\documentclass[9pt,twocolumn,twoside]{pnas-new}
\usepackage{nicefrac}
\usepackage{color}
\usepackage{tcolorbox}
\usepackage{bbm} 
\tcbuselibrary{breakable}
\usepackage{booktabs} 
\usepackage{caption,subcaption}
\usepackage{indentfirst}
\usepackage{afterpage}
\usepackage{pbox}

\newcommand{\cin}{c_\mathrm{in}}
\newcommand{\cout}{c_\mathrm{out}}
\newcommand{\eps}{\epsilon}

\newcommand{\bbone}{\ensuremath{\mathbbm{1}}}
\newcommand{\pin}{p_\mathrm{in}}
\newcommand{\pout}{p_\mathrm{out}}
\newcommand{\miin}{m_\mathrm{in}}
\newcommand{\mout}{m_\mathrm{out}}
\newcommand{\minr}{m_\mathrm{in}^{(r)}}

\newcommand{\moutrs}{m_\mathrm{out}^{(rs)}}

\usepackage{multirow}
\usepackage{multicol}
\usepackage{lipsum}
\usepackage{mathtools, cuted}
\usepackage{tikz}

\newcommand{\beginsupplement}{%
        \setcounter{table}{0}
        \renewcommand{\thetable}{S\arabic{table}}%
        \setcounter{figure}{0}
        \renewcommand{\thefigure}{S\arabic{figure}}%
        \renewcommand{\theHtable}{Supplement.\thetable}
		\renewcommand{\theHfigure}{Supplement.\thefigure}

     }
     
\renewcommand{\theHtable}{Supplement.\thetable}

\templatetype{pnasresearcharticle} 

\title{Stacking Models for Nearly Optimal Link Prediction in Complex Networks}

\author[a,b,c]{Amir Ghasemian}
\author[b]{Homa Hosseinmardi} 
\author[b]{Aram Galstyan} 
\author[c,d]{Edoardo M. Airoldi}
\author[a,e,f]{Aaron Clauset}

\affil[a]{Department of Computer Science, University of Colorado, Boulder, CO 80309, USA}
\affil[b]{Information Sciences Institute, University of Southern California, Marina del Rey, CA 90292, USA}
\affil[c]{Department of Statistics, Harvard University, Cambridge, MA 02138, USA}
\affil[d]{Department of Statistical Science, Fox School of Business, Temple University, Philadelphia, PA 19122, USA}
\affil[e]{BioFrontiers Institute, University of Colorado, Boulder, CO 80309, USA}
\affil[f]{Santa Fe Institute, Santa Fe, NM 87501, USA}
\leadauthor{Ghasemian}


\begin{abstract}
Most real-world networks are incompletely observed. Algorithms that can accurately predict which links are missing can dramatically speedup the collection of network data and improve the validity of network models. Many algorithms now exist for predicting missing links, given a partially observed network, but it has remained unknown whether a single best predictor exists, how link predictability varies across methods and networks from different domains, and how close to optimality current methods are. We answer these questions by systematically evaluating 203 individual link predictor algorithms, representing three popular families of methods, applied to a large corpus of 548 structurally diverse networks from six scientific domains. We first show that individual algorithms exhibit a broad diversity of prediction errors, such that no one predictor or family is best, or worst, across all realistic inputs. We then exploit this diversity via meta-learning to construct a series of ``stacked'' models that combine predictors into a single algorithm. Applied to a broad range of synthetic networks, for which we may analytically calculate optimal performance, these stacked models achieve optimal or nearly optimal levels of accuracy. Applied to real-world networks, stacked models are also superior, but their accuracy varies strongly by domain, suggesting that link prediction may be fundamentally easier in social networks than in biological or technological networks. These results indicate that the state-of-the-art for link prediction comes from combining individual algorithms, which achieves nearly optimal predictions. We close with a brief discussion of limitations and opportunities for further improvement of these results.
\end{abstract}

\begin{document}

\maketitle
\ifthenelse{\boolean{shortarticle}}{\ifthenelse{\boolean{singlecolumn}}{\abscontentformatted}{\abscontent}}{}

\dropcap{N}etworks provide a powerful abstraction for representing the structure of complex social, biological, and technological systems. However, data on most real-world networks is incomplete. For instance, social connections among people may be sampled, intentionally hidden, or simply unobservable~\cite{kossinets2006effects, fire2013computationally}; interactions among genes, or cells, or species must be observed or inferred by expensive experiments~\cite{lu2011link, nagarajan2015predicting}; and, connections mediated by a particular technology omit all off-platform interactions~\cite{fire2013computationally, kane2012s}. The presence of such ``missing links'' can, depending on the research question, dramatically alter scientific conclusions when analyzing a network's structure or modeling its dynamics.

Methods that accurately predict which observed pairs of unconnected nodes should, in fact, be connected have broad utility. For instance, they can improve the accuracy of predictions of future network structure and minimize the use of scarce experimental or network measurement resources~\cite{burgess2016link,mirshahvalad2012significant}. Moreover, the task of link prediction itself has become a standard for evaluating and comparing models of network structure~\cite{ghasemian2018evaluating, valles2018consistencies}, playing a role in networks that is similar to that of cross-validation in traditional statistical learning~\cite{arlot2010survey, trevor2009elements}. Hence, by helping to select more accurate network models~\cite{ghasemian2018evaluating}, methods for link prediction can shed light on the organizing principles of complex systems of all kinds.

But, predicting missing links is a statistically hard problem. Most real-world networks are relatively sparse, and the number of unconnected pairs in an observed network---each a potential missing link---grows quadratically, like $O(n^{2})$ for a network with $n$ nodes when the number of connected pairs or edges $m$ grows linearly, like $O(n)$. The probability of correctly choosing by chance a missing link is thus only $O(1/n)$---an impractically small chance even for moderate-sized systems~\cite{clauset2008hierarchical}. Despite this baseline difficulty, a plethora of link prediction methods exist~\cite{lu2011link,martinez2017survey,al2011survey}, embodied by the three main families we study here:\ (i)~topological methods~\cite{liben2007link, zhou2009predicting}, which utilize network measures like node degrees, the number of common neighbors, and the length of a shortest path; (ii)~model-based methods~\cite{clauset2008hierarchical, ghasemian2018evaluating}, such as the stochastic block model, its variants, and other models of community structure; and (iii)~embedding methods~\cite{grover2016node2vec, cai2018comprehensive}, which project a network into a latent space and predict links based on the induced proximity of its nodes.

A striking feature of this array of methods is that all appear to work relatively well~\cite{liben2007link, ghasemian2018evaluating, grover2016node2vec}. However, systematic comparisons are lacking, particularly of methods drawn from different families, and most empirical evaluations are based on relatively small numbers of networks. As a result, the general accuracy of different methods remains unclear, and we do not know whether different methods, or families, are capturing the same underlying signatures of ``missingness.'' For instance, is there a single best method or family for all circumstances? If not, then how does missing link predictability vary across methods and scientific domains, e.g., in social versus biological networks, or across network scales? And, how close to optimality are current methods?

Here, we answer these questions using a large corpus of 548 structurally and scientifically diverse real-world networks and 203 missing link predictors drawn from three large methodological families. First, we show that individual methods exploit different underlying signals of missingness, and, affirming the practical relevance of the No Free Lunch theorem~\cite{wolpert:macready:1997,peel2017ground}, no method performs best or worst on all realistic inputs. We then show that a meta-learning approach~\cite{schapire1990strength,breiman1996bagging,srivastava2014dropout} can exploit this diversity of errors by ``stacking'' individual methods into a single algorithm~\cite{wolpert1992stacked}, which we argue makes nearly optimal predictions of missing links. We support this claim with three lines of evidence:\ (i)~evaluations on synthetic data with known structure and optimal performance, (ii)~tests using real-world networks across scientific domains and network scales, and (iii)~tests of sufficiency and saturation using subsets of methods. Across these tests, model stacking is nearly always the best method on held-out links, and nearly-optimal performance can be constructed using model-based methods, topological methods, or a mixture of the two. Furthermore, we find that missing links are generally easiest to predict in social networks, where most methods perform well, and hardest in biological and technological networks. We conclude by discussing limitations and opportunities for further improvement of these results.

\section*{Methods and Materials}

As a general setting, we imagine an unobserved simple network $G$ with a set of $E$ pairwise connections among a set of $V$ nodes, with sizes $m$ and $n$, respectively. Of these, a subset $E'\subset E$ of connections is observed, chosen by some function $f$. Our task is to accurately guess, based only on the pattern of observed edges $E'$, which unconnected pairs $X=V\times V - E'$ are in fact among the missing links $Y=E - E'$. A link prediction method defines a \textit{score} function over these unconnected pairs $i,j\in X$ so that better-scoring pairs are more likely to be missing links~\cite{liben2007link}. 
In a supervised setting, the particular function that combines input predictors to produce a score is learned from the data.We evaluate the accuracy of such predictions using the standard AUC statistic, which provides a context-agnostic measure of a method's ability to distinguish a missing link $i,j\in Y$ (a true positive) from a non-edge $X-Y$ (a true negative)~\cite{clauset2008hierarchical}. Other accuracy measures may provide insight about a predictor's performance in specific settings, e.g., precision and recall at certain thresholds. We leave their investigation for future work.

The most common approach to predict missing links constructs a score function from network statistics of each unconnected node pair~\cite{liben2007link}. We study 42 of these topological predictors, which include predictions based on node degrees, common neighbors, random walks, node and edge centralities, among others (see SI Appendix, Table~\ref{table:top_feat}). Models of large-scale network structure are also commonly used for link prediction. We study 11 of these model-based methods~\cite{ghasemian2018evaluating}, which either estimate a parametric probability $\Pr(i\to j\,|\,\theta)$ that a node pair is connected~\cite{clauset2008hierarchical}, given a decomposition of a network into communities, or predict a link as missing if it would improve a measure of community structure~\cite{liben2007link} (see SI Appendix, Table~\ref{table:abbr}). Close proximity of an unconnected pair, after embedding a network's nodes into a latent space, is a third common approach to link prediction. We study 150 of these embedding-based predictors, derived from two popular graph embedding algorithms and six notions of distance or similarity in the latent space. In total, we consider 203 features of node pairs, some of which are the output of existing link prediction algorithms, while others are numerical features derived from the network structure. For our purposes, each is considered a missing link ``predictor.'' A lengthier description of these 203 methods, and the three methodological families they represent, is given in SI Appendix, section A.

Meta-learning techniques are a powerful class of machine learning algorithms that can learn from data how to combine individual predictors into a single, more accurate algorithm~\cite{breiman1996bagging,schapire1999brief}. Stacked generalization~\cite{wolpert1992stacked} combines predictors by learning a supervised model of input query characteristics and the errors that individual predictors make. In this way, model ``stacking'' treats a set of predictors as a panel of experts, and learns the kinds of questions each is most expert at answering correctly. Stacked models can thus be strictly more accurate than their component predictors~\cite{wolpert1992stacked}, making them attractive for hard problems like link prediction~\cite{koren2009bellkor}, but only if those predictors make distinct errors and are sufficiently diverse in the signals they exploit.

We evaluate individual prediction methods, and their stacked generalizations, using two types of network data. The first is a set of synthetic networks with known structure that varies along three dimensions:\ (i)~the degree distribution's variability, being low (Poisson), medium (Weibull), or high (power law); (ii)~the number of ``communities'' or modules $k\in\{1,2,4,16,32\}$; and (iii)~the fuzziness of the corresponding community boundaries~$\epsilon$, being low, medium, or high. These synthetic networks thus range from homogeneous to heterogeneous random graphs, from no modules to many modules, and from weakly to strongly modular structure (see SI Appendix, section B and Table~\ref{tab:par_set}). Moreover, because the data generating process for these networks is known, we exactly calculate the optimal accuracy that any link prediction method could achieve, as a reference point (see SI Appendix, section B).

The second is a large corpus of 548 real-world networks. This structurally diverse corpus includes social (23\%), biological (33\%), economic (22\%), technological (12\%), information (3\%), and transportation (7\%) networks~\cite{ghasemian2018evaluating}, and spans three orders of magnitude in size (see SI Appendix, section C and Fig.~\ref{fig:AD}). It is by far the largest and most diverse empirical benchmark of link prediction methods to date, and enables an assessment of how methods perform across scientific domains.

Finally, our evaluations assume a missingness function $f$ that samples edges uniformly at random from $E$ so that each edge $(i,j)\in E$ is observed with probability $\alpha$. This choice presents a hard test, as $f$ is independent of both observed edges and metadata. Other models of $f$, e.g., in which missingness correlates with edge or node characteristics, may better capture particular scientific settings and are left for future work. Our results thus provide a general, application-agnostic assessment of link predictability and method performance. In cases of supervised learning, we train a method using 5-fold cross validation by choosing as positive examples a subset of edges $E'' \subset E'$ according to the same missingness model $f$, along with all observed non-edges $V\times V - E'$ as negative examples (see SI Appendix, section D). Unless other specified, results reflect a choice of $\alpha=0.8$, i.e., 20\% of edges are unobserved (holdout set); other values produce qualitatively similar results.

\section*{Results}
\label{sec:R}

\begin{figure*}[t!]
\centering
\begin{tabular}{cc}
\includegraphics[width=1\textwidth]{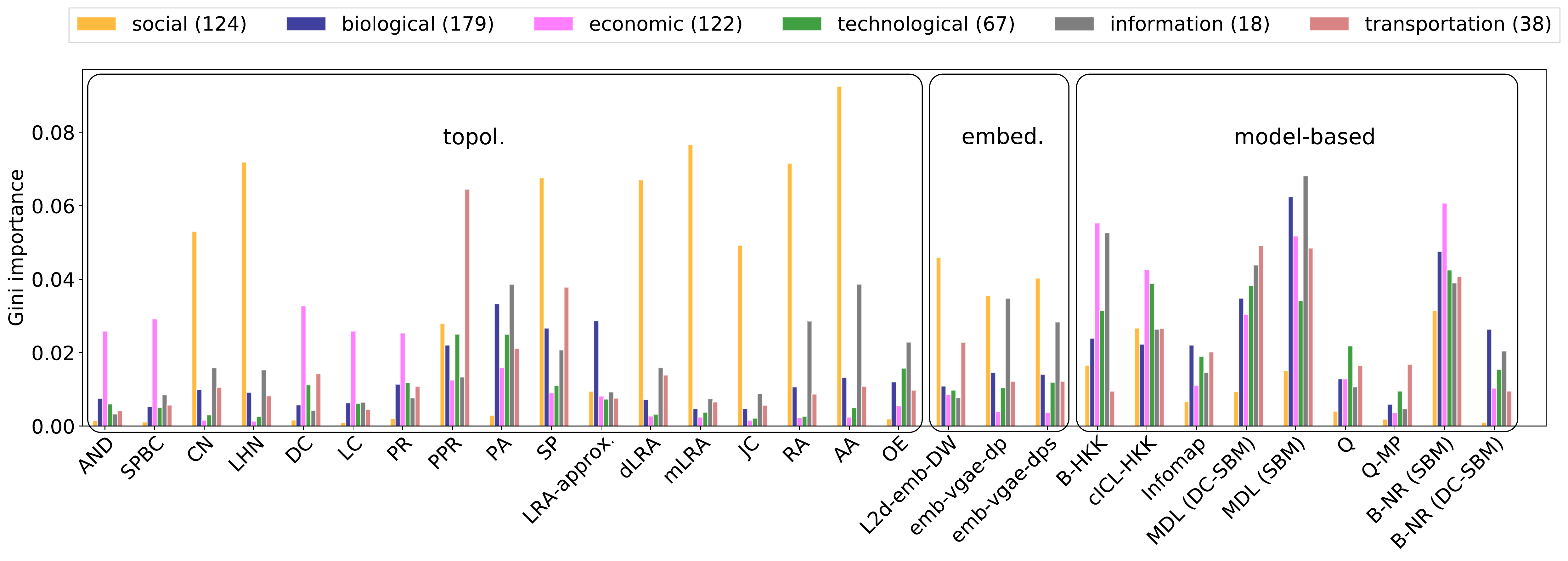}
\vspace{-3mm}
\end{tabular}
\caption{The Gini importances for predicting missing links in networks within each of six scientific domains, for the 29 most important predictors, grouped by family, under a random forest classifier trained over all 203 predictors. Across domains, predictors exhibit widely different levels of importance, indicating a diversity of errors, such that no predictor is best overall. Here, topological predictors include shortest-path betweenness centrality (SPBC), common neighbors (CN), Leicht-Holme-Newman index (LHN), personalized page rank (PPR), shortest path (SP), the mean neighbor entries within a low rank approximation (mLRA), Jaccard coefficient (JC), and the Adamic-Adar index (AA); embedding predictors include the L2 distance between embedded vectors under \mbox{emb-DW} (\mbox{L2d-emb-DW}), and the dot product (\mbox{emb-vgae-dp}) of embedded vectors under \mbox{emb-vgae}; and, model-based predictors include Infomap (Infomap), stochastic block models with (\mbox{MDL (DC-SBM), B-NR (DC-SBM)}) and without degree corrections (\mbox{MDL (SBM), B-NR (SBM)}), and modularity (Q). (A complete list of abbreviations is given in SI Appendix, Section A.)}
\label{fig:FID}
\end{figure*}

\subsection*{Prediction Error Diversity}

If all link predictors exploit a common underlying signal of missingness, then one or a few predictors will consistently perform best across realistic inputs. Optimal link prediction could then be obtained by further leveraging this universal signal. In contrast, if different predictors exploit distinct signals, they will exhibit a diversity of errors in the form of heterogenous performance across inputs, In this case, there will be no best or worst method overall, and optimal link predictions can only be obtained by combining multiple methods. This dichotomy also holds at the level of predictor families, one of which could be best overall, e.g., topological methods, even if no one family member is best.

To distinguish these possibilities, we characterize the empirical distribution of errors by training a random forest classifier over the 203 link predictors applied to each of the 548 real-world networks and separately to all networks in each of the six scientific domains within our corpus (see SI Appendix section E). In this setting, the character of a predictor's errors is captured by its learned Gini importance (mean decrease in impurity)~\cite{trevor2009elements} within the random forest:\ the higher the Gini importance, the more generally useful the predictor is for correctly identifying missing links on that network or that domain. If all methods exploit a common missingness signal (one method to rule them all), the same few predictors or predictor family will be assigned consistently greater importance across networks and domains. However, if there are multiple distinct signals (a diversity of errors), the learned importances will be highly heterogeneous across inputs, and no predictor or family will be best.

Across networks and domains, we find wide variation in both individual and family-wise predictor importances, such that no individual method and no family of methods is best, or worst, on all networks. On individual networks, predictor importances tend to be highly skewed, such that a relatively small subset of predictors account for the majority of prediction accuracy (SI Appendix, Table~\ref{tab:FIE} and Fig.~\ref{fig:LCF}).  
However, the precise composition of this subset varies widely across both networks and families (SI Appendix, Tables \ref{tab:FIE}--\ref{tab:FWE}, and Figs. \ref{fig:HFR}--\ref{fig:CFB}), implying a broad diversity of errors and multiple distinct signals of missingness. At the same time, not all predictors perform well on realistic inputs, e.g., a subset of topological methods generally receive low importances, and most embedding-based predictors are typically mediocre. Nevertheless, each family contains some members that are ranked among the most important predictors for many, but not all, networks.

Across domains, predictor importances cluster in interesting ways, such that some individual and some families of predictors perform better on specific domains. For instance, examining the 10 most-important predictors by domain (29 unique predictors; Fig.~\ref{fig:FID}), we find that topological methods, such as those based on common neighbors or localized random walks, perform well on social networks but less well on networks from other domains. In contrast, model-based methods perform relatively well across domains, but often perform less well on social networks than do topological measures and some embedding-based methods. Together, these results indicate that predictor methods exhibit a broad diversity of errors, which tend correlate somewhat with scientific domain.

This performance heterogeneity highlights the practical relevance to link prediction of the general No Free Lunch theorem~\cite{wolpert:macready:1997}, which proves that across all possible inputs, every machine learning method has the same average performance, and hence accuracy must be assessed on a per dataset basis. The observed diversity of errors indicates that none of the 203 individual predictors is a universally-best method for the subset of all inputs that are realistic. However, that diversity also implies that a nearly-optimal link prediction method for realistic inputs could be constructed by combining individual methods so that the best individual method is applied for each given input. Such a meta-learning algorithm cannot circumvent the No Free Lunch theorem, but it can achieve optimal performance on realistic inputs by effectively redistributing its worse-than-average performance onto unrealistic inputs, which are unlikely to be encountered in practice. In the following sections, we develop and investigate the near-optimal performance of such an algorithm.

\begin{figure*}[t!]
\centering
\begin{tabular}{cc}
\includegraphics[width=1\textwidth]{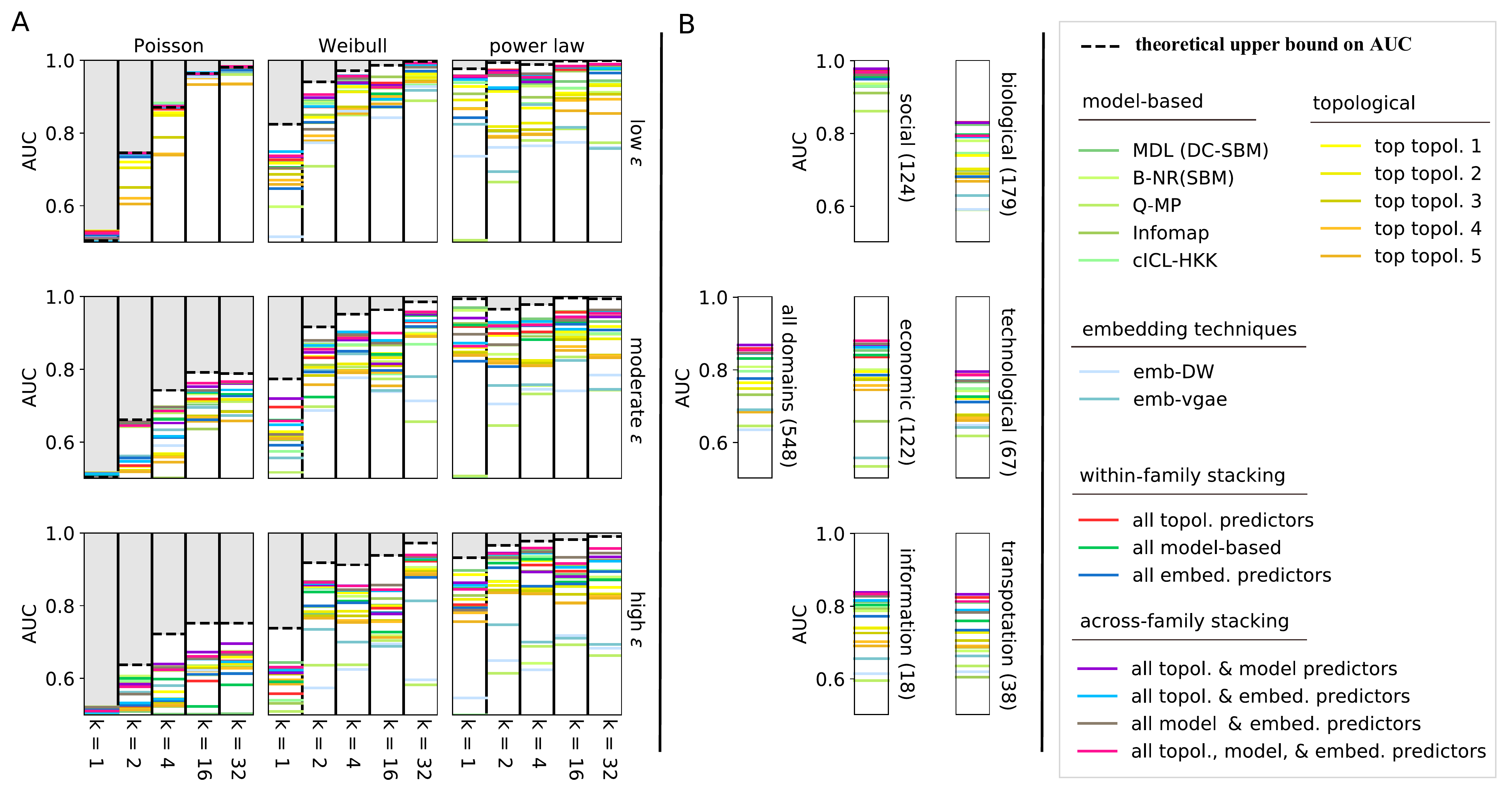}
\vspace{-3mm}
\end{tabular}
\caption{(A) On synthetic networks, the mean link prediction performance (AUC) of selected individual predictors and all stacked algorithms across three forms of structural variability:\ (left to right, by subpanel) degree distribution variability, from low (Poisson) to high (power law); (top to bottom, by subpanel) fuzziness of community boundaries, ranging from low to high ($\epsilon=\mout/\miin$, the fraction of a node's edges that connect outside its community); and (left to right, within subpanel) the number of communities $k$. Across settings, the dashed line represents the theoretical maximum performance achievable by any link prediction algorithm (SI Appendix, section B). In each instance, stacked models perform optimally or nearly optimally, and generally perform better when networks exhibit heavier-tailed degree distributions and more communities with distinct boundaries. Table~S11 lists the top five topological predictors for each synthetic network setting, which vary considerably.
(B) On real-world networks, the mean link prediction performance for the same predictors across all domains, and by individual domain. Both overall and within each domain, stacked models, particularly the across-family versions, exhibit superior performance, and they achieve nearly perfect accuracy on social networks. The performance, however, varies considerably across domains, with biological and technological networks exhibiting the lowest link predictability. Due to space limitations here, more complete results for individual topological and model-based predictors are shown in SI Appendix, Figs.~S8 and~S9, respectively.}
\label{fig:exp_PL_LN}
\end{figure*}



\subsection*{Stacking on Networks with Known Structure}
Model ``stacking'' is a meta-learning approach that learns to apply the best individual predictor according to the input's characteristics~\cite{wolpert1992stacked}. Here, we assess the accuracy of  model stacking both within and across families of prediction methods, which adds seven more prediction algorithms to our evaluation set.

Because the optimality of an algorithm's predictions can only be assessed when the underlying data generating process is known, we first characterize the accuracy of model stacking using synthetic networks with known structure, for which we calculate an exact upper bound on link prediction accuracy (see SI Appendix, section B). To provide a broad range of realistic variation in these tests, we use a structured random graph model, in which we systematically vary its degree distribution's variance, the number of communities $k$, and the fuzziness of those community boundaries $\eps$.

Across these structural variables, the upper limit on link predictability varies considerably (Fig.~\ref{fig:exp_PL_LN}A), from no better than chance in a simple random graph ($k=1$; Poisson) to nearly perfect in networks with many distinct communities and a power-law degree distribution. Predictability is generally lower (no methods can do well) with fewer communities (low $k$) or with more fuzzy boundaries (high $\eps$), but higher with increasing variance in the degree distribution (Weibull or power law). 
Most methods, whether stacked or not, perform relatively well when predictability is low. However, as potential predictability increases, methods exhibit considerable dispersion in their accuracy, particularly among topological and embedding-based methods.

Regardless of the synthetic network's structure, however, we find that stacking methods are typically among the most accurate prediction algorithms, and they often achieve optimal or nearly-optimal prediction accuracy (Fig.~\ref{fig:exp_PL_LN}A). For instance, the best model stacking method exhibits a substantially smaller gap between practical and optimal performance (all topol., model \& embed., $\Delta\textrm{AUC}=0.04$; SI Appendix, Table~\ref{tab:gap}) than the best individual predictor (MDL (DC-SBM), $\Delta\textrm{AUC}=0.07$; SI Appendix, Table~\ref{tab:best_10_AUCgap}), and is 
far better than the average non-stacked topological and model-based methods ($\langle \Delta\textrm{AUC}\rangle=0.23$; SI Appendix, Table~\ref{tab:gap}). Moreover, in all structural settings, stacking across families tends to produce slightly more accurate predictions ($\langle\textrm{AUC}\rangle = 0.83$; SI Appendix, Table~\ref{tab:AUC_synt}) than stacking within families ($\langle\textrm{AUC}\rangle = 0.80$), and only one stacked model (all embed.)\ is less accurate than the best individual predictor (marginally, with $\Delta\textrm{AUC}=0.01$, and Table~\ref{tab:AUC_synt}).

\subsection*{Stacking on Real-world Networks}
To characterize the real-world accuracy of model stacking, we apply these methods, along with the individual predictors, to our corpus of 548 structurally diverse real-world networks. We analyze the results both within and across scientific domains, and as a function of network size.

Both across all networks, and within individual domains, model stacking methods produce the most accurate predictions of missing links (Fig.~\ref{fig:exp_PL_LN}B and Table~\ref{tab:satexp}), and some individual predictors  perform relatively well, particularly model-based ones. Applied to all networks, the best model-stacking method achieves slightly better average performance (all topol.\ \& model, $\langle\textrm{AUC}\rangle = 0.87\pm0.10$) than the best individual method (MDL (DC-SBM), $\langle\textrm{AUC}\rangle = 0.84\pm0.10$), and far better performance than the average individual topological or model-based predictor ($\langle\textrm{AUC}\rangle = 0.63$; and see Tables~\ref{tab:satexp} and~\ref{tab:APR_indiv_real}). However, model stacking also achieves substantially better precision in its predictions (Table~\ref{tab:satexp}), which can be a desirable property in practice. We note that these stacking results were obtained by optimizing the standard F measure to choose the random forest's parameters. Alternatively, we may optimize the AUC itself, which produces similar results, but with slightly lower precisions in exchange for slightly higher AUC scores (see Table~\ref{tab:rf_AUC}).

\begin{table}[t!]
\centering
\caption{Link prediction performance (mean$\pm$std.\ err.), measured by AUC, precision, and recall, for link prediction algorithms applied to the 548 structurally diverse networks in our corpus.} 
\begin{tabular}{l|c|c|c}
\hline \hline
algorithm & AUC & precision & recall\\ 
\hline\hline
Q &   $ 0.7\pm0.14$  & $ 0.14 \pm0.17$  & $ 0.67\pm0.15$  \\ \hline    
Q-MR &  $0.67\pm 0.15$  & $0.12\pm  0.17$  & $0.63\pm0.13$ \\ \hline
Q-MP &  $0.64\pm 0.15$ &  $0.09\pm 0.11$  & $0.59\pm  0.17$ \\ \hline
B-NR (SBM) &  $0.81\pm 0.13$ &  $0.13\pm0.12$ &  $0.65\pm0.22$ \\ \hline    
B-NR (DC-SBM)  & $ 0.7\pm0.2$  & $0.12\pm0.12$ &  $0.61\pm0.24$ \\ \hline
cICL-HKK  &  $ 0.79\pm0.13$ &  $0.14\pm0.14$  & $ 0.58\pm 0.25$ \\ \hline
B-HKK  & $0.77\pm0.13$ &  $0.11\pm 0.1$  & $ 0.51\pm0.26$  \\ \hline
Infomap  &  $0.73\pm0.14$  & $ 0.12\pm0.12$ &  $0.68\pm0.13$  \\ \hline
MDL (SBM)  &  $ 0.79\pm 0.15$  & $0.14\pm0.13$  & $0.57\pm 0.3$  \\ \hline
MDL (DC-SBM)  &  $ 0.84\pm 0.1$  & $0.13\pm0.11$  & $0.78\pm0.12$ \\ \hline
S-NB  &   $ 0.71\pm0.19$   &  $ 0.12\pm0.13$  &  $0.66\pm0. 17$ \\ \hline \hline
mean model-based &  $ 0.74  \pm 0.16 $ & $ 0.12  \pm  0.13 $ & $ 0.63 \pm 0.21 $ \\ \hline
mean indiv. topol. &  $ 0.6  \pm 0.13 $ & $ 0.09  \pm  0.16 $ & $ 0.53 \pm 0.35 $ \\ \hline
mean indiv. topol. \& model &  $ 0.63  \pm 0.15 $ & $ 0.09  \pm  0.16 $ & $ 0.55 \pm 0.33 $ \\ \hline \hline
emb-DW  &  $0.63\pm0.23$  &  $0. 17\pm0.19$ & $0.42\pm0. 35$ \\ \hline
emb-vgae  &  $0.69\pm 0.19$  &  $0.05\pm 0.05$ & $0.69\pm0.21$ \\ \hline \hline
all topol.  &  $0.86\pm0. 11$ & $0.42\pm0.33$ &   $0.44\pm0.32$\\ \hline
all model-based &  $0.83\pm0.12$  & $0.39\pm0.34$ & $0.3\pm0.29$ \\ \hline
all embed. & $0. 77\pm 0.16$  & $0. 32\pm0. 32$ & $0.32\pm0.31$ \\ \hline
all topol. \& model &  $0.87\pm0.1$   & $0.48\pm0.36$ & $0.35\pm0.35$ \\ \hline
all topol. \& embed.   & $0. 84\pm0. 13$  & $0. 4\pm0. 34$ &  $0.39\pm0.3 3$ \\ \hline
all model \& embed.   &  $0.84\pm0.13$   &  $0.36\pm0.32$  &   $0.36\pm0.31$  \\ \hline
all topol., model \& embed. & $0.85\pm0. 14$ & $0.42\pm0.34$ &  $0.39\pm0.33$ \\ \hline
	\end{tabular}
	\label{tab:satexp}  
\end{table} 

Among the stacked models, the highest accuracy on real-world networks is achieved by stacking model-based and topological predictor families. Adding embedding-based predictors does not significantly improve  accuracy, suggesting that the network embeddings do not capture more structural information than is represented by the model-based and topological families. This behavior aligns with our results on synthetic networks above, where the performances of stacking all predictors and stacking only model-based and topological predictors were nearly identical (SI Appendix, Tables~\ref{tab:gap} and~\ref{tab:best_10_AUCgap}).

Applied to individual scientific domains, we find considerable variation in missing link predictability, which we take to be approximated by the most-accurate stacked model (Fig.~\ref{fig:exp_PL_LN}B). In particular, most predictors, both stacked and individual (SI Appendix, Figs.~\ref{tab:gap} and~\ref{tab:best_10_AUCgap}), perform well on social networks, and on these networks, model stacking achieves nearly perfect link prediction (up to $\textrm{AUC}=0.98\pm0.06$; Table~\ref{tab:AUC_soc_dom}). In contrast, this upper limit is substantially lower in non-social domains, being lowest for biological and technological networks ($\textrm{AUC}=0.83\pm0.10$; Tables~\ref{tab:AUC_bio_dom} and~\ref{tab:AUC_tech_dom}), while marginally higher for economic and information networks ($\textrm{AUC}=0.88\pm0.10$; SI Appendix, Tables~\ref{tab:AUC_eco_dom} and~\ref{tab:AUC_info_dom}).

\begin{figure}[t!]
\centering
\includegraphics[width=0.98\columnwidth]{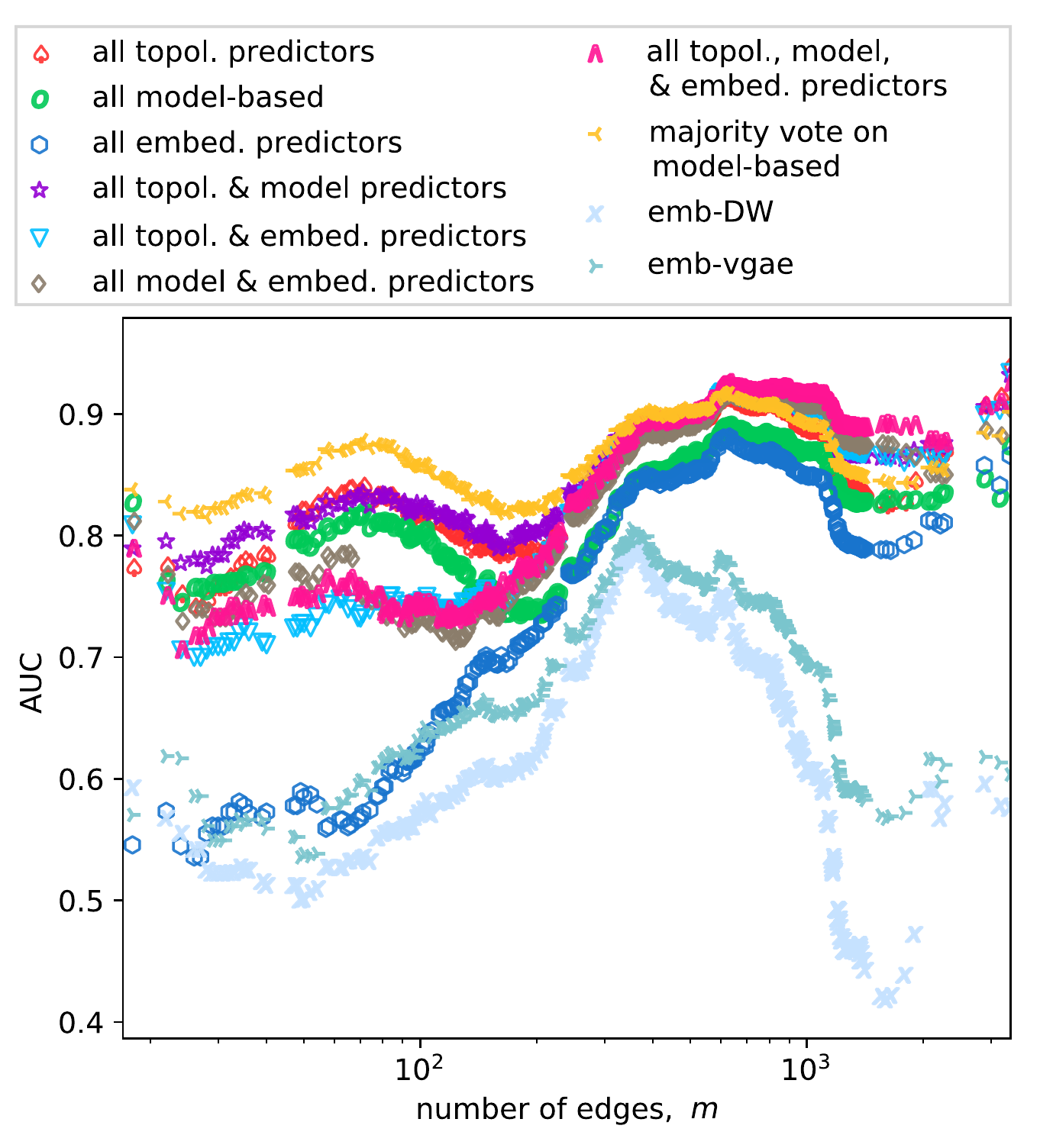}
\label{fig:aveAUC_vs_size_real}
\vspace{-3mm}
\caption{Mean link prediction performance (AUC) as a function of network size (number of edges $m$) for stacked models and select individual predictors, applied to 548 real-world networks. Generally, stacking topological predictors, model-based predictors, or both yields superior performance, but especially on larger networks where link prediction is inherently more difficult.}
\label{fig:aveAUC_vs_size}
\end{figure}

Stacked models also exhibit superior performance on link prediction across real-world networks of different scales (number of edges $m$; Fig.~\ref{fig:aveAUC_vs_size}), and generally exhibit more accurate predictions as network size increases, where link prediction is inherently harder. We note, however, that on small networks ($m<200$), an alternative algorithm based on a simple majority-vote among model-based predictors slightly outperforms all stacking methods, but performs substantially worse than the best stacked model on larger networks ($m>1000$). And, embedding-based methods perform poorly at most scales, suggesting a tendency to overfit, although stacking within that family produces higher accuracies on larger networks, but still lower than other stacked models.

\subsection*{Sufficiency and Optimality}
In practice, the optimality of a meta-learning method can only be established indirectly, over a set of considered predictors applied to a sufficiently diverse range of empirical tests cases~\cite{wolpert:macready:1997}. We assess this indirect evidence for stacked link-prediction models through two numerical experiments.

In the first, we consider how performance varies as a function of the number of predictors stacked, either within or across families. Evidence for optimality here appears as an early saturation, in which performance achieves its maximum prior to the inclusion of all available individual predictors. This behavior would indicate that a subset of predictors is sufficient to capture the same information as the total set. To test for this early-saturation signature, we first train a random forest classifier on all predictors in each of our stacked models and calculate each predictor's within-model Gini importance. For each stacked model, we then build a new sequence of sub-models in which we stack only the $k$ most important predictors at a time and assess its performance on the test corpus.

\begin{figure}[t!]
\centering
\includegraphics[width=0.98\columnwidth]{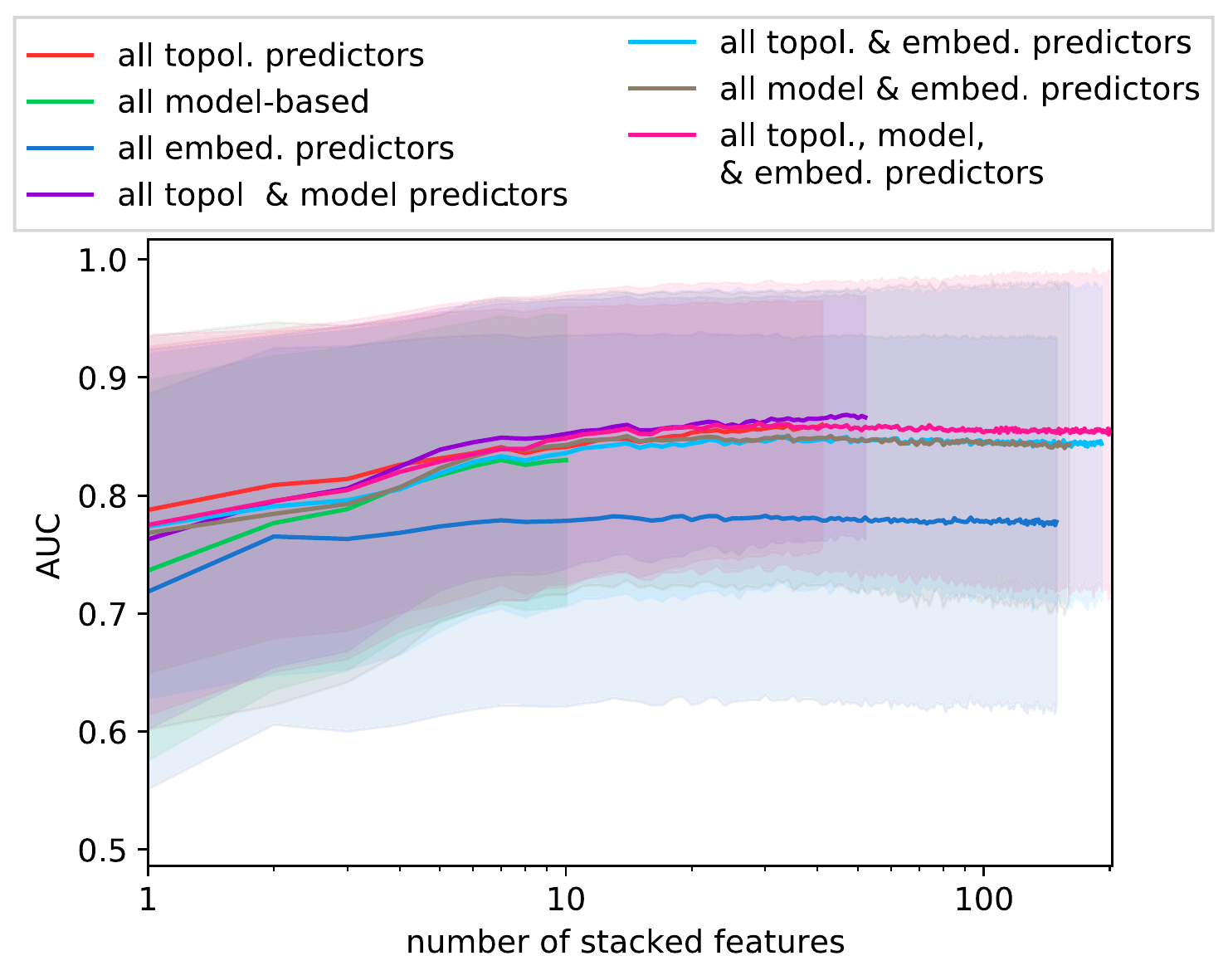}
\caption{Mean link prediction performance (AUC) as a function of the number of stacked features, for within- and across-family stacked models, applied to 548 real-world networks. The shaded regions show the standard error, and the early saturation behavior (at between 10 and 50 predictors) indicates that a small subset of predictors is sufficient to capture the same information as the total set.}
\label{fig:AUC_saturation}
\end{figure}

In each of the stacked models, performance exhibits a classic saturation pattern:\ it increases quickly as the 10 most-important predictors are included, and then stabilizes by around 30 predictors (Fig.~\ref{fig:AUC_saturation} and SI Appendix, Fig.~\ref{fig:DKstar}). Performance then degrades slightly beyond 30--50 included predictors, suggesting a slight degree of overfitting in the full models. Notably, each within and across family model exhibits a similar saturation curve, except for the embedding-only model, which saturates early and at a lower level than other stacked models. This similar behavior suggests that these families of predictors are capturing similar missingness signals, despite their different underlying representations of the network structure. As in other experiments, the best saturation behavior is achieved by stacking model-based and topological predictors.

In the second, we evaluate whether individual predictors represent ``weak'' learners in the sense that their link-prediction performance is better than random. In general, we find that nearly all of the predictors satisfy this condition (SI Appendix, Figs.~\ref{fig:HWL_me} and~\ref{fig:HWL_top}), implying that they can be combined according to the Adaboost theorem to construct an optimal algorithm~\cite{freund:schapire:1997}. Replacing the random forest algorithm within our stacking approach with a standard boosting algorithm also produces nearly identical performance on our test corpus (see Tables~\ref{tab:xgb_f}--\ref{tab:ada_AUC}). The similar performance between the two methods suggests that relatively little additional performance is likely possible using other meta-learning approaches over the same set of predictors.

\section*{Discussion}
Developing more accurate methods for predicting missing links in networks would help reduce the use of scarce resources in collecting network data, and would provide more powerful tools for evaluating and comparing network models of complex systems. The literature on such methods gives an unmistakable impression that most published algorithms produce reasonably accurate predictions. However, relatively few of these studies present systematic comparisons across different families of methods and they typically draw their test cases from a narrow set of empirical networks, e.g., social networks. As a result, it has remained unknown whether a single best predictor or family of predictors exists, how link predictability itself varies across different methods and scientific domains, or how close to optimality current methods may be.

Our broad analysis of individual link prediction algorithms, representing three large and popular families of such methods, applied to a large corpus of structurally diverse networks, shows definitively that common predictors in fact exhibit a broad diversity of errors across realistic inputs (Fig.~\ref{fig:FID} and SI Appendix, Fig.~\ref{fig:LCF}). Moreover, this diversity is such that no one predictor, and no family of predictors is overall best, or worst, in practice (SI Appendix, Table~\ref{tab:FIE} and Fig.~\ref{fig:HFR}). The common practice of evaluating link prediction algorithms using a relatively narrow range of test cases is thus problematic. The far broader range of empirical networks and algorithms considered here shows that, generally speaking, good performance on a few test cases does not generalize across inputs. The diversity of errors we find serves to highlight the practical relevance of the No Free Lunch theorem~\cite{wolpert:macready:1997} for predicting missing links in complex networks, and suggests that optimal performance on realistic inputs may only be achieved by combining methods, e.g., via meta-learning, to construct an ensemble whose domain of best performance matches the particular structural diversity of real-world networks.

Model stacking is a popular meta-learning approach, and our results indicate that it can produce highly accurate predictions of missing links by combining either topological predictors alone, model-based predictors alone, or both. Applied to structurally diverse synthetic networks, for which we may calculate optimal performance, stacking achieves optimal or near-optimal accuracy, and accuracy is generally closer to perfect when networks exhibit a highly variable degree distribution and/or many, structurally distinct communities (Fig.~\ref{fig:exp_PL_LN}A).

Similarly, applied to empirical networks, stacking produces more accurate predictions than any individual predictor (Fig.~\ref{fig:exp_PL_LN}B and Table~\ref{tab:satexp}), and these predictions appear to be nearly optimal, i.e., we find little evidence that further accuracy can be achieved using this set of predictors (Fig.~\ref{fig:AUC_saturation}), even under alternative meta-learning approaches. Of course, we cannot rule out the possibility that more accurate predictions overall could be obtained by incorporating, within the stacked models, specific new predictors or new families, if they provide better prediction coverage of some subset of input networks than do the currently considered predictors. Given the diverse set of predictors and families considered here, this possibility seems unlikely without fundamentally new ideas about how to represent the structure of networks, and therefore also signals of missingness.

Across networks drawn from different scientific domains, e.g., social vs.\ biological networks, we find substantial variation in link predictor performance, both for individual predictors and for stacked models. This heterogeneity suggests that the basic task of link prediction may be fundamentally harder in some domains of networks than others. Most algorithms produce highly accurate predictions in social networks, which are stereotypically rich in triangles (local clustering), exhibit broad degree distributions, and are composed of assortative communities, suggesting that link prediction in social networks may simply be easier~\cite{epasto2019single} than in non-social network settings. In fact, stacked models achieve nearly perfect accuracy at distinguishing true positives (missing links) from true negatives (non-edges) in social networks (Fig.~\ref{fig:exp_PL_LN}B and SI Appendix, Table~\ref{tab:AUC_soc_dom}). An alternative interpretation of this difference is that the existing families of predictors exhibit some degree of selective inference, i.e., they work well on social networks because social network data is the most common inspiration and application for link prediction methods. Our results make it clear that developing more accurate individual predictors for non-social networks, e.g., biological and informational networks, is an important direction of future work. Progress along these lines will help clarify whether link prediction is fundamentally harder in non-social domains, and why.

Across our analyses, embedding-based methods, which are instances of representation learning on networks, generally perform more poorly than do either topological or model-based predictors. This behavior is similar to recent results in statistical forecasting, which found that neural network and other machine learning methods perform less well by themselves than when combined with other, conventional statistical methods~\cite{makridakis2018m4,makridakis2018statistical}. A useful direction of future work on link prediction would specifically investigate tuning embedding-based methods to perform better on the task of link prediction.

Only strong theoretical guarantees, which currently seem out of reach, would allow us to say for certain whether the stacked models presented here actually achieve the upper bound on link prediction performance in complex networks. However, the evidence suggests that stacking achieves nearly optimal performance across a wide variety of realistic inputs. It is likely that efforts to develop new individual link prediction algorithms will continue, and these efforts will be especially beneficial in specific application domains, e.g., predicting missing links in genetic regulatory networks or in food webs. Evaluations of new predictors, however, should be carried out in the context of meta-learning, in order to assess whether they improve the overall prediction coverage embodied by the state-of-the-art stacked models applied to realistic inputs. Similarly, these evaluations should be conducted on a large and structurally diverse corpus of empirical networks, like the one considered here. More narrow evaluations are unlikely to produce reliable estimates of predictor generalization. Fortunately, stacked models can easily be extended to incorporate any new predictors, as they are developed, providing an incremental path toward fully optimal predictions.

\acknow{The authors thank David Wolpert, Brendan Tracey, and Cristopher Moore for helpful conversations, acknowledge the BioFrontiers Computing Core at the University of Colorado Boulder for providing High Performance Computing resources (NIH 1S10OD012300) supported by BioFrontiers IT, and thank the Information Sciences Institute at the University of Southern California for hosting AGh during this project. Financial support for this research was provided in part by Grant No. IIS-1452718 (AGh, AC) from the National Science Foundation. Data and code for replication purposes are provided at [https://github.com/Aghasemian/OptimalLinkPrediction].}

\showacknow{}



\section*{Supporting Information}
\beginsupplement

\section{Methods for predicting missing links}
\label{sec:A}
Here, we describe in detail the three families of link predictors and their specific members used in the analysis, including the abbreviations used in the main text. In addition, we describe in more detail the setup of the supervised stacked generalization algorithm we use to combine individual predictors into a single algorithm.

\subsection*{Topological predictors}
Topological predictors are simple functions of the observed network topology, e.g., counts of edges, measures of overlapping sets of neighbors, and measures derived from simple summarizations of the network's structure. We consider 42 topological predictors, which come in three types:\ global, pairwise, and node-based. Within these groups, the ``pairwise'' predictors include a number of topological features that are often used in the literature to directly predict missing links~\cite{liben2007link}, e.g., the number of shared neighbors of $i,j$. A listing of all topological predictors is given in Table~\ref{table:top_feat}, along with corresponding literature references.

\smallskip \noindent \textit{Global predictors.}
These predictors quantify various network-level statistics and are inherited by each pair of nodes $i,j$ that is a candidate missing link. Their primary utility is to provide global context to other predictors under supervised learning. For example, a predictor that performs well on small networks, but poorly on larger networks, can be employed appropriately under a supervised model when the global measure of the network's size is available. Or, a large variance in the degree distribution would imply that a predictor based on degree product may be useful. Or, a large clustering coefficient would imply that an assortative community detection algorithm like modularity is likely to be useful. For this reason, global predictors are not expected by themselves to be accurate predictors of missing links (see Figs.~\ref{fig:HFR} and~\ref{fig:CFB} and Tables~\ref{tab:APR_indiv_real} and~\ref{tab:APR_indiv_synt}). These global predictors are generally useful in capturing missingness in unseen networks and not on the same network link prediction experiments. These features help to learn from existing configurations in training networks and generalize them to unseen networks in experiments like Fig.~\ref{fig:FID} in the main text.

The 8 global predictors are the number of nodes (N), number of observed edges (OE), average degree (AD), variance of the degree distribution (VD), network diameter (ND), degree assortativity of graph (DA), network transitivity or clustering coefficient (NT), and average (local) clustering coefficient (ACC)~\cite{newman:networks:2018,liben2007link, al2011survey, cukierski2011graph}.

\begin{table*}[b!]\addtolength{\tabcolsep}{-5pt}
\caption{Abbreviations and descriptions of 42 topological predictors, across three types:\ \textit{global} predictors (7), which are functions of the entire network and whose utility is in providing context to other predictors; \textit{pairwise} predictors (15), which are functions of the joint topological properties of the pair $i,j$; and \text{node-based} predictors (20), which are functions of the independent topological properties of the nodes $i$ and $j$, producing one value for each node in the pair $i,j$.}
\vspace*{-3mm}
\centering
 \begin{tabular}{|p{2.42cm} | p{10.6cm} |c|c|c|c|} 
 
 \hline
Abbreviation &  Description & Global & Pairwise & Node-based & ~~Ref.~~  \\ [0.5ex] 
 \hline\hline
 N & number of nodes & $\bullet$ & & & \cite{newman:networks:2018} \\ \hline
 OE & number of observed edges & $\bullet$ & & & \cite{newman:networks:2018} \\ \hline
 AD & average degree & $\bullet$ & & & \cite{newman:networks:2018} \\ \hline
 VD & variance of degree distribution & $\bullet$ & & & \cite{newman:networks:2018} \\ \hline
 ND & network diameter & $\bullet$ & & & \cite{newman:networks:2018} \\ \hline
 DA & degree assortativity of graph & $\bullet$ & & & \cite{networkx} \\ \hline
 NT & network transitivity (clustering coefficient) & $\bullet$ & & & \cite{newman:networks:2018} \\ \hline
 ACC & average (local) clustering coefficient & $\bullet$ & & & \cite{newman:networks:2018} \\ [0.5ex]  \hline\hline
CN & common neighbors of $i,j$ & & $\bullet$ & & \cite{liben2007link} \\ \hline
SP & shortest path between $i,j$ & & $\bullet$ & & \cite{liben2007link} \\ \hline
LHN & Leicht-Holme-Newman index of neighbor sets of $i,j$& & $\bullet$ & & \cite{leicht2006vertex} \\ \hline
PPR & $j$-th entry of the personalized page rank of node $i$ & & $\bullet$ & & \cite{networkx} \\ \hline
PA & preferential attachment (degree product) of $i,j$ & & $\bullet$ & & \cite{liben2007link} \\ \hline
JC & Jaccard's coefficient of neighbor sets of $i,j$ & & $\bullet$ & & \cite{liben2007link} \\ \hline
AA & Adamic/Adar index of $i,j$ & & $\bullet$ & & \cite{liben2007link} \\ \hline
RA & resource allocation index of $i,j$ & & $\bullet$ & & \cite{networkx} \\ \hline
LRA & entry $i,j$ in low rank approximation (LRA) via singular value decomposition (SVD) & & $\bullet$ & & \cite{cukierski2011graph} \\ \hline
dLRA & dot product of columns $i$ and $j$ in LRA via SVD for each pair of nodes $i,j$ & & $\bullet$ & & \cite{cukierski2011graph} \\ \hline
mLRA & average of entries $i$ and $j$'s neighbors in low rank approximation & & $\bullet$ & & \cite{cukierski2011graph} \\ \hline
LRA-approx & an approximation of LRA & & $\bullet$ & & \cite{cukierski2011graph} \\ \hline
dLRA-approx & an approximation of dLRA & & $\bullet$ & & \cite{cukierski2011graph} \\ \hline
mLRA-approx & an approximation of mLRA & & $\bullet$ & & \cite{cukierski2011graph} \\ [0.5ex]  \hline\hline
LCC$_i$, LCC$_j$ & local clustering coefficients for $i$ and $j$ & &  & $\bullet$ & \cite{networkx} \\ \hline
AND$_i$, AND$_j$ & average neighbor degrees for $i$ and $j$ & &  & $\bullet$ & \cite{networkx} \\ \hline
SPBC$_i$, SPBC$_j$ & shortest-path betweenness centralities for $i$ and $j$ & &  & $\bullet$ & \cite{networkx} \\ \hline
CC$_i$, CC$_j$ & closeness centralities for $i$ and $j$ & &  & $\bullet$ & \cite{networkx}  \\ \hline
DC$_i$, DC$_j$ & degree centralities for $i$ and $j$ & &  & $\bullet$ & \cite{networkx} \\ \hline
EC$_i$, EC$_j$ & eigenvector centralities for $i$ and $j$ & &  & $\bullet$ & \cite{networkx} \\ \hline
KC$_i$, KC$_j$ & Katz centralities for $i$ and $j$ & &  & $\bullet$ & \cite{networkx} \\ \hline
LNT$_i$, LNT$_j$ & local number of triangles for $i$ and $j$ & &  & $\bullet$ & \cite{networkx} \\ \hline
PR$_i$, PR$_j$ & Page rank values for $i$ and $j$ & &  & $\bullet$ & \cite{networkx} \\ \hline
LC$_i$, LC$_j$ & load centralities for $i$ and $j$ & &  & $\bullet$ & \cite{networkx} \\ \hline
 \end{tabular}
 \label{table:top_feat}
\end{table*}

\smallskip \noindent \textit{Pairwise predictors.}
These predictors are functions of the joint topological properties of the pair of nodes $i,j$ being considered.

The 14 pairwise predictors are the number of common neighbors of $i,j$ (CN), shortest path between $i,j$ (SP), Leicht-Holme-Newman index of neighbor sets of $i,j$ (LHN), personalized page rank (PPR),%
\footnote{By using a biased random walk we can find the personalized PageRank algorithm. This centrality could reflect the importance of nodes with respect to a biased set of specific nodes or a single specific node.
In this paper using personalized page rank we consider $j$-th entry of the personalized page rank for node $i$ as one of the edge-based features.}
preferential attachment or degree product of $i,j$ (PA), Jaccard coefficient of the neighbor sets of $i,j$ (JC), Adamic-Adar index of $i,j$ (AA), resource allocation index of $i,j$ (RA), the entry $i,j$ in a low rank approximation (LRA) via a singular value decomposition (SVD) (LRA), the dot product of the $i,j$ columns in the LRA via SVD (dLRA), the mean of entries $i$ and $j$'s neighbors in the LRA (mLRA), and simple approximations of the latter three predictors (LRA-approx, dLRA-approx, mLRA-approx)~\cite{newman:networks:2018,liben2007link, al2011survey, cukierski2011graph}.

We omit from consideration several pairwise predictors found in the literature, e.g. edge betweenness centrality, due to their large computational complexity for an evaluation as large as ours.

\smallskip \noindent \textit{Node-based predictors.} 
These predictors are functions of the independent topological properties of the individual nodes $i$ and $j$, and thus produce a pair of predictor values. Unlike many of the pairwise predictors, which can be used as standalone algorithms to predict missing links, these node-based predictors do not directly score the likelihood that $i,j$ is a missing link. Instead, the particular function that converts the pair of node-based predictors into a score is learned within the supervised framework.

The 20 node-based predictors are two instance each of the local clustering coefficient (LCC), average neighbor degree (AND), shortest-path betweenness centrality (SPBC), closeness centrality (CC), degree centrality (DC), eigenvector centrality (EC), Katz centrality (KC), local number of triangles (LNT), Page rank (PR), and load centrality (LC)~\cite{newman:networks:2018,liben2007link, al2011survey, cukierski2011graph}.

\subsection*{Model-based predictors}
Model-based predictors are a broad class of prediction algorithms that rely on models of large-scale network structure to score pairs $i,j$ that are more or less likely to be missing. To make link predictions, model-based algorithms employ one of two strategies:\ likelihood or optimization. In the first case, a method estimates a parametric probability $\Pr(i\to j\,|\,\theta)$ that a node pair should be connected, given a decomposition of a network into communities, as in the stochastic block model and its variants. In the second it predicts a link as missing if it would improve its measure of community structure, as in Infomap and modularity.

We consider 11 model-based predictors for missing links, which include many state-of-the-art in community detection algorithms~\cite{ghasemian2018evaluating}, are sufficiently scalable to be applied in an evaluation as large as ours, and each of which has previously been used as a standalone link prediction algorithm. A listing of all model-based predictors is given in Table~\ref{table:abbr}, along with the corresponding literature references.

For the model-based predictors that make predictions by likelihood, we follow Ref.~\cite{ghasemian2018evaluating} to employ a ``model-specific'' score function for each method. Under this approach, a particular method first decomposes the network into a set of communities using its corresponding parametric model, and then extracts from that same parametric model a score $\Pr(i\to j\,|\,\theta)$ for each candidate pair $i,j$. See Ref.~\cite{ghasemian2018evaluating} for additional details.

\begin{table*}[h!]\addtolength{\tabcolsep}{-5pt}
\caption{Abbreviations and descriptions of 11 model-based predictors, across two types:\ \textit{likelihood} predictors (7), which score each pair $i,j$ according to a parametric model $\Pr(i\to j\,|\,\theta)$ learned by decomposing the network under a probabilistic generative model of network structure such as the stochastic block model or its variants; and, \textit{optimization} predictors (4), which score each pair $i,j$ according to whether adding them would increase a corresponding (non-probabilistic) community structure objective function, as in the Map Equation or the modularity function.}
\vspace*{-3mm}
\centering
 \begin{tabular}{|p{2.42cm} | p{8.4cm} |c|c|c|} 
 
 \hline
Abbreviation &  Description & Likelihood & Optimization & ~~Ref.~~  \\ [0.5ex] 
 \hline\hline
Q & modularity, Newman-Girvan & & $\bullet$ & \cite{newman2004finding} \\  \hline
Q-MR & modularity, Newman's multiresolution & & $\bullet$ & \cite{newman2016community}   \\  \hline
Q-MP & modularity, message passing  & & $\bullet$ &  \cite{zhang2014scalable}  \\  \hline
B-NR (SBM) & Bayesian stochastic block model, Newman and Reinert & $\bullet$ & & \cite{newman2016estimating}   \\  \hline
B-NR (DC-SBM)  & Bayesian degree-corrected stochastic block model, Newman and Reinert & $\bullet$ & & \cite{newman2016estimating}   \\  \hline
B-HKK (SBM) & Bayesian stochastic block model, Hayashi, Konishi and Kawamoto & $\bullet$ & & \cite{hayashi2016tractable} \\  \hline
cICL-HKK (SBM) & Corrected integrated classification likelihood, stochastic block model & $\bullet$ & & \cite{hayashi2016tractable} \\  \hline
Infomap & Map equation & & $\bullet$ & \cite{rosvall2008maps} \\  \hline
MDL (SBM)  & Minimum description length, stochastic block model & $\bullet$ & & \cite{peixoto2013parsimonious} \\  \hline
MDL (DC-SBM)  & Minimum description length, degree-corrected stochastic block model & $\bullet$ & & \cite{peixoto2013parsimonious} \\  \hline
S-NB & Spectral with non-backtracking matrix & $\bullet$ & & \cite{Krzakala2013} \\  \hline
\end{tabular}
\label{table:abbr}
\end{table*}

\subsection*{Embedding-based predictors}
Embedding-based predictors are derived from graph embedding techniques, which attempt to automate the feature engineering phase of learning with graphs by projecting a network's nodes into a relatively low-dimensional latent space, with the goal of locally preserving the node neighborhoods. Embedding-based predictors are thus either node coordinates in such an embedding, or measure of distance between embedded pairs. We consider a total of 150 embedding-based predictors, all derived from 2 popular graph embedding algorithms, DeepWalk (emb-DeepWalk)~\cite{perozzi2014deepwalk}---a special case of node2vec (emb-node2vec)~\cite{grover2016node2vec}---and the variational graph auto encoder (emb-vgae)~\cite{kipf2016variational}.

Using emb-DeepWalk and emb-vgae, we embed each network into a 128-dimensional and 16-dimensional space, respectively. For each pair of nodes $i,j$, we then apply a Hadamard product function to the corresponding pair of coordinates to obtain 144 link predictors as features for supervised learning~\cite{hamilton2017representation}. To these, we add 6 more predictors by applying, for each of the 2 embedding methods, a different distance or similarity function to the corresponding pair of coordinate vectors:\ an inner product, an inner product with a sigmoid function, and Euclidean distance.

\subsection*{Stacked generalization and meta-learning for link prediction}
Meta-learning or ensemble techniques are a powerful class of supervised machine learning algorithms that can learn from data how to combine individual predictors into a single, more accurate algorithm~\cite{breiman1996bagging,schapire1999brief,dietterich2000ensemble,sewell2008ensemble}. By treating the output of individual prediction algorithms as features of the input instances themselves, a supervised meta-learning algorithm can construct a correlation function that relates which individual algorithm is most accurate on which subset of inputs. Of the several approaches to meta-learning, we focus on the approach of stacked generalization or model ``stacking''~\cite{wolpert1992stacked}, and we consider two boosting approaches (see below) as a robustness check. We leave further investigation of other meta-learning algorithms for future work.

Stacking aims to minimize the generalization error of a set of component learners. In the classic setting, the two training levels can be summarized as follows. Given a dataset $\mathcal{D}=\{(y_{\ell},x_{\ell}), \ell\in\{1, ... ,L\} \}$, where $x_{\ell}$ is the feature vector of the $\ell$-th example and $y_{\ell}$ is its label, randomly split $\mathcal{D}$ into $J$ ``folds'' appropriate for $J$-fold cross validation. Each fold $j$ contributes once as a test set $\mathcal{D}^j$ and the rest contributes once as a training set $\mathcal{D}^{-j} = \mathcal{D} \smallsetminus \mathcal{D}^j$. For each base classifier $r$, where $r\in\{1,...,R\}$, called a level-$0$ generalizer, we fit it to the $j$th fold in the training set $\mathcal{D}^{-j}$ to build a model $\mathcal{M}_r^{-j}$, called a level-$0$ model. Now for each data point $x_{\ell}$ in the $j$th test set, we employ these level-$0$ models $\mathcal{M}_r^{-j}$ to predict the output $z_{r\ell}$. The new data set $\mathcal{D}_{CV} =\{(y_{\ell},z_{1\ell},...,z_{R\ell}), \ell\in\{1, ... ,L\}\}$, is now prepared for the next training level, called a level-$1$ generalizer. In the second training phase, an algorithm learns a new model from this data, denoted as $\tilde{\mathcal{M}}$. Now, we again train the base classifiers using the whole data $\mathcal{D}$, noted as $\mathcal{M}_r$, we complete the training phase and the models are ready to classify a new data point $x$. The new data point will first be fed into the trained base classifiers $\mathcal{M}_r$ and then the output of these level-$0$ models will construct the input for the next level model $\tilde{\mathcal{M}}$.

In the network setting of link prediction, the classifiers (predictors) in the first level are all unsupervised, and therefore, we alter the stacked generalization algorithm as follows to account for this difference and to adapt it to a network setting. For a given network $G=(V,E)$, we sample the edges uniformly and construct the observed network $G'=(V,E')$, where $| E |=\alpha | E |$ ($\alpha=0.8$ in our experiments). Here, we use only the uniform edge-removal model and leave the analysis of any non-uniform edge removal model for future work. The removed edges $E\setminus E'$ are considered as held-out data in the link prediction task. Then, in order to train a model, we remove $1-\alpha'$ ($\alpha'=0.8$ in our experiments) of the edges as our positive examples and take all non-edges in the observed network $G'$ as negative examples. Although this procedure makes the negative samples noisy, since the networks are sparse, it introduces a negligible error in the learned model, and should not significantly effect the model's performance. In our setting, the unsupervised classifiers in the first level are our level-$0$ predictors, and we use the scores coming from these link prediction techniques as our meta features. The second training phase is conducted through supervised learning with $5$-fold cross validation on the training set. We use a standard supervised random forest algorithm for the meta-learning step, and assess the learning process on 3 within-family models (topol.\ only, model-based only, and embed.\ only) and on 4 across-family models  (all families, and each of topol.\ \& model, topol.\ \& embed, and model \& embed.), for a total of 7 stacked models.

\smallskip \noindent \textit{Model selection.} In order to choose the best parameters of the model using 5-fold cross validation, we can choose the parameters of the model through optimizing the AUC performance or the F measure. In the main text all figures and tables show results for a standard random forest with the parameters chosen through F measure optimization and the results are reported on the test set. Results for, instead, optimizing using the AUC are given in Table~\ref{tab:rf_AUC} which can be compared with Table~\ref{tab:satexp} in the main text.

\smallskip \noindent \textit{Alternative meta-learning algorithms.} In addition to a standard random forest, we also evaluate two methods of boosting, XGBoost~\cite{chen2016xgboost} and AdaBoost~\cite{freund1997decision}, for learning a single algorithm over the individual predictors. The results from these meta-learning algorithms are provided in Tables~\ref{tab:xgb_f}-\ref{tab:ada_AUC} for different choices of model selection through AUC or F measure.

\section{Tests on synthetic data}
\label{sec:synthetic}
We evaluate individual predictors and their stacked generalization on a set of synthetic networks with known structure that varies along three dimensions: (i)~the degree distribution's variability, being low (Poisson), medium (Weibull), or high (power law); (ii)~the number of ``communities'' or modules $ k \in  \{1, 2, 4, 16, 32\}$; and (iii)~the fuzziness of the corresponding community boundaries $\epsilon=\mout/\miin$, the fraction of a node's edges that connect outside its community, being low, medium, or high. These synthetic networks thus range from homogeneous to heterogeneous random graphs (degree distribution), from no modules to many modules ($k$), and from weakly to strongly modular structure ($\epsilon$).

We generate these networks using the degree-corrected stochastic block model (DC-SBM)~\cite{karrer2011stochastic}, which allows us to systematically control each of these parameters to generate a synthetic network. Moreover, because both the data generating process and the missing function $f$ (here, uniform at random) are known, we may exactly calculate the theoretical upper limit that any link prediction algorithm could achieve for a given parameterization of the generative process. This upper bound provides an unambiguous reference point for how optimal any particular link prediction algorithm is, and the three structural dimensions of the synthetic networks allow us to extract some general insights as to what properties increase or decrease the predictability of missing links.

In this section, we first describe the generative processes, and then detail the calculations for optimal predictions. For completeness, we first specify the mathematical forms of the Weibull and power-law degree distributions used in some settings. The Poisson distribution is fully specified by the choice of the mean degree parameter $c$.

The Weibull distribution can be written as 
\begin{align}
\label{eq:WeibD}
f(r) = c r ^{\beta-1}e^{-\lambda r ^{\beta}} \enspace ,
\end{align} 
where the constant $c$ is the corresponding normalization constant when $r$ is the degree of a node, and the parameters $\lambda,\beta$ specify the shape of the distribution. When $\beta<1$, this distribution decays more slowly than simple exponential, meaning it exhibits greater variance, but not as much variance as can a power-law distribution. See Table~\ref{tab:par_set} for the particular values used in our synthetic data.

The power-law distribution can be written as 
\begin{align}
f(r) = c r ^{-\gamma} \enspace ,
\end{align}
where, again, $c$ is the corresponding normalization constant when $r$ is the degree of a node, and $\gamma$ is the ``scaling'' exponent that governs the shape of the distribution. When $\gamma\in(2,3)$, the mean is finite but the variance is infinite. See Table~\ref{tab:par_set} for the particular values used in our synthetic data.

\begin{table*}[t!]
\centering
\caption{Parameters used to generate the synthetic networks, via the DC-SBM structured random graph model, used to evaluate the link prediction methods studied here. Redundant information (derivable from the other parameters) is listed parenthetically, for convenience. See Section~\ref{sec:synthetic}.}
\begin{tabular}{l|l|l|l}
\hline \hline
Region & Model &  Number of modules $k$ & Parameters \\ 
\hline\hline
low $\epsilon$&Poisson&1&  $n=505$, $p=0.008$ \\ \hline    
low $\epsilon$&Poisson&2&  $n=512, p_{\text{in}}=0.03, p_{\text{out}}=0.0003~~(\epsilon=0.009)$  \\ \hline    
low $\epsilon$&Poisson&4&  $n=512, p_{\text{in}}=0.06, p_{\text{out}}=0.0003~~(\epsilon=0.015)$ \\ \hline    
low $\epsilon$&Poisson&16& $n=512, p_{\text{in}}=0.25, p_{\text{out}}=0.0003~~(\epsilon=0.015)$ \\ \hline    
low $\epsilon$&Poisson&32& $n=512, p_{\text{in}}=0.49, p_{\text{out}}=0.0003~~(\epsilon=0.019)$ \\ \hline    
low $\epsilon$&Weibull&1& $n=497$, $\lambda=1$, $\beta=0.5$, $\omega=2350$ \\ \hline    
low $\epsilon$&Weibull&2&$n=520$, $\lambda=1$, $\beta=0.4$, $\epsilon=0.002$ \\ \hline    
low $\epsilon$&Weibull&4&$n=604$, $\lambda=1$, $\beta=0.4$, $\epsilon=0.002$ \\ \hline    
low $\epsilon$&Weibull&16&$n=773$, $\lambda=1$, $\beta=0.4$, $\epsilon=0.04$ \\ \hline    
low $\epsilon$&Weibull&32&$n=939$, $\lambda=1$, $\beta=0.15$, $\epsilon=0.0005$ \\ \hline    
low $\epsilon$&power law&1& $n=507$, $\beta=1.6$, $\omega=5436$ \\ \hline    
low $\epsilon$&power law&2& $n=511$, $\beta=1.7$, $\epsilon=0.0003$ \\ \hline    
low $\epsilon$&power law&4& $n=511$, $\beta=1.8$, $\epsilon=0.002$ \\ \hline    
low $\epsilon$&power law&16& $n=983$, $\beta=1.6$, $\epsilon=0.0015$ \\ \hline    
low $\epsilon$&power law&32& $n=1029$, $\beta=1.41$, $\epsilon=0.0015$ \\ \hline    
moderate $\epsilon$&Poisson&1& $n=511$, $p=0.016$ \\ \hline    
moderate $\epsilon$&Poisson&2& $n=512$, $p_{\text{in}}=0.03, p_{\text{out}}=0.005~~(\epsilon=0.20)$ \\ \hline    
moderate $\epsilon$&Poisson&4& $n=512$, $p_{\text{in}}=0.04, p_{\text{out}}=0.006~~(\epsilon=0.39)$  \\ \hline    
moderate $\epsilon$&Poisson&16&$n=512$, $p_{\text{in}}=0.16, p_{\text{out}}=0.006~~(\epsilon=0.6)$  \\ \hline    
moderate $\epsilon$&Poisson&32&$n=511$, $p_{\text{in}}=0.31, p_{\text{out}}=0.006~~(\epsilon=0.62)$  \\ \hline  
moderate $\epsilon$&Weibull&1& $n=510$, $\lambda=1$, $\beta=0.7$, $\omega=1424$  \\ \hline    
moderate $\epsilon$&Weibull&2& $n=501$, $\lambda=1$, $\beta=0.4$, $\epsilon=0.06$ \\ \hline    
moderate $\epsilon$&Weibull&4&$n=593$, $\lambda=1$, $\beta=0.4$, $\epsilon=0.08$  \\ \hline    
moderate $\epsilon$&Weibull&16&$n=589$, $\lambda=1$, $\beta=0.4$, $\epsilon=0.2$  \\ \hline    
moderate $\epsilon$&Weibull&32&$n=640$, $\lambda=1$, $\beta=0.22$, $\epsilon=0.05$  \\ \hline   
moderate $\epsilon$&power law&1& $n=545$, $\beta=1.9$, $\omega=1428$ \\ \hline     
moderate $\epsilon$&power law&2& $n=506$, $\beta=1.7$, $\epsilon=0.05$ \\ \hline    
moderate $\epsilon$&power law&4& $n=540$, $\beta=1.8$, $\epsilon=0.05$ \\ \hline    
moderate $\epsilon$&power law&16& $n=655$, $\beta=1.7$, $\epsilon=0.01$ \\ \hline    
moderate $\epsilon$&power law&32& $n=702$, $\beta=1.41$, $\epsilon=0.01$ \\ \hline 
high $\epsilon$&Poisson&1& $n=512$, $p=0.03$  \\ \hline       
high $\epsilon$&Poisson&2&$n=512$, $p_{\text{in}}=0.025, p_{\text{out}}=0.006~~(\epsilon=0.25)$  \\ \hline    
high $\epsilon$&Poisson&4&$n=512$, $p_{\text{in}}=0.04, p_{\text{out}}=0.007~~(\epsilon=0.48)$  \\ \hline    
high $\epsilon$&Poisson&16&$n=512$, $p_{\text{in}}=0.14, p_{\text{out}}=0.007~~(\epsilon=0.75)$  \\ \hline    
high $\epsilon$&Poisson&32&$n=512$, $p_{\text{in}}=0.27, p_{\text{out}}=0.007~~(\epsilon=0.93)$  \\ \hline    
high $\epsilon$&Weibull&1& $n=489$, $\lambda=1$, $\beta=0.9$, $\omega=1216$ \\ \hline    
high $\epsilon$&Weibull&2& $n=506$, $\lambda=1$, $\beta=0.4$, $\epsilon=0.2$ \\ \hline    
high $\epsilon$&Weibull&4& $n=590$, $\lambda=1$, $\beta=0.4$, $\epsilon=0.32$ \\ \hline    
high $\epsilon$&Weibull&16& $n=600$, $\lambda=1$, $\beta=0.4$, $\epsilon=0.5$  \\ \hline    
high $\epsilon$&Weibull&32& $n=631$, $\lambda=1$, $\beta=0.22$, $\epsilon=0.13$  \\ \hline  
high $\epsilon$&power law&1& $n=514$, $\beta=2.2$, $\omega=1722$ \\ \hline      
high $\epsilon$&power law&2& $n=536$, $\beta=1.7$, $\epsilon=0.08$ \\ \hline    
high $\epsilon$&power law&4& $n=526$, $\beta=1.8$, $\epsilon=0.14$ \\ \hline    
high $\epsilon$&power law&16& $n=626$, $\beta=1.7$, $\epsilon=0.1$ \\ \hline    
high $\epsilon$&power law&32& $n=673$, $\beta=1.5$, $\epsilon=0.05$ \\ \hline    

	\end{tabular}
	\label{tab:par_set}  
\end{table*} 

\subsection*{Generating synthetic networks}
\label{subsec:GP}
Although each type of synthetic network can be generated under the DC-SBM model, for some choices of the number of communities $k$ and degree distribution, the generative process simplifies greatly. Below, we describe the generation procedures for the synthetic networks according to the simplest generative model available for a given choice of parameters, which is noted in the subsection heading.

\medskip \noindent \textit{Generating ER networks ($k=1$, Poisson)}
\begin{itemize}
\itemsep-0.1pt
\item choose number of nodes $n$, and average degree $c$, or the interaction probability $p=c/(n-1)$,
\item connect each pair of nodes independently with probability $p$.
\end{itemize}

\medskip \noindent \textit{Generating DC-ER networks ($k=1$, Weibull or power law)}
\begin{itemize}
\itemsep-0.1pt
\item choose number of nodes $n$, and average degree $c$,
\item compute the parameters of degree distribution for the average degree $c$,
\item generate a degree sequence with length $n$ with the computed parameters in the previous step,
\item compute the number of edges for the network as $m=\dfrac{1}{2}\sum_i d_i$,
\item make a multi-edge between each pair of nodes $i,j$ independently with the Poisson probability with rate $\lambda = \dfrac{d_i}{d_{g_i}}\dfrac{d_j}{d_{g_j}} \omega$, where $\omega = 2m$. \\
We then convert this multigraph into a simple (unweighted) network by collapsing multi-edges. Because these networks are parameterized to be sparse, this operation does not substantially alter the network's structure, as only a small fraction of all edges are multi-edges.
\end{itemize}

\medskip \noindent \textit{Generating SBM networks ($k>1$, Poisson)}
\begin{itemize}
\itemsep-0.1pt
\item choose number of nodes $n$, number of clusters $k$, average degree $c$, and $\tilde{\epsilon}$, the ratio of number of edges connected to a node outside and inside its cluster, i.e., $\tilde{\epsilon} = \dfrac{\pout(n/k)}{\pin(n/k)}=\dfrac{\pout}{\pin}$; by choosing $c$ and $\tilde{\epsilon}$, the mixing probabilities can then be computed as $\pin = \dfrac{c}{(n/k)(1+\tilde{\epsilon}(k-1))}$ and $\pout = \tilde{\epsilon} \, \pin$,
\item generate the type of the nodes independently with prior probabilities $q_r$ for $r=\{1,...,k\}$,
\item connect each pair of nodes $i,j$ independently with probability $p_{g_i g_j}$, where 
$$p_{g_i,g_j} = \begin{cases} \pin& \mbox{if } g_i=g_j \, \\ 
\pout & \mbox{if } g_i \neq g_j \end{cases}\, . $$
\end{itemize}

\medskip \noindent \textit{Generating DC-SBM networks ($k>1$, Weibull or power law)}
\begin{itemize}
\itemsep-0.1pt
\item choose number of nodes $n$, average degree $c$, and $\epsilon$, the ratio of number of edges between the clusters and inside the clusters, i.e., $\epsilon = \mout/\miin$, where $\miin$ is the number of edges inside the clusters, and $\mout$ is the number of edges between the clusters,
\item generate the type of nodes independently with prior probabilities $q_r$ for $r=\{1,...,k\}$,
\item compute the parameters of degree distribution for average degree $c$,
\item generate a degree sequence with length $n$ with the computed parameters in previous step, and compute the aggregate degrees for each cluster noted as $d_r = \sum_{i:g_i = r} d_i$,
\item compute the total number of edges for the network as $m=\dfrac{1}{2}\sum_i d_i$,
\item using $\epsilon$, compute the number of edges inside and outside the clusters, denoted as $\miin$, and $\mout$, as $\miin=m/(1+\epsilon)$ and $\mout = \epsilon \, \miin$,
\item because we do not assume heterogeneity for the size and volume of clusters in the generating process (node types are randomized uniformly and edges are created uniformly inside and between clusters), then we may approximate the number of edges inside each cluster $r$ as $\minr = \miin/k$, the number of edges between cluster $r$ and any other cluster as $\mout^{(r)}=\dfrac{\mout}{k/2}$, and the number of edges between each pair of clusters $r$ and $s$ as $\moutrs=\mout \left/ \binom{k}{2}\right. $,
\item make a multi-edge between each pair of nodes $i,j$ with types $r,s$, independently with the Poisson probability with rate $\lambda_{r,s}(d_i,d_j) = \dfrac{d_i}{d_{r}}\dfrac{d_j}{d_{s}} \omega_{r,s}$, where 
$$\omega_{r,s} = \begin{cases} 2\minr& \mbox{if } r=s \, \\ 
\moutrs& \mbox{if } r \neq s \end{cases}\, . $$ \\
We then convert this multigraph into a simple (unweighted) network by collapsing multi-edges. Because these networks are parameterized to be sparse, this operation does not substantially alter the network's structure, as only a small fraction of all edges are multi-edges.
\end{itemize}
It is worthwhile to mention that $\epsilon$ in \mbox{DC-SBM} is related to $\tilde{\epsilon}$ in SBM as $\epsilon = \mout/\miin = (k-1)\pout/\pin=(k-1)\tilde{\epsilon}$. Therefore, for the results, we used $\epsilon=\mout/\miin$ for both SBM and \mbox{DC-SBM}.

\subsection*{Optimal link prediction accuracy on a synthetic network}
\label{secB:OPLPSN}
To calculate an upper bound on link prediction accuracy that any algorithm could achieve in one of our synthetic networks, we exploit the mathematical equivalence of the Area Under the ROC Curve (AUC) and the binary classification probability that a prediction algorithm $\mathcal{A}$ assigns a higher score to a missing link (true positive) than to a non-edge (true negative):
\begin{align}
\label{eq:gen_AUC}
\textrm{AUC} = \Pr({\rm tes} > {\rm tnes}) \enspace ,
\end{align}
where $\rm tes$ and $\rm tnes$ denote the scores assigned to a missing edge (te; true positive) and to a non-edge (tne; true negative). 
To derive the optimal AUC for any possible link prediction algorithm, it suffices to calculate this probability under a given parametric generative model $\mathcal{M}(\theta)$ and missingness function $f$.

\medskip \noindent \textit{Assumptions and definitions.}
In the calculations that follow, we treat separately the three generative process subcases of the 
\begin{tikzpicture}
    \draw (0,0) -- (0:8.7cm);
\end{tikzpicture}

\noindent
DC-SBM described above, and we define $n=|V|$, $m=|E|$. If two edges assigned the same score by the generative model, we assume that such ties are broken uniformly at random.

For these calculations, we also assume that algorithm $\mathcal{A}$ has access to the planted partition assignment $\mathcal{P}$ of the $k$ clusters used to generate the edges. In practice, this assumption implies that our upper bound may be unachievable in cases where the detectability of $\mathcal{P}$ is either computational hard or information-theoretically impossible (see Ref.~\cite{decelle2011asymptotic}), e.g., when community boundaries are fuzzy (high $\epsilon$).

Given this partition, we define $n_i$, $m_i$, and $\tilde{m}_i$ to be the number of nodes, number of edges, and number of non-edges, respectively, within community $i$. And, we define $m_{ij}$ and $\tilde{m}_{ij}$ to be the number of edges and non-edges, respectively, that span communities $i$ and $j$.

Finally, when we estimate \eqref{eq:gen_AUC} via Monte Carlo sampling, we select 100,000 uniformly random te (true positive) and tne (true negative) pairs.

\medskip \noindent \textit{Optimal AUC for ER.}
The AUC for an Erd\H{o}s-R\'enyi random graph is
\begin{align}
\textrm{AUC} = \Pr({\rm tes} > {\rm tnes}) = 1/2 \enspace .
\end{align}
In words:\ because the generative model assigns the same score $p=c/(n-1)$ to every edge and every non-edge, and because ties are broken at random, the maximum AUC can be no better than chance.

\medskip \noindent \textit{Optimal AUC for DC-ER.}
As in the ER case, this random graph has $k=1$ communities, but unlike the ER case, the degree distribution here is heterogeneous (Weibull or power law). We calculate the maximum AUC for any algorithm $\mathcal{A}$ on this synthetic network via \eqref{eq:OAUC_DCER} (below), which we estimate numerically via Monte Carlo sampling on
\eqref{eq:gen_AUC}.

\begin{strip}
\begin{align}\label{eq:OAUC_DCER}
\textrm{AUC} \stackrel{}{=}&  \Pr({\rm tes} > {\rm tnes})  \nonumber\\ \stackrel{}{=}&  \sum_{u_1. v_1, u_2, v_2} p(u_1 v_1 > u_2 v_2, d_{i_1} = u_1, d_{j_1} = v_1, d_{i_2} = u_2, d_{j_2}=v_2 \mid (i_1,j_1) \in E , (i_2,j_2) \notin E)\nonumber \\
\stackrel{}{=}&  \sum_{u_1, v_1, u_2, v_2} \bbone(u_1 v_1 > u_2 v_2) p(d_{i_1} = u_1, d_{j_1} = v_1, d_{i_2} = u_2, d_{j_2}=v_2 \mid (i_1,j_1) \in E , (i_2,j_2) \notin E)\nonumber \\
\stackrel{\left(\mbox{Bayes Thm.}\right)}{=}& \sum_{u_1, v_1, u_2, v_2} \bbone(u_1 v_1 > u_2 v_2) \dfrac{p((i_1,j_1) \in E , (i_2,j_2) \notin E \mid d_{i_1} = u_1, d_{j_1} = v_1, d_{i_2} = u_2, d_{j_2}=v_2)}{p((i_1,j_1) \in E , (i_2,j_2) \notin E)}
\nonumber\\&\times p(d_{i_1} = u_1)p(d_{j_1} = v_1)p(d_{i_2} = u_2)p(d_{j_2} = v_2) \enspace .
\end{align}
\end{strip}

\medskip \noindent \textit{Optimal AUC for SBM.}
In the general case ($k>1$) of the stochastic block model (SBM), the te and tne probabilities under the generative model depend on the mixing matrix of edge densities between and within \hspace{0.4mm} communities. When \hspace{0.5mm} these densities \hspace{0.5mm} are set such  \hspace{0.2mm} that the  \vspace{3mm}

\noindent
\begin{tikzpicture}
    \draw (0,0) -- (0:8.7cm);
\end{tikzpicture} 

\noindent
planted partition $\mathcal{P}$ is easily recoverable by a community detection algorithm (a range of parameters called the ``deep detectable regime'' (DDR)~\cite{decelle2011asymptotic}, where $\epsilon \to 0$), \eqref{eq:gen_AUC} can be rewritten as \eqref{eq:OAUC_SBM}:

\begin{strip}
\begin{align}\label{eq:OAUC_SBM}
\textrm{AUC} = & \Pr({\rm tes} > {\rm tnes}) \nonumber\\
=&\Pr({\rm tes} > {\rm tnes} | \mbox{both inside})\Pr(\mbox{both inside}) \times \mbox{number of possibilities}  \nonumber\\
&+ \Pr({\rm tes} > {\rm tnes} | \mbox{both outside})\Pr(\mbox{both outside}) \times \mbox{number of possibilities} \nonumber \\
&+ \Pr({\rm tes} > {\rm tnes} |  \mbox{te inside}, \mbox{tne outside})\Pr(\mbox{te inside}, \mbox{tne outside}) \times \mbox{number of possibilities} \nonumber\\
&+ \Pr({\rm tes} > {\rm tnes} | \mbox{te outside}, \mbox{tne inside})\Pr(\mbox{te outside}, \mbox{tne inside}) \times \mbox{number of possibilities} \enspace . 
\end{align}
\end{strip}

\begin{strip}

\vspace{2cm}

The four terms of \eqref{eq:OAUC_SBM} can then be computed as follows, where $\alpha$ is the sampling rate of observed edges:

\begin{itemize}

	\begin{item} First term:
	
	\begin{align}
	 \Pr&({\rm tes} > {\rm tnes} \,|\, \mbox{both inside})\Pr(\mbox{both inside}) \nonumber\\
	 &= \frac{m_i \alpha}{\sum_i m_i \alpha + \sum_{i\neq j} m_{ij} \alpha}\times 
	\frac{\tilde{m}_i}{\sum_i \tilde{m}_i + \sum_{i\neq j}\tilde{m}_{ij}} \nonumber\\
	 &= \frac{\binom{n_i}{2}\pin}{\sum_i \binom{n_i}{2}\pin + \sum_{i\neq j}n_in_j\pout}\times\frac{\binom{n_i}{2}(1-\pin)}{\binom{n_i}{2}(1-\pin)+n_in_j(1-\pout)} \nonumber\\
	 &= \frac{1}{2}\left(\frac{\pin}{k^3(\pin+(k-1)\pout)}\right) = \frac{\cin}{2k^4c} \enspace , \footnotemark
	\end{align}
	
\footnotetext{Note that in the last line we assume equally-sized homogeneous clusters i.e. $n_i=n/k$.}
Finally, because the $\mbox{number of possibilities}$ is $k^2$, the first term simplifies as $ \frac{1}{2}\left(\frac{\pin}{k^3(\pin+(k-1)\pout)}\right) \times k^2 \approx 1/2k$.
	\end{item}
	
	\begin{item} Second term: 
	
	\begin{align}
	 \Pr&({\rm tes} > {\rm tnes} \,|\, \mbox{both outside})\Pr(\mbox{both outside}) \nonumber\\
	 &= \frac{m_{ij} \alpha}{\sum_i m_i \alpha + \sum_{i\neq j} m_{ij} \alpha}\times 
	\frac{\tilde{m}_{ij}}{\sum_i \tilde{m}_i + \sum_{i\neq j}\tilde{m}_{ij}} \nonumber\\
	 &= \frac{n_i n_j\pout}{\sum_i \binom{n_i}{2}\pin + \sum_{i\neq j}n_in_j\pout}\times\frac{n_i n_j(1-\pout)}{\binom{n_i}{2}(1-\pin)+n_in_j(1-\pout)} \nonumber\\
	 &=2\left(\frac{\pout}{k^3(\pin+(k-1)\pout)}\right) = \frac{2\cout}{k^4c}.
	\end{align}
	 
Finally, because the $\mbox{number of possibilities}$ is $\binom{k}{2}^2\approx\frac{k^4}{4}$ the second term simplifies as $\frac{2\pout}{k^3(pin+(k-1)\pout)} \times \frac{k^4}{4} = \frac{k\pout}{(\pin+(k-1)\pout)}\approx 0$.
	\end{item}
	
	\begin{item} Third term: 
	
	\begin{align}
	  \Pr&({\rm tes} > {\rm tnes} \,|\, \mbox{tes inside}, \mbox{tnes outside})\Pr(\mbox{tes inside}, \mbox{tnes outside}) \nonumber\\
	 &= \frac{m_{i} \alpha}{\sum_i m_i \alpha + \sum_{i\neq j} m_{ij} \alpha}\times 
	\frac{\tilde{m}_{ij}}{\sum_i \tilde{m}_i + \sum_{i\neq j}\tilde{m}_{ij}} \nonumber\\
	 &= \frac{\binom{n_i}{2}\pin}{\sum_i \binom{n_i}{2}\pin + \sum_{i\neq j}n_in_j\pout}\times\frac{n_i n_j(1-\pout)}{\binom{n_i}{2}(1-\pin)+n_in_j(1-\pout)} \nonumber\\
	 &=2\left(\frac{\pin}{k^3(\pin+(k-1)\pout)}\right) = \frac{2\cin}{k^4c}.
	\end{align}
Finally, because the $\mbox{number of possibilities}$ is $k\binom{k}{2}=k^2(k-1)/2$ the third term simplifies as $\frac{2 \pin}{k^3(\pin+(k-1)\pout)} \times \frac{k^2(k-1)}{2} = (k-1)/k$.
	\end{item}
	
	\begin{item} Last term:
	
	\begin{align}
	\Pr({\rm tes} > {\rm tnes} \,|\, \mbox{tes outside}, \mbox{tnes inside})\Pr(\mbox{tes outside}, \mbox{tnes inside}) = 0
	\end{align}

Finally, when the SBM parameters are such that the model is in the deep detectable regime (DDR), 
the fourth term is zero, because the assigned scores to the outer edges are smaller than the assigned scores to inner edges, under the assumption that the algorithm $\mathcal{A}$ can recover the planted partition (which occurs with probability $1$ in DDR).
	\end{item}
	\end{itemize}

\end{strip}

	Given the above simplifications, we arrive at the final expression to compute the optimal AUC for the SBM in the DDR:
	\begin{align}
	\textrm{AUC} &= \Pr( {\rm tes} > {\rm tnes} ) \nonumber\\
	&=  \frac{1}{2k} + \frac{k-1}{k} \nonumber\\
	&= \frac{2k-1}{2k}.
	\end{align}
	For example, the upper bounds on link predictability under this model for $k=\{2,4,8,16,32\}$ are $\textrm{AUC}=\{0.75, 0.875, 0.94, 0.97, 0.98\}$, respectively. Because these values are computed in the deep detectable regime, they are accurate only when $\epsilon$ is low (sharp community boundaries, or $\mathcal{P}$ is known or recoverable).
	
	For any value of $\epsilon$, we may numerically calculate the upper bound on AUC using Monte Carlo sampling via the \eqref{eq:gen_AUC}, applied to the generated networks. The corresponding values represent conservative upper bounds on the maximum AUC
\begin{tikzpicture}
    \draw (0,0) -- (0:8.7cm);
\end{tikzpicture} 

\noindent
because under Monte Carlo because we assume that $\mathcal{P}$ is known.
 
	In practice, a community detection algorithm would need to infer that from the observed data, and this event is not guaranteed when $\epsilon$ is higher~\cite{decelle2011asymptotic}, due to a phase transition in the detectability (recoverability) of the planted partition structure that maximizes the predictability of missing links. We suggest that the gap observed in Fig.~\ref{fig:exp_PL_LN} between this conservative upper bound and accuracy of the best stacked models in the high-$\epsilon$ settings can be attributed to this difference. That is, the stacked models are closer to the true upper bound than our calculations suggest.

\medskip \noindent \textit{Optimal AUC for DC-SBM.}
In the general case ($k>1$) of the degree-corrected SBM, the te and tne probabilities under the generative model depend on the specified degree distribution (Weibull or power law) and the mixing matrix of edge densities between and within communities.

In this setting, \eqref{eq:gen_AUC} can be rewritten as \eqref{eq:OAUC_SBM} when $\epsilon \to 0$; however to compute each term we must also condition on the degrees of the nodes. Following the same logic as in the SBM analysis, we compute each term separately as follows.

\begin{strip}
\begin{itemize}

\item First term:
 
\begin{align}
\Pr&({\rm tes} > {\rm tnes} \,|\, \mbox{both inside}) \nonumber\\ 
=& \sum_{u_1, v_1, u_2, v_2} \bbone(u_1 v_1 > u_2 v_2)
\dfrac{p((i_1,j_1) \in E , (i_2,j_2) \notin E \mid d_{i_{1,2}} = u_{1,2}, d_{j_{1,2}} = v_{1,2})}{p((i_1,j_1) \in E , (i_2,j_2) \notin E)}
\nonumber\\&\times p(d_{i_1} = u_1)p(d_{j_1} = v_1)p(d_{i_2} = u_2)p(d_{j_2} = v_2)\enspace ,
\label{eq:1AUC_DCSBM}
\end{align}
 
where $d_{i_{1,2}} = u_{1,2}$ means $d_{i_1}=u_1$ and $d_{i_2}=u_2$.  

\item Second term:
 
\begin{align}
\Pr&({\rm tes} > {\rm tnes} \,|\, \mbox{both outside}) \nonumber\\ 
=& \sum_{u_1, v_1, u_2, v_2} \bbone(u_1 v_1 > u_2 v_2)
\dfrac{p((i_1,j_1) \in E , (i_2,j_2) \notin E \mid d_{i_{1,2}} = u_{1,2}, d_{j_{1,2}} = v_{1,2})}{p((i_1,j_1) \in E , (i_2,j_2) \notin E)}
\nonumber\\&\times p(d_{i_1} = u_1)p(d_{j_1} = v_1)p(d_{i_2} = u_2)p(d_{j_2} = v_2)\enspace .
\label{eq:2AUC_DCSBM}
\end{align}

\item Third term:
 
\begin{align}
\Pr&({\rm tes} > {\rm tnes} \,|\, \mbox{te inside}, \mbox{tne outside}) \nonumber\\ 
=& \sum_{u_1, v_1, u_2, v_2} \bbone(u_1 v_1 m_{rr}> u_2 v_2 m_{rs})
\dfrac{p((i_1,j_1) \in E , (i_2,j_2) \notin E \mid d_{i_{1,2}} = u_{1,2}, d_{j_{1,2}} = v_{1,2})}{p((i_1,j_1) \in E , (i_2,j_2) \notin E)}
\nonumber\\&\times p(d_{i_1} = u_1)p(d_{j_1} = v_1)p(d_{i_2} = u_2)p(d_{j_2} = v_2)\enspace .
\label{eq:3AUC_DCSBM}
\end{align}

\item Fourth term:
 
\begin{align}
\Pr&({\rm tes} > {\rm tnes} \,|\, \mbox{te outside}, \mbox{tne inside}) \nonumber\\ 
=& \sum_{u_1, v_1, u_2, v_2} \bbone(u_1 v_1 m_{rs}> u_2 v_2 m_{rr})
\dfrac{p((i_1,j_1) \in E , (i_2,j_2) \notin E \mid d_{i_{1,2}} = u_{1,2}, d_{j_{1,2}} = v_{1,2})}{p((i_1,j_1) \in E , (i_2,j_2) \notin E)}
\nonumber\\&\times p(d_{i_1} = u_1)p(d_{j_1} = v_1)p(d_{i_2} = u_2)p(d_{j_2} = v_2)\enspace .
\label{eq:4AUC_DCSBM}
\end{align}
 
\end{itemize}
\end{strip}
We compute these terms numerically using Monte Carlo samples of the generated networks to calculate \eqref{eq:gen_AUC}.

\section{Empirical corpus for link prediction evaluations}
\begin{figure*}[h!]
\centering
\begin{tabular}{cc}
\includegraphics[width=0.6\textwidth]{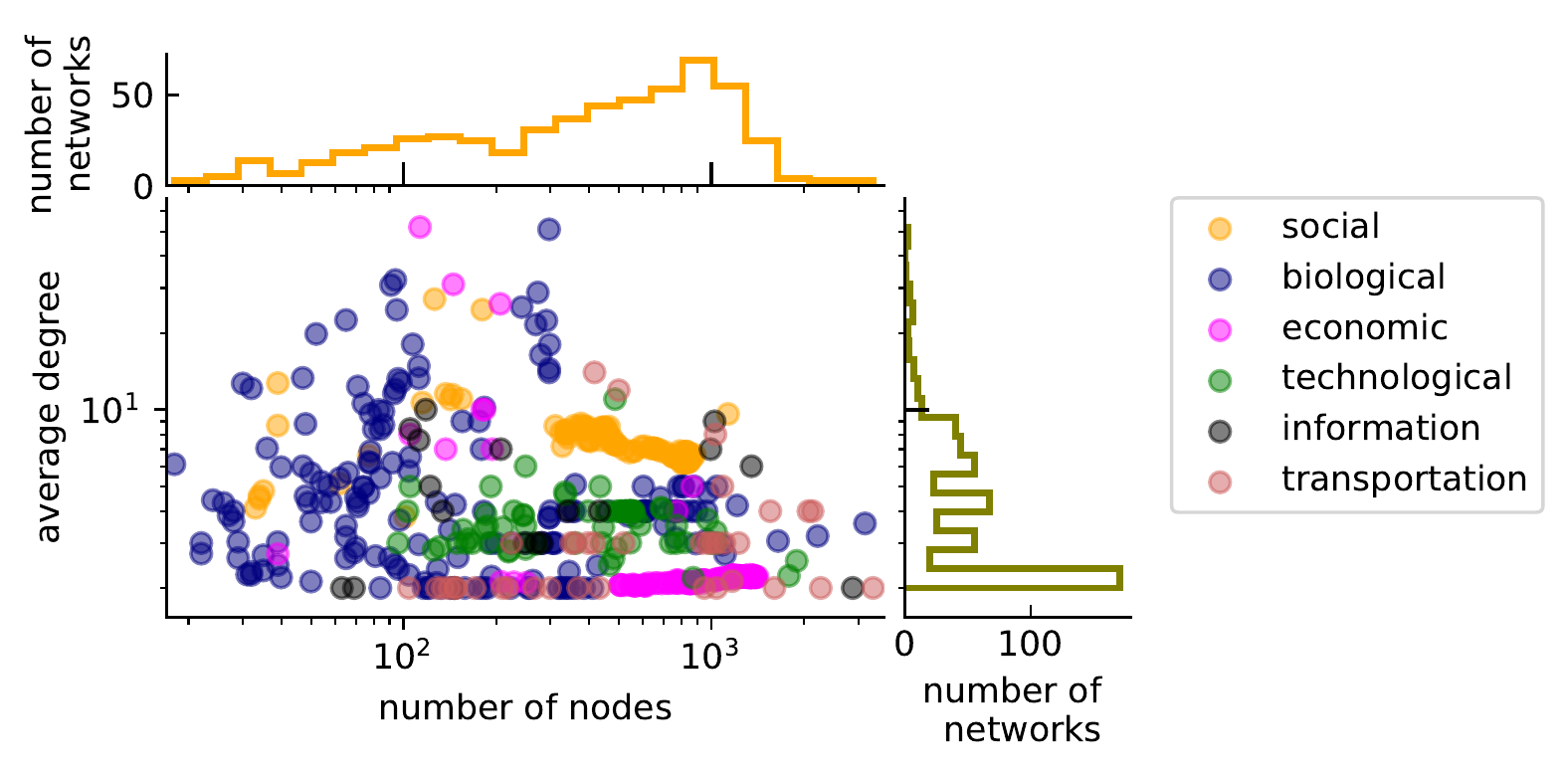}
\end{tabular}
\caption{Average degree versus number of nodes for a subset of CommunityFitNet corpus~\cite{ghasemian2018evaluating} consisting of 548 real-world networks drawn from the Index of Complex Networks (ICON)~\cite{clauset2016ICON}, including social, biological, economic, technological, information, and transportation graphs.}
\label{fig:AD}
\end{figure*}

To evaluate and compare the different link prediction algorithms in a practical setting, we have selected 548 networks%
\footnote{Available at \url{https://github.com/Aghasemian/OptimalLinkPrediction}}
from the ``CommunityFitNet corpus,'' a novel data set%
\footnote{Available at \url{https://github.com/AGhasemian/CommunityFitNet}}
containing 572 real-world 
\noindent
\begin{tikzpicture}
    \draw (0,0) -- (0:8.7cm);
\end{tikzpicture} 
networks drawn from the Index of Complex Networks (ICON)~\cite{clauset2016ICON}. 
This corpus spans a variety of network sizes and structures, with 22\% social, 21\% economic, 34\% biological, 12\% technological, 4\% information, and 7\% transportation graphs (Fig.~\ref{fig:AD}). 

\section{Evaluation of the link prediction algorithms}
\label{sec:SI_ELPA}
In practical settings, the true missingness function $f$ may not be known, and $f$ is likely to vary with the scientific domain, the manner in which the network data is collected, and the scientific question of interest. Here, we do not consider all possible functions $f$, and instead analyze an $f$
that samples edges uniformly at random from $E$ so that each edge $(i,j)\in E$ is observed with probability $\alpha$.%
\footnote{Unless otherwise specified, results reflect a choice of $\alpha=0.8$, i.e., 20\% of edges are unobserved (holdout set); other values produce qualitatively similar results.}
This choice presents a hard test for link prediction algorithms, as $f$ is independent of both observed edges and metadata. Other models of $f$, e.g., in which missingness correlates with edge or node characteristics, may better capture particular scientific settings and are left for future application-specific work. Our results thus provide a general, application-agnostic assessment of link predictability and method performance.

Most of the predictors we consider are predictive only under a supervised learning approach, in which we learn a model of how these node-pair features correlate with edge missingness. This supervised approach to link prediction 
poses one specific technical challenge. 
Under supervised learning, we train a method using 5-fold cross validation by choosing as positive examples a subset of edges $E'' \subset E'$ according to the same (uniformly random) missingness model $f$.
But applying this missingness function can only create positive training examples (missing links), while supervised learning also needs negative examples (non-links). Other approaches to supervised link prediction have made specific assumptions to mitigate this issue. For example, in a temporal network, an algorithm can be trained using the links and non-links observed during an earlier training time frame~\cite{al2006link, ahmed2016supervised}, or if some missing links are known \textit{a priori}, they may be used as training examples~\cite{lichtenwalter2010new}. 
However, such approaches require information, e.g., the evolution of a network over time, that are not commonly available, and thus they do not generalize well to the broad evaluation setting of this study. Here, we use a different, more general approach to evaluate and compare supervised link prediction methods on a large set of static networks.

Specifically, we exploit two features of our empirical networks (Fig.~\ref{fig:AD}) to construct reasonably reliable training sets. 
First, all observed non-edges $V\times V - E'$ in observed graph $G'=(V,E')$ are taken as negative examples (non-links). 
If $G$ is a snapshot of an evolving network, then links that form in the future of $G$ will form between pairs that are not currently connected. 
Therefore,  the non-links of $G$ can reasonably be considered as negative examples up to the time of observation. 
Second, most real-world networks, for which link prediction is relevant, are sparse. In this case, considering the non-links as negative examples includes only a small number of negative examples in the training set, which are in fact positive examples in the test set. Although these mislabeled edges are not true negative examples, their sparsity in the training set (because the size of the non-links set is $O(n^{2})$ compared to the $O(n)$ size of the missing links set, this approach can only induce a $O(1/n)$ bias in the learning) is likely compensated for by the improved generalizability of taking a supervised learning approach compared to an unsupervised approach.

\section{Diversity in prediction error}
\label{sec:SI_DPE}

\begin{figure*}[t!]
\centering
\begin{tabular}{cc}
\includegraphics[width=0.4\textwidth]{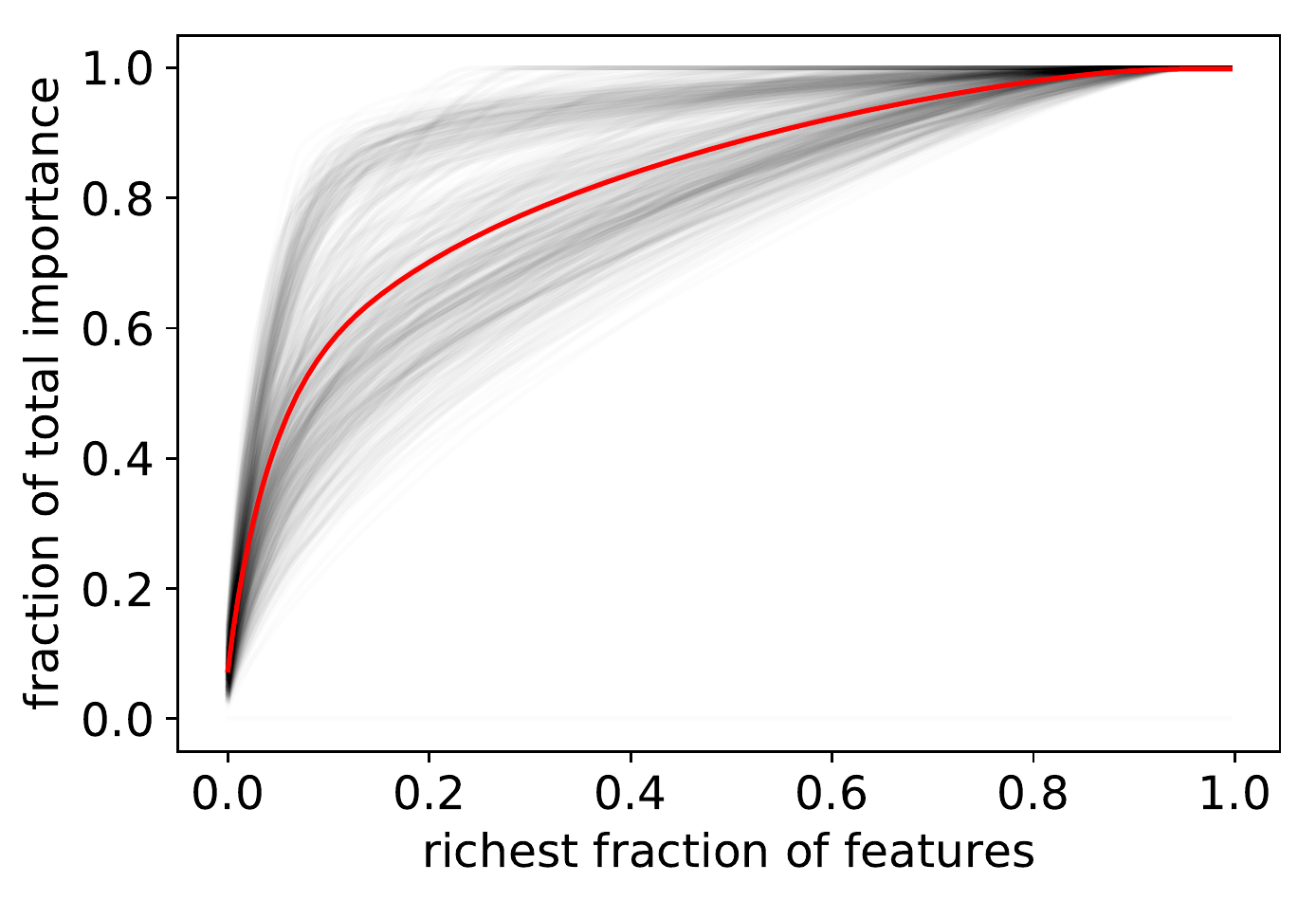}
\end{tabular}
\caption{Lorenz curves of the importance of the features. These curves illustrate
that in a large portion of empirical networks, a very large 
fraction of learned ``importance'' belongs to a small fraction of predictors. 
The red solid line shows the average Lorenz curve over the 548 networks.} 
\label{fig:LCF}
\end{figure*}

A Lorenz curve is a  standard method to visualize the skewness of predictor importances on individual networks. 
Fig.~\ref{fig:LCF} shows the set of 548 curves for the learned importances for each of the networks in our empirical corpus, along with the average curve across the ensemble (red solid line).  
This ensemble exhibits a mean Gini coefficient of $0.64\pm0.14$ (mean$\pm$stddev), and illustrates that the importances tend to be highly skewed, such that a relatively small subset of predictors account for the majority of prediction accuracy.

\medskip The entropy of a distribution is a standard summary statistic of such variation, and provides a compact measure for comparing different distributions. 
Given a discrete random variable $X$ drawn from a probability distribution $p$, the entropy is defined as $H(X)=E_{p(X)}[-\log(p(X))]$, and can be interpreted as the amount of uncertainty in $X$, the average number of bits we need to store $X$, or the minimum number of binary questions on average to guess a draw from $X$~\cite{cover2012elements}. The maximum entropy of discrete random variable occurs for a uniform distribution, and is simply $\log_2(L)$ where $L$ is the number of possible outcomes for $X$.

We begin by computing the learned feature importance entropies for each domain in each family (Table~\ref{tab:FIE}). To calculate these, we first choose all the networks in a domain $j$, as either social, biological, economic, technological, information, or transportation networks) and a set of predictors $\ell$, as either (i)~all 203 predictors, (ii)~the 42 topological predictors, (iii)~the 11 model-based predictors, or (iv)~the 150 embedding predictors. We then learn the feature importances of this set for each domain via supervised learning, as described above.%
\footnote{Unless otherwise noted, the reported results are based on a training a random forest. We also used AdaBoost and XGBoost and similar results have been observed (see Tables~\ref{tab:xgb_f}-\ref{tab:ada_AUC}).}
We denote the feature importances of all predictors in a family $\ell$ and for networks in domain $j$ by a vector $X^{(\ell)}_j$. The ``probability'' associated with the $i$-th predictor in family $\ell$ and for networks in domain $j$ is then computed as $p^{(\ell)}_{ij} = X^{(\ell)}_{ij} \left/ \sum_i X^{(\ell)}_{ij} \right.$. For each setting, the entropy of the corresponding distribution is reported in Table~\ref{tab:FIE}.

Comparing the entropy of the learned importances with the simple upper-limit entropy given by a uniform distribution illustrates the diversity of learned importances among the predictors. To provide a more intuitive sense of how skewed the distribution is, we compare the empirical entropy value with that of a simple piece-wise 
artificial distribution. 
Specifically, we consider a distribution in which at least 90\% of the density is allocated uniformly across
the best $x$\% of the predictors, with the remaining density allocated uniformly across the rest. We then choose the $x$ that minimizes the difference between this model entropy and the empirical entropy.

\medskip \noindent \textit{All predictors.} Applied to the importances of all predictors, only $9$\% of predictors account for 90\% of the importance  in social networks. Other domains require far  more predictors, e.g., $37$\% for biological and $45$\% for technological networks. Notably, the top $x$\% in each family of predictors are different across domains. The values in Table~\ref{tab:FIE} show that most of the variation in importances can be explained by at most 91 of 203 total predictors for technological networks, 26 out of 42 topol.\ predictors for biological networks, 8 of 11 model-based predictors for biological, economic, and technological networks,  and 78 of 150 embed.\ predictors for technological networks.   
Also we see that across these predictor sets most of the uncertainty can be explained by at least 19 out of 203, 9 out of 42, 7 out of 11, and 14 out of 150 predictors for social networks, which illustrates  the simplicity of link prediction in social networks.

\begin{table*}[htbp]
\small
\caption{Predictor importance entropy for each domain in each family. For each family, the ``entropy'' column measures the uncertainty in predictor importance of each domain. Also we consider an artificial distribution on predictors that explain (uniformly) the 90\% of the probability by the best $x$\% of the predictors, and (uniformly) 10\% of the probability by the rest.  We choose $x$ such that the artificial entropy be as close as possible to the empirical entropy. The column ``top $x$\%'', shows the percentage of best predictors with 90\% probability. The ($n$) value shows the corresponding number for the top $x$\%. The ``uniform'' column reports the entropy if predictor importance were uniform. The ``feature-wise'' row within each family reports the entropy held by each predictor, summing across domains. Entropies are reported in units of bits.} 
\centering
\begin{tabular}{c|l|l|l|c}
\hline
Family & Domain & Entropy & Top $x$\% (n) & Uniform\\ \hline
\multirow{ 7}{*}{\rotatebox[origin=c]{90}{\parbox[c]{2cm}{\centering all topol., model, and embed. predictors (203)}}} & social & $5.03$ & $9.36 (19)$ & \multirow{ 7}{*}{$7.66$} \\
& biology & $6.79$ & $37.44 (76)$ &  \\ 
& economy & $6.57$ & $31.03 (63)$ &    \\ 
& technology & $7$ & $44.83 (91)$  &   \\ 
& information & $6.41$ & $27.59 (56)$ &   \\ 
& transportation & $6.67$ & $33.99 (69)$ &    \\
& feature-wise &  $6.71$ & $34.98 (71)$ &     \\ \hline
Family & Domain & Entropy & Top $x$\% (n) & Uniform\\ \hline
\multirow{ 6}{*}{\rotatebox[origin=c]{90}{\parbox[c]{2cm}{\centering all topol. predictors (42) }}} & social & $3.85$ & $21.43 (9)$ & \multirow{ 6}{*}{$5.39$} \\
& biology & $5.09$ & $61.9 (26)$&    \\ 
& economy & $4.65$ & $42.86 (18)$ &    \\ 
& technology & $5.01$ & $57.14 (24)$ &    \\ 
& information & $4.81$ & $47.62 (20)$&     \\ 
& transportation & $4.9$ & $52.38 (22)$ &      \\ 
& feature-wise & $5.08$  &  $61.9 (26)$ &     \\ \hline
Family & Domain & Entropy & Top $x$\% (n) & Uniform\\ \hline
\multirow{ 6}{*}{\rotatebox[origin=c]{90}{\parbox[c]{2cm}{\centering all model-based predictors (11) }}} & social & $3.14$ & $63.64 (7)$ & \multirow{ 6}{*}{$3.46$ } \\
& biology & $3.35$ &  $72.73 (8)$&    \\ 
& economy & $3.36$ & $72.73 (8)$ &    \\ 
& technology & $3.37$ & $72.73 (8)$ &    \\ 
& information & $2.94$ & $54.55 (6)$ &    \\ 
& transportation & $3.24$ & $63.64 (7)$ &     \\ 
& feature-wise &  $3.31$  & $72.73 (8)$ &     \\ \hline
Family & Domain & Entropy & Top $x$\% (n) & Uniform\\ \hline
\multirow{ 6}{*}{\rotatebox[origin=c]{90}{\parbox[c]{2cm}{\centering all embed. predictors (150) }}} & social & $4.65$ &  $9.33 (14)$ & \multirow{ 6}{*}{$7.23$} \\
& biology & $6.51$ & $42.67  (64)$ &    \\ 
& economy & $6.38$ & $38.67 (58)$ &    \\ 
& technology & $6.74$ & $52  (78)$ &     \\ 
& information & $5.19$ & $14.67  (22)$ &    \\ 
& transportation & $6.62$ & $46.67  (70)$ &     \\ 
& feature-wise &  $6.23$  &  $34 (51)$ &     \\ \hline
\end{tabular}
\label{tab:FIE}
\end{table*}

\medskip \noindent \textit{Feature-wise entropy.} 
We also compute the feature-wise entropy for each family $\ell$, which captures the distribution of learned predictor importances, summing across domains. We denote the predictor importance of all predictors in a family $\ell$ by a vector $X^{\ell}$. The importance ``probability'' in the $i$-th entry of this vector for family $\ell$ can be computed as $p^{(\ell)}_{i} = \sum_j X^{(\ell)}_{ij}\left/ \sum_{ij} X^{(\ell)}_{ij}\right.$, which quantifies the proportion of total importance of predictor $i$ in all domains versus the total importance of all predictors. For each family, the entropy of the corresponding probability distribution is reported in Table~\ref{tab:FIE}. And, as before, comparing the entropy of the learned importances with the simple upper-limit entropy given by a uniform distribution illustrates that regardless of the domain, the importances are spread widely across predictors.

\medskip \noindent \textit{Family-wise entropy.} 
Finally, we compute the family-wise entropies for each domain $j$, under an alternative  formulation. Denoting the importance of predictor $i$ in domain $j$ as $X_{ij}$, and the set of all predictors in family $\ell$ as $\mathcal{P}^{\ell}$, we compute the importance ``probability'' of the predictor $i$ in domain $j$ as $p_{ij} = \sum_{i \in \mathcal{P}^{\ell}}X_{ij} \left/  \sum_{i}X_{ij} \right.$. Then, the family-wise entropy can be defined using this distribution (see Table~\ref{tab:FWE}). 
As before, comparing the entropy of  each domain $j$ with the simple upper-limit entropy given by a uniform distribution  illustrates the variance of predictor importances among different families. Moreover, these entropies also illustrate that the variation of importances in social networks is smaller (the most important predictors are in topological and embedding families [see Fig.~\ref{fig:exp_PL_LN} in the main text]), compared to that of non-social networks.

\begin{table}[htbp]
\small
\caption{Family wise entropy. Importance entropy of all features in a family for each domain. The ``uniform'' column reports the entropy if predictor importance were uniform. Entropies are reported in units of bits.} 
\centering
\begin{tabular}{c|l|l|l}
\hline
  & Domain & Entropy &  Uniform\\ \hline
\multirow{ 6}{*}{\rotatebox[origin=c]{90}{\parbox[c]{2cm}{\centering family wise }}} & social & $1.27$ &\multirow{ 6}{*}{$1.58$} \\
& biology & $1.57$ &    \\ 
& economy & $1.58$ &    \\ 
& technology & $1.55$ &    \\ 
& information & $1.58$ &  \\ 
& transportation & $1.56$ &    \\  \hline
\end{tabular}
\label{tab:FWE}
\end{table}

\begin{figure*}[t!]
\centering
\begin{tabular}{cc}
\includegraphics[width=0.6\textwidth]{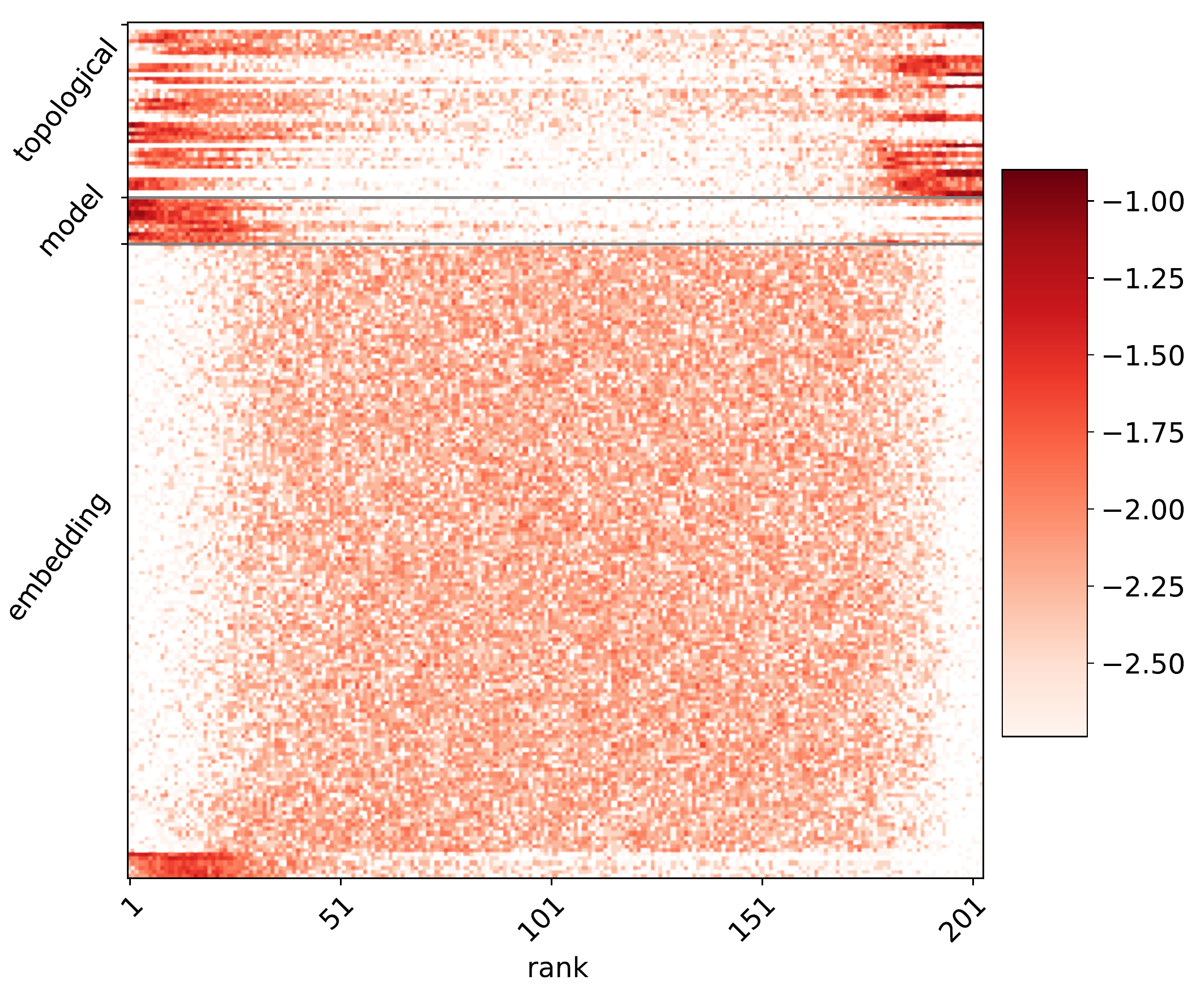}
\end{tabular}
\caption{Distribution of the importance ranks of each predictor across
548 networks. The most important features typically belong to model-based and topological predictor families. 
Among embedding predictor,  the most important correspond 
to the distance measures among the embedded vectors. Almost all vector embedding predictors 
have middling levels of importance, although they are rarely the worst predictors.  
The distribution of the ranks is logarithmic (base 10). }
\label{fig:HFR}
\end{figure*}

\bigskip 
Taking the learned importances for all predictors, Fig.~\ref{fig:HFR} plots the distributions of the importance-ranks (how often predictor $i$ was the $j$th most important predictor) for all 203 predictors, applied to all 548 networks. This visualization reveals that among  the most important predictors (high importance-rank across networks) are those in the model-based family, along with a subset of topological predictors, and the six notions of distance or similarity for embedding-based predictors. The least important predictors (low importance-rank across networks) fall primarily in  the topological family and a few model-based predictors.  
None of embedding predictors ranked among the least important, and instead nearly all of them rank in the broad middle of overall importance. 
Most embedding-based predictors do rank highly for a few individual networks, but it is a different subset of embedding predictors for each network. Thus, across networks, embedding predictors are uniformly middling in their importance, and none are dominant in a network domain. 

\begin{figure*}[t!]
\centering
\begin{tabular}{cc}
\includegraphics[width=1\textwidth]{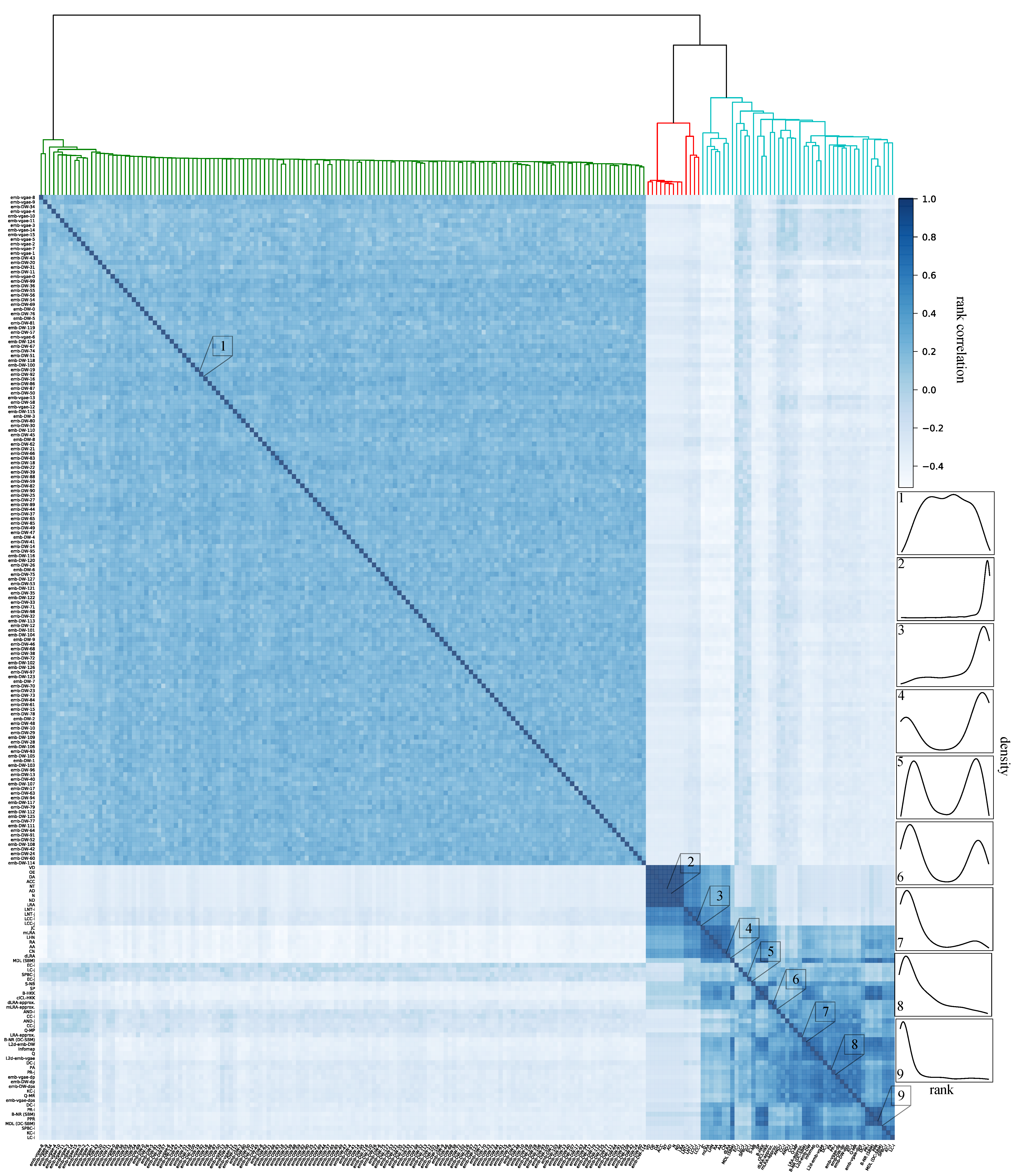}
\end{tabular}
\caption{A clustering of features based on the similarities of their rank in Gini importance.  
Clusters show similar importance distribution among 548 networks. Embedding predictors appear in middle ranks uniformly for different networks (inset 1). The red cluster shows the worst importance among different networks (insets 2--3). The cyan cluster shows better importance and the most important features are located to the right of this group (insets 4--9).}
\label{fig:CFB}
\end{figure*}

\medskip \noindent \textit{Categorizing predictors by their importance-rank distributions.}
To analyze and identify the most important predictors in comparison with the least important predictors on average, we extract a  hierarchical clustering of the rank similarities of Fig.~\ref{fig:HFR}, which is shown in Fig.~\ref{fig:CFB}. The large group of predictors (green cluster) on the left of the hierarchy correspond to the embedding-based predictors, whose distribution of importances is concentrated in the middle range (Fig.~\ref{fig:CFB}, inset panel 1). 
A second group (red cluster) corresponds to the predictors that are nearly always the least important across networks, such as VD, OE, DA, and ACC (Fig.~\ref{fig:CFB}, inset panel 2). And a third group (cyan cluster) includes to predictors that receive high importance across nearly all 548 networks (Fig.~\ref{fig:CFB}, inset panels 7--9), as well as some with more bimodal importance (Fig.~\ref{fig:CFB}, inset panels 4--6).

\begin{figure*}[t!]
\centering
\begin{tabular}{cc}
\includegraphics[width=0.6\textwidth]{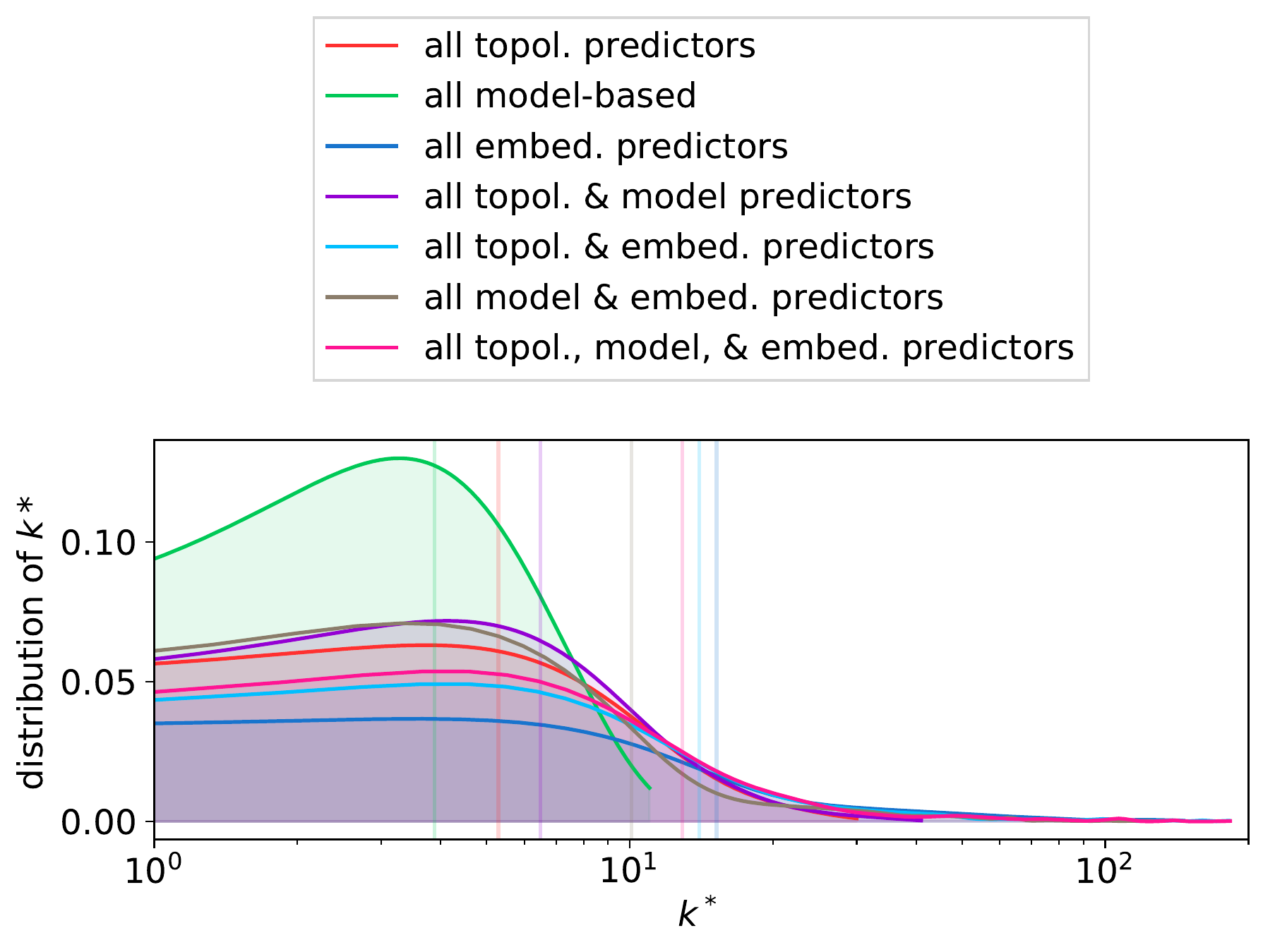}
\end{tabular}
\caption{Distribution of the minimum number of features $k^{\ast}$ that is needed to achieve at least 95\% of final AUC for each family of stacking methods.}
\label{fig:DKstar}
\end{figure*}

\medskip \noindent \textit{Minimal number of features for stacking.}
Fig.~\ref{fig:DKstar} shows the distribution of the minimum number of predictors $k^{\ast}$, that is needed to achieve at least 95\% of final AUC for each family of stacking methods. These curves highlight that in a large portion of networks, we can achieve high predictability using roughly 10 predictors. 

\begin{figure*}[t!]
\centering
\begin{tabular}{cc}
\includegraphics[width=0.6\textwidth]{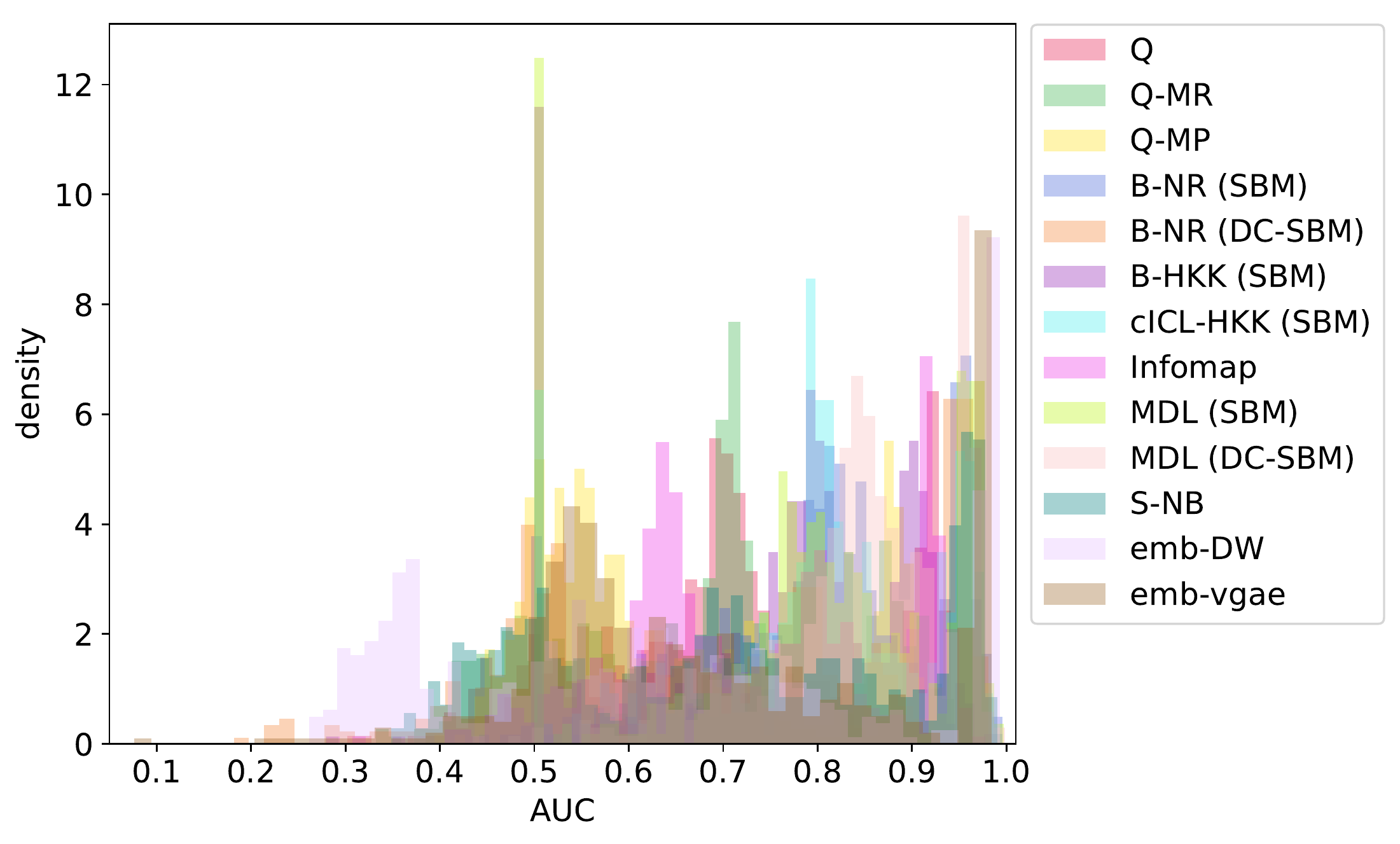}
\end{tabular}
\caption{Histogram of AUC performances on 548 empirical networks for all 11 model-based ``weak learners'' and two embedding link predictors used in model stacking.}
\label{fig:HWL_me}
\end{figure*}

\begin{figure*}[t!]
\centering
\begin{tabular}{cc}
\includegraphics[width=0.8\textwidth]{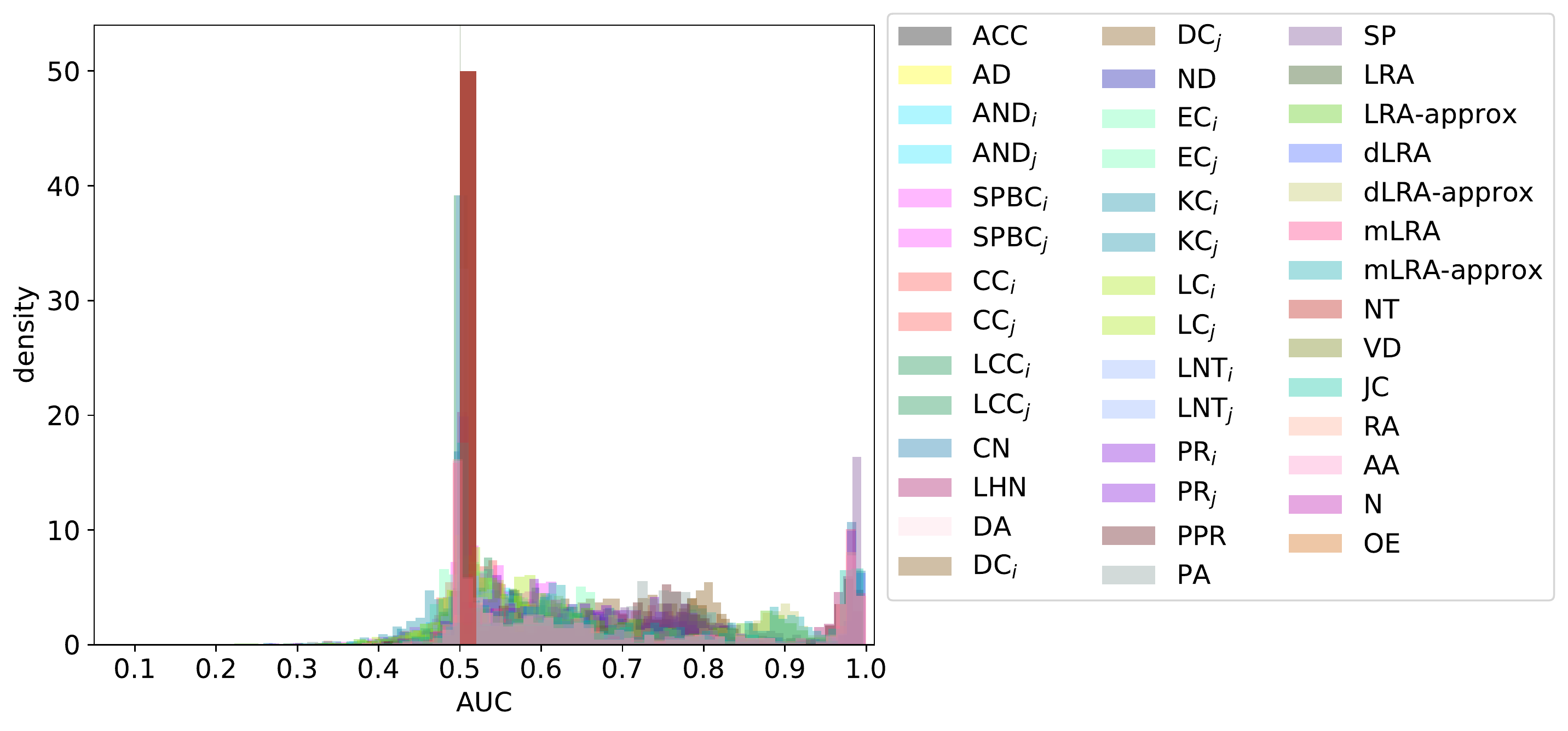}
\end{tabular}
\caption{Histogram of AUC performances on 548 empirical networks for all 42 individual topological ``weak learners'' used in model stacking.}
\label{fig:HWL_top}
\end{figure*}

\medskip \noindent \textit{Performance as weak learners.}
Considering each predictor as a ``weak learner'' from the perspective of the Adaboost theorem, Figs.~\ref{fig:HWL_me} and~\ref{fig:HWL_top} show the histogram of AUC performances of all model-based and topological individual predictors across the 548 networks in our empirical corpus. 
The large majority of these predictors have AUC larger than $0.5$, while a modest portion of individual topological predictors fall below this threshold, meaning that they are not useful in link prediction for a given network. (These topological predictors are of the ``global''  type (see SI Appendix, section~\ref{sec:A} and Table~\ref{table:top_feat}) and are not expected to be individually predictive. 
We note, however, that these predictors are likely to be useful in any transfer learning setting, in which we train on a subset of networks and apply the model to unseen networks. 
Transfer learning for link prediction is out of scope of the present work and we leave it for future study.  

\medskip \noindent \textit{AUC, precision, and recall across tests.}
Tables~\ref{tab:APR_indiv_real} and~\ref{tab:APR_indiv_synt} present the link prediction performance measured by AUC, precision, and recall, for all individual topological predictors applied to the 548 real-world networks in our empirical corpus and the 45 generated synthetic data.

\begin{table*}[t!]
\centering
\caption{Link prediction performance (mean$\pm$std.\ err.), measured by AUC, precision, and recall, for individual topological predictors applied to the 548 structurally diverse networks in our corpus.
}
\begin{tabular}{l|c|c|c}
\hline \hline
algorithm & AUC & precision & recall\\ 
\hline\hline
ACC & $ 0.5 \pm 0.0 $ & $ 0.01 \pm 0.02 $ & $ 0.48 \pm 0.5 $ \\ \hline
AD & $ 0.5 \pm 0.0 $ & $ 0.02 \pm 0.02 $ & $ 0.51 \pm 0.5 $ \\ \hline
AND$_i$ & $ 0.6 \pm 0.12 $ & $ 0.05 \pm 0.04 $ & $ 0.46 \pm 0.21 $ \\ \hline
AND$_j$ & $ 0.61 \pm 0.12 $ & $ 0.05 \pm 0.05 $ & $ 0.48 \pm 0.2 $ \\ \hline
SPBC$_i$ & $ 0.58 \pm 0.09 $ & $ 0.06 \pm 0.06 $ & $ 0.44 \pm 0.16 $ \\ \hline
SPBC$_j$ & $ 0.55 \pm 0.08 $ & $ 0.05 \pm 0.06 $ & $ 0.41 \pm 0.24 $ \\ \hline
CC$_i$ & $ 0.56 \pm 0.08 $ & $ 0.04 \pm 0.03 $ & $ 0.5 \pm 0.23 $ \\ \hline
CC$_j$ & $ 0.6 \pm 0.1 $ & $ 0.05 \pm 0.03 $ & $ 0.54 \pm 0.21 $ \\ \hline
LCC$_i$ & $ 0.55 \pm 0.07 $ & $ 0.05 \pm 0.07 $ & $ 0.44 \pm 0.36 $ \\ \hline
LCC$_j$ & $ 0.53 \pm 0.05 $ & $ 0.04 \pm 0.06 $ & $ 0.45 \pm 0.37 $ \\ \hline
CN & $ 0.68 \pm 0.19 $ & $ 0.21 \pm 0.27 $ & $ 0.7 \pm 0.37 $ \\ \hline
LHN & $ 0.66 \pm 0.18 $ & $ 0.25 \pm 0.3 $ & $ 0.68 \pm 0.37 $ \\ \hline
DA & $ 0.5 \pm 0.0 $ & $ 0.01 \pm 0.02 $ & $ 0.49 \pm 0.5 $ \\ \hline
DC$_i$ & $ 0.68 \pm 0.11 $ & $ 0.06 \pm 0.05 $ & $ 0.61 \pm 0.2 $ \\ \hline
DC$_j$ & $ 0.68 \pm 0.1 $ & $ 0.06 \pm 0.04 $ & $ 0.58 \pm 0.18 $ \\ \hline
ND & $ 0.5 \pm 0.0 $ & $ 0.02 \pm 0.02 $ & $ 0.52 \pm 0.5 $ \\ \hline
EC$_i$ & $ 0.56 \pm 0.09 $ & $ 0.05 \pm 0.06 $ & $ 0.37 \pm 0.17 $ \\ \hline
EC$_j$ & $ 0.6 \pm 0.08 $ & $ 0.05 \pm 0.05 $ & $ 0.5 \pm 0.21 $ \\ \hline
KC$_i$ & $ 0.56 \pm 0.09 $ & $ 0.05 \pm 0.06 $ & $ 0.47 \pm 0.19 $ \\ \hline
KC$_j$ & $ 0.59 \pm 0.1 $ & $ 0.05 \pm 0.06 $ & $ 0.54 \pm 0.22 $ \\ \hline
LC$_i$ & $ 0.58 \pm 0.09 $ & $ 0.06 \pm 0.06 $ & $ 0.44 \pm 0.16 $ \\ \hline
LC$_j$ & $ 0.55 \pm 0.07 $ & $ 0.05 \pm 0.06 $ & $ 0.41 \pm 0.24 $ \\ \hline
LNT$_i$ & $ 0.55 \pm 0.07 $ & $ 0.04 \pm 0.05 $ & $ 0.51 \pm 0.35 $ \\ \hline
LNT$_j$ & $ 0.54 \pm 0.07 $ & $ 0.04 \pm 0.05 $ & $ 0.51 \pm 0.36 $ \\ \hline
PR$_i$ & $ 0.64 \pm 0.1 $ & $ 0.06 \pm 0.05 $ & $ 0.48 \pm 0.18 $ \\ \hline
PR$_j$ & $ 0.63 \pm 0.11 $ & $ 0.06 \pm 0.04 $ & $ 0.51 \pm 0.18 $ \\ \hline
PPR & $ 0.75 \pm 0.15 $ & $ 0.21 \pm 0.26 $ & $ 0.57 \pm 0.28 $ \\ \hline
PA & $ 0.69 \pm 0.1 $ & $ 0.06 \pm 0.05 $ & $ 0.61 \pm 0.19 $ \\ \hline
SP & $ 0.76 \pm 0.15 $ & $ 0.15 \pm 0.18 $ & $ 0.73 \pm 0.3 $ \\ \hline
LRA & $ 0.5 \pm 0.0 $ & $ 0.01 \pm 0.02 $ & $ 0.51 \pm 0.5 $ \\ \hline
LRA-approx & $ 0.67 \pm 0.15 $ & $ 0.17 \pm 0.19 $ & $ 0.42 \pm 0.3 $ \\ \hline
dLRA & $ 0.68 \pm 0.19 $ & $ 0.2 \pm 0.27 $ & $ 0.71 \pm 0.36 $ \\ \hline
dLRA-approx & $ 0.69 \pm 0.14 $ & $ 0.15 \pm 0.19 $ & $ 0.56 \pm 0.31 $ \\ \hline
mLRA & $ 0.67 \pm 0.19 $ & $ 0.21 \pm 0.28 $ & $ 0.68 \pm 0.38 $ \\ \hline
mLRA-approx & $ 0.68 \pm 0.14 $ & $ 0.14 \pm 0.18 $ & $ 0.56 \pm 0.3 $ \\ \hline
NT & $ 0.5 \pm 0.0 $ & $ 0.01 \pm 0.02 $ & $ 0.48 \pm 0.5 $ \\ \hline
VD & $ 0.5 \pm 0.0 $ & $ 0.01 \pm 0.02 $ & $ 0.46 \pm 0.5 $ \\ \hline
JC & $ 0.67 \pm 0.19 $ & $ 0.23 \pm 0.29 $ & $ 0.68 \pm 0.38 $ \\ \hline
RA & $ 0.67 \pm 0.19 $ & $ 0.24 \pm 0.31 $ & $ 0.68 \pm 0.38 $ \\ \hline
AA & $ 0.67 \pm 0.19 $ & $ 0.24 \pm 0.31 $ & $ 0.68 \pm 0.38 $ \\ \hline
N & $ 0.5 \pm 0.0 $ & $ 0.02 \pm 0.02 $ & $ 0.52 \pm 0.5 $ \\ \hline
OE & $ 0.5 \pm 0.0 $ & $ 0.01 \pm 0.02 $ & $ 0.45 \pm 0.5 $ \\ \hline
	\end{tabular}
	\label{tab:APR_indiv_real}  
\end{table*} 
\begin{table*}[t!]
\centering
\caption{
Link prediction performance (mean$\pm$std.\ err.), measured by AUC, precision, and recall, for individual topological predictors applied to 
the 45 synthetic networks.
}
\begin{tabular}{l|c|c|c}
\hline \hline
algorithm & AUC & precision & recall\\ 
\hline\hline
ACC & $ 0.5 \pm 0.0 $ & $ 0.02 \pm 0.02 $ & $ 0.44 \pm 0.5 $ \\ \hline
AD & $ 0.5 \pm 0.0 $ & $ 0.02 \pm 0.02 $ & $ 0.49 \pm 0.5 $ \\ \hline
AND$_i$ & $ 0.57 \pm 0.07 $ & $ 0.06 \pm 0.03 $ & $ 0.41 \pm 0.16 $ \\ \hline
AND$_j$ & $ 0.56 \pm 0.07 $ & $ 0.06 \pm 0.03 $ & $ 0.4 \pm 0.12 $ \\ \hline
SPBC$_i$ & $ 0.61 \pm 0.1 $ & $ 0.09 \pm 0.06 $ & $ 0.44 \pm 0.13 $ \\ \hline
SPBC$_j$ & $ 0.61 \pm 0.1 $ & $ 0.08 \pm 0.06 $ & $ 0.46 \pm 0.17 $ \\ \hline
CC$_i$ & $ 0.61 \pm 0.11 $ & $ 0.05 \pm 0.03 $ & $ 0.59 \pm 0.22 $ \\ \hline
CC$_j$ & $ 0.6 \pm 0.11 $ & $ 0.05 \pm 0.03 $ & $ 0.63 \pm 0.23 $ \\ \hline
LCC$_i$ & $ 0.6 \pm 0.1 $ & $ 0.09 \pm 0.07 $ & $ 0.51 \pm 0.2 $ \\ \hline
LCC$_j$ & $ 0.62 \pm 0.1 $ & $ 0.08 \pm 0.05 $ & $ 0.48 \pm 0.21 $ \\ \hline
CN & $ 0.71 \pm 0.14 $ & $ 0.19 \pm 0.15 $ & $ 0.54 \pm 0.28 $ \\ \hline
LHN & $ 0.7 \pm 0.14 $ & $ 0.22 \pm 0.16 $ & $ 0.55 \pm 0.29 $ \\ \hline
DA & $ 0.5 \pm 0.0 $ & $ 0.02 \pm 0.02 $ & $ 0.49 \pm 0.5 $ \\ \hline
DC$_i$ & $ 0.67 \pm 0.13 $ & $ 0.08 \pm 0.05 $ & $ 0.55 \pm 0.16 $ \\ \hline
DC$_j$ & $ 0.67 \pm 0.12 $ & $ 0.08 \pm 0.05 $ & $ 0.57 \pm 0.16 $ \\ \hline
ND & $ 0.5 \pm 0.0 $ & $ 0.02 \pm 0.02 $ & $ 0.47 \pm 0.5 $ \\ \hline
EC$_i$ & $ 0.62 \pm 0.11 $ & $ 0.08 \pm 0.05 $ & $ 0.46 \pm 0.14 $ \\ \hline
EC$_j$ & $ 0.62 \pm 0.12 $ & $ 0.08 \pm 0.05 $ & $ 0.46 \pm 0.16 $ \\ \hline
KC$_i$ & $ 0.57 \pm 0.08 $ & $ 0.07 \pm 0.08 $ & $ 0.42 \pm 0.11 $ \\ \hline
KC$_j$ & $ 0.57 \pm 0.09 $ & $ 0.06 \pm 0.03 $ & $ 0.44 \pm 0.16 $ \\ \hline
LC$_i$ & $ 0.61 \pm 0.1 $ & $ 0.09 \pm 0.06 $ & $ 0.45 \pm 0.13 $ \\ \hline
LC$_j$ & $ 0.61 \pm 0.1 $ & $ 0.08 \pm 0.05 $ & $ 0.46 \pm 0.14 $ \\ \hline
LNT$_i$ & $ 0.63 \pm 0.11 $ & $ 0.08 \pm 0.07 $ & $ 0.57 \pm 0.21 $ \\ \hline
LNT$_j$ & $ 0.63 \pm 0.11 $ & $ 0.07 \pm 0.04 $ & $ 0.57 \pm 0.2 $ \\ \hline
PR$_i$ & $ 0.65 \pm 0.12 $ & $ 0.09 \pm 0.06 $ & $ 0.51 \pm 0.14 $ \\ \hline
PR$_j$ & $ 0.65 \pm 0.12 $ & $ 0.09 \pm 0.06 $ & $ 0.5 \pm 0.15 $ \\ \hline
PPR & $ 0.74 \pm 0.14 $ & $ 0.16 \pm 0.14 $ & $ 0.54 \pm 0.23 $ \\ \hline
PA & $ 0.72 \pm 0.16 $ & $ 0.1 \pm 0.07 $ & $ 0.62 \pm 0.18 $ \\ \hline
SP & $ 0.75 \pm 0.14 $ & $ 0.13 \pm 0.13 $ & $ 0.72 \pm 0.18 $ \\ \hline
LRA & $ 0.5 \pm 0.0 $ & $ 0.01 \pm 0.02 $ & $ 0.38 \pm 0.48 $ \\ \hline
LRA-approx & $ 0.69 \pm 0.14 $ & $ 0.15 \pm 0.16 $ & $ 0.51 \pm 0.21 $ \\ \hline
dLRA & $ 0.71 \pm 0.14 $ & $ 0.18 \pm 0.14 $ & $ 0.54 \pm 0.27 $ \\ \hline
dLRA-approx & $ 0.73 \pm 0.13 $ & $ 0.17 \pm 0.12 $ & $ 0.51 \pm 0.19 $ \\ \hline
mLRA & $ 0.68 \pm 0.13 $ & $ 0.18 \pm 0.15 $ & $ 0.51 \pm 0.28 $ \\ \hline
mLRA-approx & $ 0.7 \pm 0.12 $ & $ 0.12 \pm 0.11 $ & $ 0.49 \pm 0.19 $ \\ \hline
NT & $ 0.5 \pm 0.0 $ & $ 0.02 \pm 0.02 $ & $ 0.49 \pm 0.5 $ \\ \hline
VD & $ 0.5 \pm 0.0 $ & $ 0.02 \pm 0.02 $ & $ 0.6 \pm 0.49 $ \\ \hline
JC & $ 0.69 \pm 0.14 $ & $ 0.21 \pm 0.16 $ & $ 0.5 \pm 0.29 $ \\ \hline
RA & $ 0.7 \pm 0.14 $ & $ 0.21 \pm 0.16 $ & $ 0.52 \pm 0.28 $ \\ \hline
AA & $ 0.71 \pm 0.14 $ & $ 0.21 \pm 0.16 $ & $ 0.51 \pm 0.28 $ \\ \hline
N & $ 0.5 \pm 0.0 $ & $ 0.02 \pm 0.02 $ & $ 0.62 \pm 0.48 $ \\ \hline
OE & $ 0.5 \pm 0.0 $ & $ 0.02 \pm 0.02 $ & $ 0.49 \pm 0.5 $ \\ \hline
	\end{tabular}
	\label{tab:APR_indiv_synt}  
\end{table*} 

\begin{figure*}[t!]
\centering
\begin{tabular}{cc}
\includegraphics[width=1\textwidth]{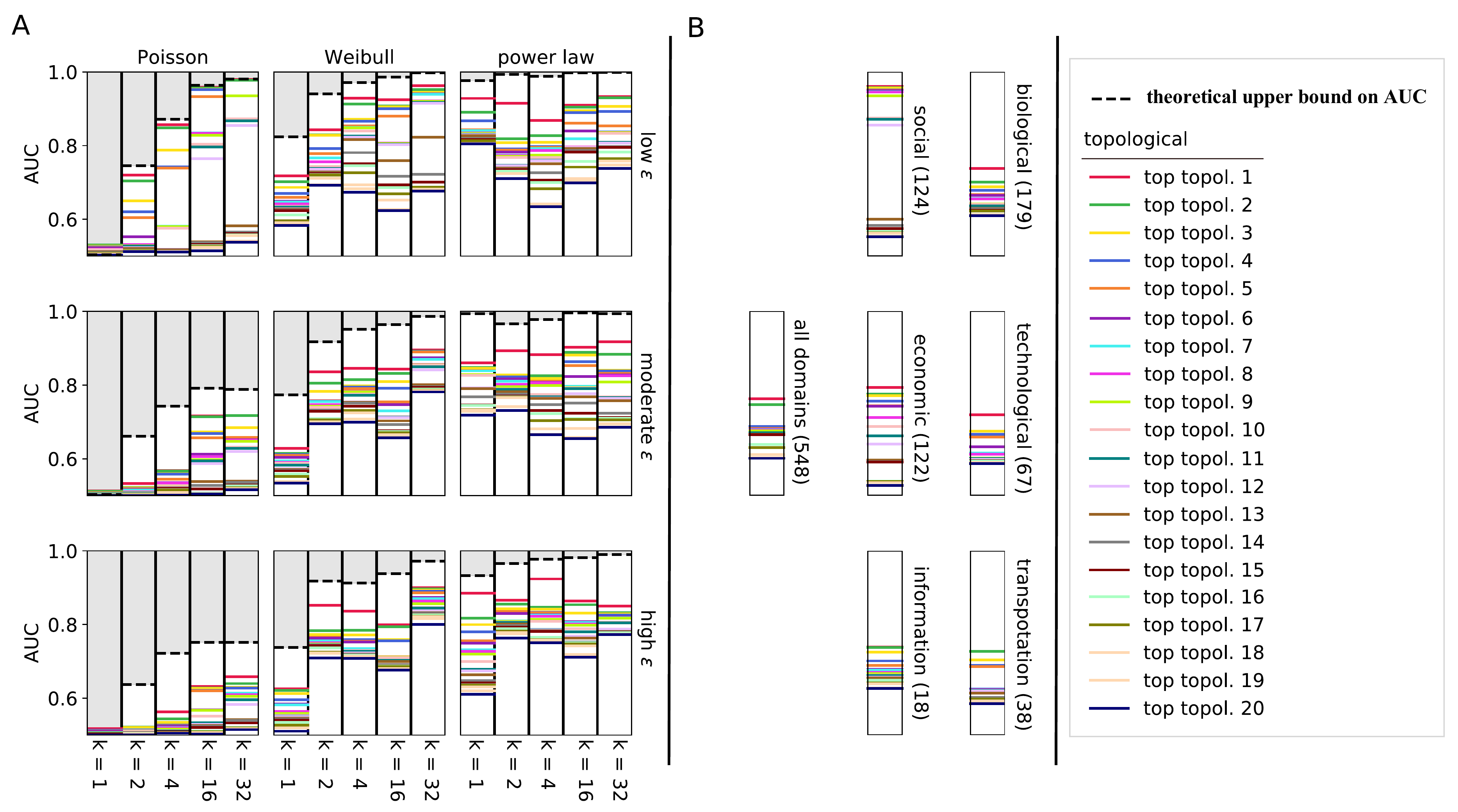}
\vspace{-3mm}
\end{tabular}
 \caption{(A) On synthetic networks, the mean link prediction performance (AUC) of topological individual predictors and all stacked algorithms across three forms of structural variability:\ (left to right, by subpanel) degree distribution variability, from low (Poisson) to high (power law); (top to bottom, by subpanel) fuzziness of community boundaries, ranging from low to high ($\epsilon=\mout/\miin$, the fraction of a node's edges that connect outside its community); and (left to right, within subpanel) the number of communities $k$. Across settings, the dashed line represents the theoretical maximum performance achievable by any link prediction algorithm (SI Appendix, section B). In each instance, stacked models perform optimally or nearly optimally, and generally perform better when networks exhibit heavier-tailed degree distributions and more communities with distinct boundaries. 
(B) On real-world networks, the mean link prediction performance for the same predictors across all domains, and by individual domain. Both overall and within each domain, stacked models, particularly the across-family versions, exhibit superior performance, and they achieve nearly perfect accuracy on social networks. The performance, however, varies considerably across domains, with biological, technological, transportation, and information networks exhibiting the lowest link predictability.}
\label{fig:AUC_topol}
\end{figure*}

\begin{figure*}[t!]
\centering
\begin{tabular}{cc}
\includegraphics[width=1\textwidth]{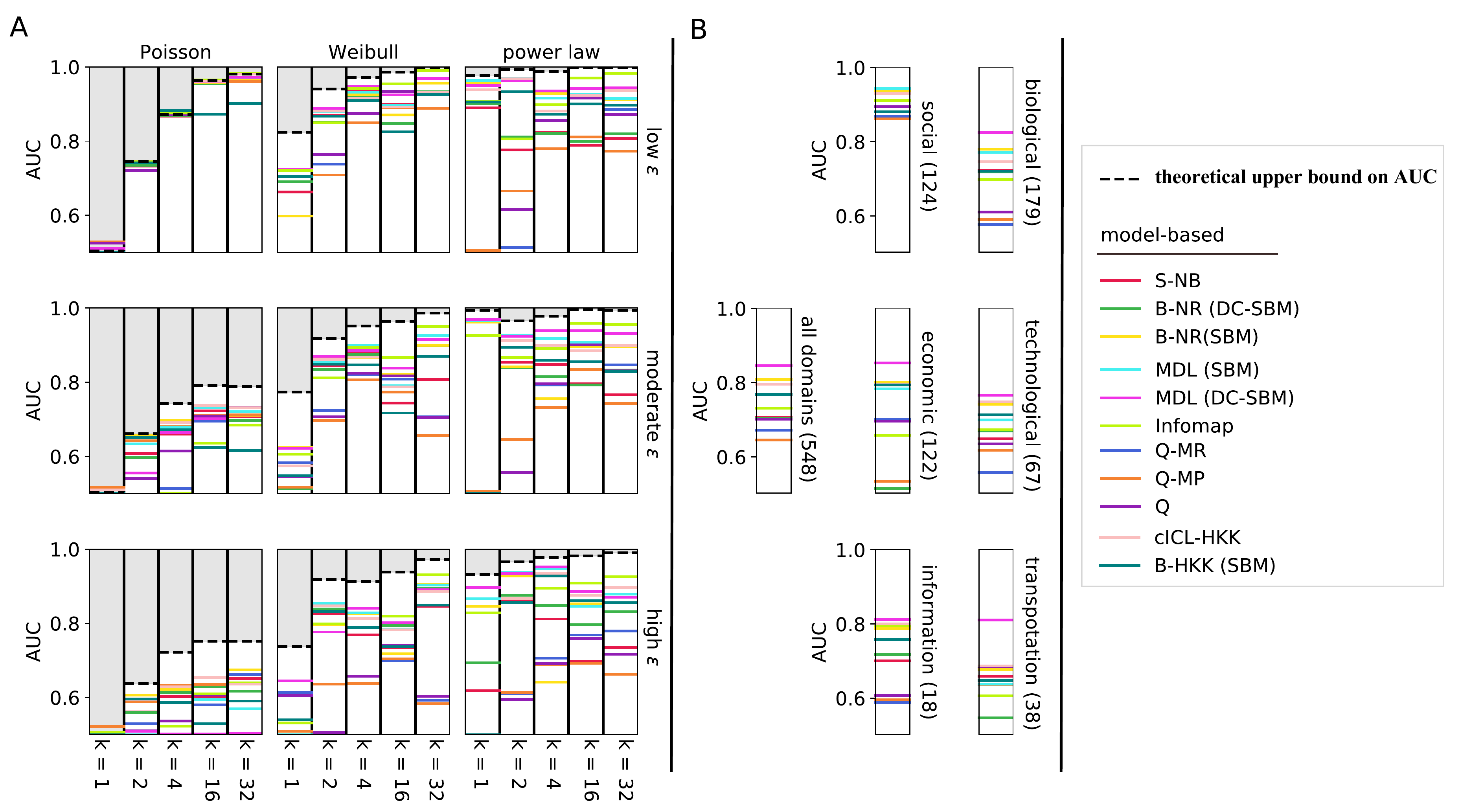}
\vspace{-3mm}
\end{tabular}
     \caption{(A) On synthetic networks, the mean link prediction performance (AUC) of model-based individual predictors and all stacked algorithms across three forms of structural variability:\ (left to right, by subpanel) degree distribution variability, from low (Poisson) to high (power law); (top to bottom, by subpanel) fuzziness of community boundaries, ranging from low to high ($\epsilon=\mout/\miin$, the fraction of a node's edges that connect outside its community); and (left to right, within subpanel) the number of communities $k$. Across settings, the dashed line represents the theoretical maximum performance achievable by any link prediction algorithm (SI Appendix, section B). In each instance, stacked models perform optimally or nearly optimally, and generally perform better when networks exhibit heavier-tailed degree distributions and more communities with distinct boundaries. 
(B) On real-world networks, the mean link prediction performance for the same predictors across all domains, and by individual domain. Both overall and within each domain, stacked models, particularly the across-family versions, exhibit superior performance, and they achieve nearly perfect accuracy on social networks. The performance, however, varies considerably across domains, with technological networks exhibiting the lowest link predictability.}
\label{fig:AUC_MB}
\end{figure*}

\begin{table*}[t!]
\centering
\caption{Mean performance gap for each method in synthetic data.}
\begin{tabular}{l|l}
\hline \hline
Algorithm & Average gap$\langle \Delta \text{AUC} \rangle$\\ 
\hline\hline
Q & $0.187$\\ \hline
Q-MR & $0.198$\\ \hline
Q-MP &  $0.191$ \\ \hline
B-NR (SBM) & $0.085$  \\ \hline 
B-NR (DC-SBM) & $0.123$\\ \hline   
cICL-HKK  & $0.09$   \\ \hline
B-HKK & $0.12$\\ \hline
Infomap  & $0.083$   \\ \hline
MDL (SBM) & $0.075$\\ \hline
MDL (DC-SBM)  &  $0.071$  \\ \hline
S-NB & $0.138$ \\ \hline \hline
mean indiv. model. &  $ 0.124 $ \\ \hline
mean indiv. topol. &  $ 0.257 $ \\ \hline
mean indiv. topol. \& model &  $ 0.229 $ \\ \hline \hline
emb-DW  &  $0.2$  \\ \hline
emb-vgae  &  $0.172$  \\ \hline
all topol.   & $0.066$ \\ \hline
all model-based & $0.069$  \\ \hline
all embed.  & $0.09$  \\ \hline
all topol. \& model &  $0.049$ \\ \hline
all topol. \& embed.   & $0.057$ \\ \hline
all model \& embed.   &  $0.05$ \\ \hline
all topol., model \& embed.   & $0.044$ \\ \hline
	\end{tabular}
	\label{tab:gap}  
\end{table*}

\begin{table*}[t!]
\centering
\caption{The AUC gap of the best 10 predictors with the upper-bound AUC for synthetic data.}
\begin{tabular}{l|l|l}
\hline \hline
Rank&Algorithm & Average gap ($\langle \Delta \text{AUC} \rangle$) \\ 
\hline\hline
1& MDL (DC-SBM)  &  $0.071$  \\ \hline
2& MDL (SBM) & $0.075$ \\ \hline
3& Infomap  & $0.083$   \\ \hline
4& B-NR (SBM) & $0.085$  \\ \hline  
5& cICL-HKK  & $0.09$   \\ \hline
6& B-HKK & $0.12$ \\ \hline
7& SP & $0.121$ \\ \hline
8& B-NR (DC-SBM) & $0.123$\\ \hline
9& S-NB & $0.138$ \\ \hline
10& PPR & $0.139$ \\ \hline
	\end{tabular}
	\label{tab:best_10_AUCgap}  
\end{table*} 

\begin{table*}[t!]
\centering
\caption{Link prediction performance (mean$\pm$std.\ err.), measured by AUC, precision, and recall, for link prediction algorithms applied to the 45 synthetic networks.}
\begin{tabular}{l|c|c|c}
\hline \hline
Algorithm & AUC & Precision & Recall\\
\hline\hline
Q &$ 0.69  \pm 0.16 $ & $ 0.11  \pm  0.14 $ & $ 0.66 \pm 0.15 $ \\ \hline
Q-MR& $ 0.68  \pm 0.17 $ & $ 0.11  \pm  0.14 $ & $ 0.66 \pm 0.15 $ \\ \hline
Q-MP& $ 0.69  \pm 0.13 $ & $ 0.11  \pm  0.1 $ & $ 0.65 \pm 0.15 $ \\ \hline
B-NR (SBM)& $ 0.79  \pm 0.14 $ & $ 0.16  \pm  0.12 $ & $ 0.67 \pm 0.24 $ \\ \hline
B-NR (DC-SBM)& $ 0.75  \pm 0.15 $ & $ 0.17  \pm  0.12 $ & $ 0.7 \pm 0.16 $ \\ \hline
cICL-HKK& $ 0.79  \pm 0.15 $ & $ 0.17  \pm  0.14 $ & $ 0.61 \pm 0.27 $ \\ \hline
B-HKK& $ 0.76  \pm 0.15 $ & $ 0.13  \pm  0.1 $ & $ 0.56 \pm 0.26 $ \\ \hline
Infomap& $ 0.79  \pm 0.17 $ & $ 0.18  \pm  0.17 $ & $ 0.75 \pm 0.16 $ \\ \hline
MDL (SBM) &$ 0.8  \pm 0.16 $ & $ 0.17  \pm  0.13 $ & $ 0.65 \pm 0.3 $ \\ \hline
MDL (DC-SBM)& $ 0.8  \pm 0.16 $ & $ 0.15  \pm  0.11 $ & $ 0.76 \pm 0.16 $\\ \hline
S-NB &$ 0.74  \pm 0.15 $ & $ 0.14  \pm  0.13 $ & $ 0.67 \pm 0.15 $\\ \hline \hline
mean model-based &  $ 0.75  \pm 0.16 $ & $ 0.15  \pm  0.13 $ & $ 0.67 \pm 0.21 $ \\ \hline
mean indiv. topol. &  $ 0.62  \pm 0.13 $ & $ 0.09  \pm  0.11 $ & $ 0.51 \pm 0.3 $ \\ \hline
mean indiv. topol. \& model &  $ 0.65  \pm 0.15 $ & $ 0.1  \pm  0.11 $ & $ 0.54 \pm 0.29 $ \\ \hline \hline
emb-DW & $0.68\pm0.14$ & $0.15\pm0.14$ & $0.36\pm0.28$ \\ \hline
emb-vgae & $0.7\pm0.16$ & $0.06\pm0.03$ & $0.72\pm0.17$ \\ \hline \hline
all topol. & $ 0.81 \pm 0.16 $ & $ 0.4 \pm 0.26 $ & $ 0.46 \pm 0.21 $ \\ \hline
all model-based & $ 0.81 \pm 0.15 $ & $ 0.51 \pm 0.33 $ & $ 0.38 \pm 0.28 $ \\ \hline
all embed. & $ 0.79 \pm 0.15 $ & $ 0.35 \pm 0.27 $ & $ 0.29 \pm 0.26 $ \\ \hline
all topol. \& model  & $ 0.83 \pm 0.14 $ & $ 0.49 \pm 0.33 $ & $ 0.42 \pm 0.25 $ \\ \hline
all topol. \& embed.  & $ 0.82 \pm 0.15 $ & $ 0.41 \pm 0.28 $ & $ 0.42 \pm 0.23 $ \\ \hline
all model \& embed.  & $ 0.83 \pm 0.15 $ & $ 0.47 \pm 0.31 $ & $ 0.36 \pm 0.26 $ \\ \hline
all topol., model \& embed. & $ 0.83 \pm 0.15 $ & $ 0.48 \pm 0.3 $ & $ 0.4 \pm 0.24 $ \\ \hline
 \end{tabular}
	\label{tab:AUC_synt}  
\end{table*}

\begin{table*}[t!]
\centering
\caption{The detailed information of the top 5 topological predictors for synthetic data as presented in Fig. 2 in the manuscript.}
\begin{tabular}{l|l|l|l}
\hline \hline
Region & Model & Number of clusters $k$ & Predictors \\ 
\hline\hline
low $\epsilon$&Poisson&1&[mLRA-approx., PPR, PA, dLRA-approx., PR-j]\\ \hline    
low $\epsilon$&Poisson&2&[PPR, SP, dLRA-approx., LRA-approx., mLRA-approx.] \\ \hline    
low $\epsilon$&Poisson&4&[PPR, SP, LRA-approx., mLRA-approx., dLRA-approx.]\\ \hline    
low $\epsilon$&Poisson&16&[PPR, SP, dLRA-approx., mLRA-approx., LRA-approx.]\\ \hline    
low $\epsilon$&Poisson&32&[PPR, SP, RA, LHN, mLRA]\\ \hline    
low $\epsilon$&Weibull&1&[PR-i, PA, DC-j, EC-i, KC-j]\\ \hline    
low $\epsilon$&Weibull&2&[SP, PA, PPR, LRA-approx., DC-i]\\ \hline    
low $\epsilon$&Weibull&4&[SP, dLRA-approx., LRA-approx., mLRA-approx., PA]\\ \hline    
low $\epsilon$&Weibull&16&[SP, dLRA-approx., mLRA-approx., PPR, LRA-approx.]\\ \hline    
low $\epsilon$&Weibull&32&[PPR, SP, dLRA-approx., AA, LRA-approx.]\\ \hline    
low $\epsilon$&power law&1&[PA, LHN, CN, dLRA, dLRA-approx.]\\ \hline    
low $\epsilon$&power law&2&[PA, DC-i, PR-j, LHN, AND-i]\\ \hline    
low $\epsilon$&power law&4&[PA, PR-i, SP, DC-i, AA]\\ \hline    
low $\epsilon$&power law&16&[SP, LRA-approx., dLRA-approx., PPR, PA]\\ \hline    
low $\epsilon$&power law&32&[dLRA-approx., SP, PPR, LRA-approx., PA]\\ \hline    
moderate $\epsilon$&Poisson&1&[EC-i, SPBC-i, LNT-j, EC-j, AA]\\ \hline    
moderate $\epsilon$&Poisson&2&[LRA-approx., SP, AND-i, PPR, KC-i]\\ \hline    
moderate $\epsilon$&Poisson&4&[dLRA-approx., SP, mLRA-approx., LRA-approx., PPR]\\ \hline    
moderate $\epsilon$&Poisson&16&[mLRA-approx., dLRA-approx., LRA-approx., SP, PPR]\\ \hline    
moderate $\epsilon$&Poisson&32&[PPR, SP, LRA-approx., dLRA, CN]\\ \hline  
moderate $\epsilon$&Weibull&1&[PA, DC-i, dLRA-approx., SPBC-j, mLRA-approx.]\\ \hline    
moderate $\epsilon$&Weibull&2&[PA, SP, PPR, CN, dLRA]\\ \hline    
moderate $\epsilon$&Weibull&4&[SP, PA, mLRA-approx., dLRA-approx., PPR]\\ \hline    
moderate $\epsilon$&Weibull&16&[PPR, SP, dLRA-approx., LRA-approx., PA]\\ \hline    
moderate $\epsilon$&Weibull&32&[dLRA-approx., AA, dLRA, CN, PPR]\\ \hline   
moderate $\epsilon$&power law&1&[PA, EC-i, CC-i, JC, DC-j]\\ \hline     
moderate $\epsilon$&power law&2&[PA, AA, RA, LHN, CN]\\ \hline    
moderate $\epsilon$&power law&4&[PA, LHN, SP, AA, RA]\\ \hline    
moderate $\epsilon$&power law&16&[PPR, dLRA-approx., SP, mLRA-approx., LRA-approx.]\\ \hline    
moderate $\epsilon$&power law&32&[PPR, SP, CN, LHN, JC]\\ \hline
high $\epsilon$&Poisson&1&[EC-i, DC-j, KC-j, LCC-j, RA]\\ \hline        
high $\epsilon$&Poisson&2&[PA, EC-i, DC-j, dLRA, KC-j]\\ \hline    
high $\epsilon$&Poisson&4&[LRA-approx., SP, PPR, DC-j, LNT-j]\\ \hline    
high $\epsilon$&Poisson&16&[SP, PPR, dLRA-approx., LRA-approx., mLRA-approx.]\\ \hline    
high $\epsilon$&Poisson&32&[PPR, SP, mLRA-approx., dLRA-approx., AA]\\ \hline    
high $\epsilon$&Weibull&1&[SP, dLRA-approx., mLRA-approx., PPR, DC-i]\\ \hline    
high $\epsilon$&Weibull&2&[PA, dLRA-approx., SP, DC-j, PR-i]\\ \hline    
high $\epsilon$&Weibull&4&[PA, dLRA-approx., SP, DC-i, DC-j]\\ \hline    
high $\epsilon$&Weibull&16&[SP, PPR, dLRA-approx., PA, AA]\\ \hline    
high $\epsilon$&Weibull&32&[RA, AA, PA, CN, dLRA]\\ \hline  
high $\epsilon$&power law&1&[PA, DC-i, PR-i, LCC-i, LNT-i]\\ \hline      
high $\epsilon$&power law&2&[PA, LHN, AA, dLRA, CN]\\ \hline    
high $\epsilon$&power law&4&[PA, SP, LHN, AA, RA]\\ \hline    
high $\epsilon$&power law&16&[PPR, PA, SP, AA, CN]\\ \hline    
high $\epsilon$&power law&32&[PA, SP, CN, dLRA, JC]\\ \hline    

	\end{tabular}
	\label{tab:AUC_synt_best5}  
\end{table*}  

\begin{table*}[t!]
\centering
\caption{Average AUC, precision, and recall performances of the link prediction algorithms over 124 social networks as a subset of CommunityFitNet corpus. A random forest is used for supervised stacking of methods. Here, the predictors are adjusted for maximum F measure using a model selection through a cross validation on training set. The results are reported on 20\% holdout test set.}
\begin{tabular}{l|c|c|c}
\hline \hline
Algorithm & AUC & Precision & Recall\\ 
\hline\hline
Q & $ 0.89  \pm 0.07 $ & $ 0.42  \pm  0.13 $ & $ 0.85 \pm 0.08 $    \\ \hline    
Q-MR & $ 0.87  \pm 0.07 $ & $ 0.38  \pm  0.16 $ & $ 0.78 \pm 0.07 $  \\ \hline
Q-MP & $ 0.86  \pm 0.08 $ & $ 0.25  \pm  0.07 $ & $ 0.83 \pm 0.09 $   \\ \hline
B-NR (SBM) & $ 0.93  \pm 0.06 $ & $ 0.3  \pm  0.08 $ & $ 0.85 \pm 0.12 $  \\ \hline    
B-NR (DC-SBM)  & $ 0.93  \pm 0.07 $ & $ 0.28  \pm  0.08 $ & $ 0.88 \pm 0.08 $  \\ \hline
cICL-HKK  & $ 0.93  \pm 0.08 $ & $ 0.34  \pm  0.1 $ & $ 0.85 \pm 0.14 $   \\ \hline
B-HKK  & $ 0.88  \pm 0.07 $ & $ 0.17  \pm  0.05 $ & $ 0.79 \pm 0.17 $ \\ \hline
Infomap  &  $ 0.91  \pm 0.04 $ & $ 0.29  \pm  0.08 $ & $ 0.83 \pm 0.05 $   \\ \hline
MDL (SBM)  & $ 0.94  \pm 0.07 $ & $ 0.31  \pm  0.09 $ & $ 0.87 \pm 0.16 $    \\ \hline
MDL (DC-SBM)  & $ 0.93  \pm 0.09 $ & $ 0.26  \pm  0.09 $ & $ 0.89 \pm 0.11 $  \\ \hline
S-NB  &  $ 0.94  \pm 0.07 $ & $ 0.3  \pm  0.1 $ & $ 0.87 \pm 0.08 $   \\ \hline\hline
mean model-based &  $ 0.91  \pm 0.08 $ & $ 0.3  \pm  0.12 $ & $ 0.84 \pm 0.12 $ \\ \hline
mean indiv. topol. &  $ 0.64  \pm 0.19 $ & $ 0.2  \pm  0.27 $ & $ 0.56 \pm 0.33 $ \\ \hline
mean indiv. topol. \& model &  $ 0.7  \pm 0.21 $ & $ 0.22  \pm  0.25 $ & $ 0.62 \pm 0.32 $ \\ \hline \hline
emd-DW & $ 0.95 \pm 0.1 $ & $ 0.45 \pm 0.16 $ & $ 0.92 \pm 0.13 $ \\ \hline
emb-vgae & $ 0.95 \pm 0.08 $ & $ 0.09 \pm 0.02 $ & $ 0.96 \pm 0.09 $ \\ \hline \hline
all topol. & $ 0.97 \pm 0.08 $ & $ 0.89 \pm 0.21 $ & $ 0.88 \pm 0.2 $ \\ \hline
all model-based & $ 0.95 \pm 0.07 $ & $ 0.76 \pm 0.2 $ & $ 0.68 \pm 0.17 $ \\ \hline
all embed. & $ 0.95 \pm 0.11 $ & $ 0.75 \pm 0.23 $ & $ 0.74 \pm 0.23 $ \\ \hline
all topol. \& model  & $ 0.98 \pm 0.06 $ & $ 0.89 \pm 0.22 $ & $ 0.88 \pm 0.19 $ \\ \hline
all topol. \& embed.  & $ 0.96 \pm 0.1 $ & $ 0.86 \pm 0.22 $ & $ 0.83 \pm 0.25 $ \\ \hline
all model \& embed.  & $ 0.96 \pm 0.09 $ & $ 0.78 \pm 0.21 $ & $ 0.74 \pm 0.22 $ \\ \hline
all topol., model \& embed. & $ 0.97 \pm 0.09 $ & $ 0.86 \pm 0.23 $ & $ 0.84 \pm 0.23 $ \\ \hline
	\end{tabular}
	\label{tab:AUC_soc_dom}  
\end{table*} 

\begin{table*}[t!]
\centering
\caption{Average AUC, precision, and recall performances of the link prediction algorithms over 179 biological networks as a subset of CommunityFitNet corpus. A random forest is used for supervised stacking of methods. Here, the predictors are adjusted for maximum F measure using a model selection through a cross validation on training set. The results are reported on 20\% holdout test set.}
\begin{tabular}{l|c|c|c}
\hline \hline
Algorithm & AUC & Precision & Recall\\
\hline\hline
Q &  $ 0.61  \pm 0.12 $ & $ 0.06  \pm  0.09 $ & $ 0.58 \pm 0.13 $    \\ \hline    
Q-MR & $ 0.57  \pm 0.11 $ & $ 0.05  \pm  0.09 $ & $ 0.56 \pm 0.12 $   \\ \hline
Q-MP &  $ 0.59  \pm 0.09 $ & $ 0.06  \pm  0.07 $ & $ 0.52 \pm 0.13 $ \\ \hline
B-NR (SBM) & $ 0.78  \pm 0.13 $ & $ 0.09  \pm  0.09 $ & $ 0.6 \pm 0.21 $  \\ \hline    
B-NR (DC-SBM)  &  $ 0.72  \pm 0.17 $ & $ 0.1  \pm  0.09 $ & $ 0.63 \pm 0.21 $ \\ \hline
cICL-HKK  & $ 0.74  \pm 0.13 $ & $ 0.09  \pm  0.09 $ & $ 0.47 \pm 0.24 $   \\ \hline
B-HKK  & $ 0.72  \pm 0.14 $ & $ 0.11  \pm  0.12 $ & $ 0.39 \pm 0.26 $ \\ \hline
Infomap  &  $ 0.7  \pm 0.12 $ & $ 0.07  \pm  0.09 $ & $ 0.68 \pm 0.11 $   \\ \hline
MDL (SBM)  & $ 0.77  \pm 0.14 $ & $ 0.11  \pm  0.1 $ & $ 0.51 \pm 0.29 $   \\ \hline
MDL (DC-SBM)  & $ 0.82  \pm 0.09 $ & $ 0.09  \pm  0.07 $ & $ 0.75 \pm 0.11 $  \\ \hline
S-NB  & $ 0.72  \pm 0.14 $ & $ 0.09  \pm  0.1 $ & $ 0.64 \pm 0.16 $    \\ \hline\hline
mean model-based &  $ 0.7  \pm 0.15 $ & $ 0.08  \pm  0.09 $ & $ 0.58 \pm 0.21 $ \\ \hline
mean indiv. topol. &  $ 0.59  \pm 0.11 $ & $ 0.06  \pm  0.08 $ & $ 0.51 \pm 0.35 $ \\ \hline
mean indiv. topol. \& model &  $ 0.62  \pm 0.13 $ & $ 0.06  \pm  0.08 $ & $ 0.52 \pm 0.32 $ \\ \hline \hline
emd-DW & $ 0.59 \pm 0.15 $ & $ 0.07 \pm 0.08 $ & $ 0.39 \pm 0.25 $ \\ \hline
emb-vgae & $ 0.63 \pm 0.16 $ & $ 0.04 \pm 0.06 $ & $ 0.62 \pm 0.2 $ \\ \hline \hline
all topol. & $ 0.83 \pm 0.1 $ & $ 0.27 \pm 0.23 $ & $ 0.34 \pm 0.24 $ \\ \hline
all model-based & $ 0.79 \pm 0.12 $ & $ 0.29 \pm 0.29 $ & $ 0.24 \pm 0.25 $ \\ \hline
all embed. & $ 0.68 \pm 0.16 $ & $ 0.17 \pm 0.25 $ & $ 0.12 \pm 0.17 $ \\ \hline
all topol. \& model  & $ 0.83 \pm 0.1 $ & $ 0.35 \pm 0.31 $ & $ 0.23 \pm 0.23 $ \\ \hline
all topol. \& embed.  & $ 0.79 \pm 0.13 $ & $ 0.23 \pm 0.27 $ & $ 0.18 \pm 0.2 $ \\ \hline
all model \& embed.  & $ 0.79 \pm 0.14 $ & $ 0.23 \pm 0.26 $ & $ 0.18 \pm 0.2 $ \\ \hline
all topol., model \& embed. & $ 0.79 \pm 0.15 $ & $ 0.25 \pm 0.27 $ & $ 0.18 \pm 0.2 $ \\ \hline
	\end{tabular}
	\label{tab:AUC_bio_dom}  
\end{table*} 

\begin{table*}[t!]
\centering
\caption{Average AUC, precision, and recall performances of the link prediction algorithms over 122 economic networks as a subset of CommunityFitNet corpus. A random forest is used for supervised stacking of methods. Here, the predictors are adjusted for maximum F measure using a model selection through a cross validation on training set. The results are reported on 20\% holdout test set.}
\begin{tabular}{l|c|c|c}
\hline \hline
Algorithm & AUC & Precision & Recall\\
\hline\hline
Q  & $ 0.69  \pm 0.06 $ & $ 0.04  \pm  0.02 $ & $ 0.69 \pm 0.08 $ \\ \hline  
Q-MR &  $ 0.7  \pm 0.06 $ & $ 0.05  \pm  0.02 $ & $ 0.67 \pm 0.06 $ \\ \hline  
Q-MP &  $ 0.53  \pm 0.06 $ & $ 0.03  \pm  0.02 $ & $ 0.51 \pm 0.11 $ \\ \hline  
B-NR (SBM) & $ 0.8  \pm 0.05 $ & $ 0.07  \pm  0.05 $ & $ 0.6 \pm 0.16 $ \\ \hline  
B-NR (DC-SBM) & $ 0.51  \pm 0.1 $ & $ 0.04  \pm  0.05 $ & $ 0.35 \pm 0.13 $ \\ \hline  
cICL-HKK &$ 0.79  \pm 0.06 $ & $ 0.06  \pm  0.04 $ & $ 0.45 \pm 0.12 $ \\ \hline  
B-HKK & $ 0.79  \pm 0.06 $ & $ 0.06  \pm  0.03 $ & $ 0.44 \pm 0.11 $ \\ \hline  
Infomap & $ 0.66  \pm 0.05 $ & $ 0.05  \pm  0.04 $ & $ 0.62 \pm 0.06 $ \\ \hline  
MDL (SBM) & $ 0.78  \pm 0.05 $ & $ 0.07  \pm  0.05 $ & $ 0.49 \pm 0.14 $ \\ \hline  
MDL (DC-SBM) &  $ 0.85  \pm 0.06 $ & $ 0.09  \pm  0.04 $ & $ 0.79 \pm 0.06 $ \\ \hline  
S-NB & $ 0.49  \pm 0.11 $ & $ 0.03  \pm  0.05 $ & $ 0.55 \pm 0.07 $ \\ \hline  \hline
mean model-based &  $ 0.69  \pm 0.14 $ & $ 0.05  \pm  0.04 $ & $ 0.56 \pm 0.16 $ \\ \hline
mean indiv. topol. &  $ 0.58  \pm 0.12 $ & $ 0.04  \pm  0.06 $ & $ 0.6 \pm 0.39 $ \\ \hline
mean indiv. topol. \& model &  $ 0.6  \pm 0.13 $ & $ 0.04  \pm  0.05 $ & $ 0.59 \pm 0.35 $ \\ \hline \hline
emd-DW & $ 0.37 \pm 0.11 $ & $ 0.09 \pm 0.06 $ & $ 0.12 \pm 0.16 $ \\ \hline
emb-vgae & $ 0.56 \pm 0.07 $ & $ 0.03 \pm 0.02 $ & $ 0.55 \pm 0.1 $ \\ \hline \hline
all topol. & $ 0.83 \pm 0.05 $ & $ 0.31 \pm 0.08 $ & $ 0.28 \pm 0.14 $ \\ \hline
all model-based & $ 0.84 \pm 0.07 $ & $ 0.27 \pm 0.26 $ & $ 0.14 \pm 0.17 $ \\ \hline
all embed. & $ 0.78 \pm 0.07 $ & $ 0.17 \pm 0.1 $ & $ 0.34 \pm 0.18 $ \\ \hline
all topol. \& model  & $ 0.87 \pm 0.05 $ & $ 0.38 \pm 0.25 $ & $ 0.12 \pm 0.15 $ \\ \hline
all topol. \& embed.  & $ 0.86 \pm 0.07 $ & $ 0.3 \pm 0.1 $ & $ 0.41 \pm 0.15 $ \\ \hline
all model \& embed.  & $ 0.87 \pm 0.09 $ & $ 0.21 \pm 0.12 $ & $ 0.42 \pm 0.23 $\\ \hline
all topol., model \& embed. & $ 0.88 \pm 0.1 $ & $ 0.31 \pm 0.11 $ & $ 0.41 \pm 0.18 $ \\ \hline
	\end{tabular}
	\label{tab:AUC_eco_dom}  
\end{table*} 

\begin{table*}[t!]
\centering
\caption{Average AUC, precision, and recall performances of the link prediction algorithms over 67 technological networks as a subset of CommunityFitNet corpus. A random forest is used for supervised stacking of methods. Here, the predictors are adjusted for maximum F measure using a model selection through a cross validation on training set. The results are reported on 20\% holdout test set.}
\begin{tabular}{l|c|c|c}
\hline \hline
Algorithm & AUC & Precision & Recall\\
\hline\hline
Q  & $ 0.63  \pm 0.11 $ & $ 0.04  \pm  0.03 $ & $ 0.58 \pm 0.12 $ \\ \hline
Q-MR  & $ 0.56  \pm 0.11 $ & $ 0.03  \pm  0.02 $ & $ 0.54 \pm 0.09 $ \\ \hline
Q-MP  & $ 0.62  \pm 0.08 $ & $ 0.04  \pm  0.03 $ & $ 0.57 \pm 0.08 $ \\ \hline
B-NR (SBM)  & $ 0.74  \pm 0.11 $ & $ 0.06  \pm  0.05 $ & $ 0.62 \pm 0.2 $ \\ \hline
B-NR (DC-SBM)  & $ 0.67  \pm 0.12 $ & $ 0.06  \pm  0.06 $ & $ 0.63 \pm 0.13 $ \\ \hline
cICL-HKK  & $ 0.75  \pm 0.1 $ & $ 0.08  \pm  0.08 $ & $ 0.59 \pm 0.18 $ \\ \hline
B-HKK  & $ 0.71  \pm 0.11 $ & $ 0.08  \pm  0.08 $ & $ 0.5 \pm 0.2 $ \\ \hline
Infomap  & $ 0.67  \pm 0.13 $ & $ 0.05  \pm  0.04 $ & $ 0.6 \pm 0.12 $ \\ \hline
MDL (SBM)  & $ 0.7  \pm 0.15 $ & $ 0.07  \pm  0.07 $ & $ 0.45 \pm 0.32 $ \\ \hline
MDL (DC-SBM)  & $ 0.77  \pm 0.1 $ & $ 0.07  \pm  0.07 $ & $ 0.68 \pm 0.12 $ \\ \hline
S-NB  & $ 0.65  \pm 0.09 $ & $ 0.04  \pm  0.04 $ & $ 0.56 \pm 0.1 $ \\ \hline \hline
mean model-based &  $ 0.68  \pm 0.13 $ & $ 0.06  \pm  0.06 $ & $ 0.58 \pm 0.17 $ \\ \hline
mean indiv. topol. &  $ 0.58  \pm 0.09 $ & $ 0.05  \pm  0.07 $ & $ 0.48 \pm 0.34 $ \\ \hline
mean indiv. topol. \& model &  $ 0.6  \pm 0.11 $ & $ 0.05  \pm  0.07 $ & $ 0.5 \pm 0.31 $ \\ \hline \hline
emd-DW & $ 0.65 \pm 0.1 $ & $ 0.07 \pm 0.1 $ & $ 0.26 \pm 0.17 $ \\ \hline
emb-vgae & $ 0.64 \pm 0.1 $ & $ 0.03 \pm 0.02 $ & $ 0.63 \pm 0.12 $ \\ \hline \hline
all topol. & $ 0.79 \pm 0.1 $ & $ 0.24 \pm 0.19 $ & $ 0.27 \pm 0.22 $ \\ \hline
all model-based & $ 0.72 \pm 0.13 $ & $ 0.28 \pm 0.33 $ & $ 0.13 \pm 0.15 $ \\ \hline
all embed. & $ 0.71 \pm 0.11 $ & $ 0.2 \pm 0.21 $ & $ 0.13 \pm 0.13 $ \\ \hline
all topol. \& model  & $ 0.79 \pm 0.09 $ & $ 0.32 \pm 0.31 $ & $ 0.18 \pm 0.17 $ \\ \hline
all topol. \& embed.  & $ 0.77 \pm 0.11 $ & $ 0.24 \pm 0.23 $ & $ 0.17 \pm 0.15 $ \\ \hline
all model \& embed.  & $ 0.77 \pm 0.11 $ & $ 0.24 \pm 0.23 $ & $ 0.16 \pm 0.16 $ \\ \hline
all topol., model \& embed. & $ 0.78 \pm 0.1 $ & $ 0.27 \pm 0.23 $ & $ 0.17 \pm 0.15 $ \\ \hline
	\end{tabular}
	\label{tab:AUC_tech_dom}  
\end{table*} 

\begin{table*}[t!]
\centering
\caption{Average AUC, precision, and recall performances of the link prediction algorithms over 18 information networks as a subset of CommunityFitNet corpus. A random forest is used for supervised stacking of methods. Here, the predictors are adjusted for maximum F measure using a model selection through a cross validation on training set. The results are reported on 20\% holdout test set.}
\begin{tabular}{l|c|c|c}
\hline \hline
Algorithm & AUC & Precision & Recall\\ 
\hline\hline
Q  & $ 0.61  \pm 0.1 $ & $ 0.06  \pm  0.08 $ & $ 0.58 \pm 0.13 $ \\ \hline
Q-MR  & $ 0.59  \pm 0.1 $ & $ 0.04  \pm  0.05 $ & $ 0.57 \pm 0.15 $ \\ \hline
Q-MP  & $ 0.59  \pm 0.1 $ & $ 0.06  \pm  0.07 $ & $ 0.54 \pm 0.11 $ \\ \hline
B-NR (SBM)  & $ 0.79  \pm 0.14 $ & $ 0.13  \pm  0.2 $ & $ 0.58 \pm 0.2 $ \\ \hline
B-NR (DC-SBM)  & $ 0.72  \pm 0.14 $ & $ 0.12  \pm  0.19 $ & $ 0.61 \pm 0.17 $ \\ \hline
cICL-HKK  & $ 0.8  \pm 0.12 $ & $ 0.15  \pm  0.2 $ & $ 0.59 \pm 0.24 $ \\ \hline
B-HKK  & $ 0.76  \pm 0.13 $ & $ 0.18  \pm  0.19 $ & $ 0.46 \pm 0.24 $ \\ \hline
Infomap  & $ 0.79  \pm 0.08 $ & $ 0.09  \pm  0.1 $ & $ 0.74 \pm 0.11 $ \\ \hline
MDL (SBM)  & $ 0.8  \pm 0.13 $ & $ 0.16  \pm  0.2 $ & $ 0.57 \pm 0.25 $ \\ \hline
MDL (DC-SBM)  & $ 0.81  \pm 0.12 $ & $ 0.13  \pm  0.2 $ & $ 0.75 \pm 0.13 $ \\ \hline
S-NB  & $ 0.7  \pm 0.12 $ & $ 0.08  \pm  0.08 $ & $ 0.6 \pm 0.14 $ \\ \hline \hline
mean model-based &  $ 0.72  \pm 0.15 $ & $ 0.11  \pm  0.16 $ & $ 0.6 \pm 0.2 $ \\ \hline
mean indiv. topol. &  $ 0.61  \pm 0.12 $ & $ 0.07  \pm  0.13 $ & $ 0.48 \pm 0.31 $ \\ \hline
mean indiv. topol. \& model &  $ 0.63  \pm 0.13 $ & $ 0.08  \pm  0.14 $ & $ 0.51 \pm 0.29 $ \\ \hline \hline
emd-DW & $ 0.61 \pm 0.15 $ & $ 0.08 \pm 0.13 $ & $ 0.33 \pm 0.21 $ \\ \hline
emb-vgae & $ 0.65 \pm 0.15 $ & $ 0.04 \pm 0.04 $ & $ 0.65 \pm 0.19 $ \\ \hline \hline
all topol. & $ 0.83 \pm 0.12 $ & $ 0.32 \pm 0.25 $ & $ 0.39 \pm 0.25 $ \\ \hline
all model-based & $ 0.8 \pm 0.11 $ & $ 0.38 \pm 0.33 $ & $ 0.18 \pm 0.18 $ \\ \hline
all embed. & $ 0.77 \pm 0.12 $ & $ 0.3 \pm 0.28 $ & $ 0.17 \pm 0.27 $ \\ \hline
all topol. \& model  & $ 0.84 \pm 0.11 $ & $ 0.39 \pm 0.3 $ & $ 0.23 \pm 0.23 $ \\ \hline
all topol. \& embed.  & $ 0.81 \pm 0.15 $ & $ 0.32 \pm 0.27 $ & $ 0.27 \pm 0.26 $ \\ \hline
all model \& embed.  & $ 0.83 \pm 0.12 $ & $ 0.34 \pm 0.32 $ & $ 0.2 \pm 0.22 $ \\ \hline
all topol., model \& embed. & $ 0.83 \pm 0.12 $ & $ 0.36 \pm 0.28 $ & $ 0.26 \pm 0.27 $ \\ \hline
	\end{tabular}
	\label{tab:AUC_info_dom}  
\end{table*} 

\begin{table*}[t!]
\centering
\caption{Average AUC, precision, and recall performances of the link prediction algorithms over 38 transportation networks as a subset of CommunityFitNet corpus. A random forest is used for supervised stacking of methods. Here, the predictors are adjusted for maximum F measure using a model selection through a cross validation on training set. The results are reported on 20\% holdout test set. }
\begin{tabular}{l|c|c|c}
\hline \hline
Algorithm & AUC & Precision & Recall\\ 
\hline\hline
Q  & $ 0.68  \pm 0.09 $ & $ 0.07  \pm  0.07 $ & $ 0.6 \pm 0.09 $ \\ \hline
Q-MR  & $ 0.63  \pm 0.08 $ & $ 0.05  \pm  0.04 $ & $ 0.54 \pm 0.08 $ \\ \hline
Q-MP  & $ 0.63  \pm 0.1 $ & $ 0.07  \pm  0.07 $ & $ 0.56 \pm 0.11 $ \\ \hline
B-NR (SBM)  & $ 0.68  \pm 0.14 $ & $ 0.09  \pm  0.11 $ & $ 0.44 \pm 0.31 $ \\ \hline
B-NR (DC-SBM)  & $ 0.55  \pm 0.23 $ & $ 0.09  \pm  0.1 $ & $ 0.48 \pm 0.25 $ \\ \hline
cICL-HKK  & $ 0.69  \pm 0.13 $ & $ 0.1  \pm  0.14 $ & $ 0.52 \pm 0.26 $ \\ \hline
B-HKK  & $ 0.65  \pm 0.13 $ & $ 0.09  \pm  0.15 $ & $ 0.36 \pm 0.28 $ \\ \hline
Infomap  & $ 0.6  \pm 0.13 $ & $ 0.08  \pm  0.1 $ & $ 0.53 \pm 0.12 $ \\ \hline
MDL (SBM)  & $ 0.64  \pm 0.15 $ & $ 0.08  \pm  0.11 $ & $ 0.33 \pm 0.35 $ \\ \hline
MDL (DC-SBM)  & $ 0.81  \pm 0.07 $ & $ 0.09  \pm  0.1 $ & $ 0.72 \pm 0.1 $ \\ \hline
S-NB &  $ 0.66  \pm 0.12 $ & $ 0.07  \pm  0.08 $ & $ 0.61 \pm 0.1 $ \\ \hline \hline
mean model-based &  $ 0.66  \pm 0.15 $ & $ 0.08  \pm  0.1 $ & $ 0.52 \pm 0.24 $ \\ \hline
mean indiv. topol. &  $ 0.58  \pm 0.1 $ & $ 0.09  \pm  0.15 $ & $ 0.48 \pm 0.35 $ \\ \hline
mean indiv. topol. \& model &  $ 0.6  \pm 0.12 $ & $ 0.09  \pm  0.14 $ & $ 0.49 \pm 0.33 $ \\ \hline \hline
emd-DW & $ 0.62 \pm 0.15 $ & $ 0.2 \pm 0.21 $ & $ 0.29 \pm 0.2 $ \\ \hline
emb-vgae & $ 0.66 \pm 0.11 $ & $ 0.04 \pm 0.04 $ & $ 0.67 \pm 0.14 $ \\ \hline \hline
all topol. & $ 0.82 \pm 0.09 $ & $ 0.29 \pm 0.28 $ & $ 0.34 \pm 0.25 $ \\ \hline
all model-based & $ 0.76 \pm 0.11 $ & $ 0.29 \pm 0.28 $ & $ 0.22 \pm 0.23 $ \\ \hline
all embed. & $ 0.73 \pm 0.1 $ & $ 0.33 \pm 0.28 $ & $ 0.18 \pm 0.16 $ \\ \hline
all topol. \& model  & $ 0.83 \pm 0.09 $ & $ 0.34 \pm 0.33 $ & $ 0.25 \pm 0.24 $ \\ \hline
all topol. \& embed.  & $ 0.79 \pm 0.12 $ & $ 0.33 \pm 0.28 $ & $ 0.24 \pm 0.22 $ \\ \hline
all model \& embed.  & $ 0.78 \pm 0.11 $ & $ 0.35 \pm 0.27 $ & $ 0.22 \pm 0.21 $ \\ \hline
all topol., model \& embed. & $ 0.81 \pm 0.11 $ & $ 0.35 \pm 0.28 $ & $ 0.24 \pm 0.21 $ \\ \hline
	\end{tabular}
	\label{tab:AUC_trans_dom}  
\end{table*} 
\begin{table*}[t!]
\centering
\caption{Average AUC performance of the link prediction supervised stacking methods over 548 networks as a subset of CommunityFitNet corpus. A random forest is used for supervised stacking of methods. Here, the predictors are adjusted for maximum AUC using a model selection through a cross validation on training set. The results are reported on 20\% holdout test set. }
\begin{tabular}{l|l|l|l}
\hline \hline
Algorithm & AUC & Precision & Recall\\
\hline\hline
all topol. & $ 0.88 \pm 0.1 $ & $ 0.32 \pm 0.31 $ & $ 0.65 \pm 0.27 $ \\ \hline
all model-based & $ 0.87 \pm 0.11 $ & $ 0.25 \pm 0.26 $ & $ 0.64 \pm 0.28 $ \\ \hline
all embed. & $ 0.78 \pm 0.17 $ & $ 0.27 \pm 0.33 $ & $ 0.25 \pm 0.35 $ \\ \hline
all topol. \& model  & $ 0.89 \pm 0.09 $ & $ 0.33 \pm 0.32 $ & $ 0.64 \pm 0.28 $ \\ \hline
all topol. \& embed.  & $ 0.85 \pm 0.15 $ & $ 0.35 \pm 0.33 $ & $ 0.47 \pm 0.35 $ \\ \hline
all model \& embed.  & $ 0.85 \pm 0.14 $ & $ 0.31 \pm 0.31 $ & $ 0.46 \pm 0.34 $ \\ \hline
all topol., model \& embed. & $ 0.87 \pm 0.13 $ & $ 0.36 \pm 0.32 $ & $ 0.51 \pm 0.34 $ \\ \hline
	\end{tabular}
	\label{tab:rf_AUC}  
\end{table*}

\begin{table*}[t!]
\centering
\caption{Average AUC, precision, and recall performances of the link prediction algorithms over 548 networks as a subset of CommunityFitNet corpus. A XGBoost is used for supervised stacking of methods. Here, the predictors are adjusted for maximum F measure using a model selection through a cross validation on training set. The results are reported on 20\% holdout test set. }
\begin{tabular}{l|l|l|l}
\hline \hline
Algorithm & AUC & Precision & Recall\\ 
\hline\hline
all topol. & $ 0.85 \pm 0.11 $ & $ 0.45 \pm 0.32 $ & $ 0.39 \pm 0.33 $ \\ \hline
all model-based & $ 0.82 \pm 0.13 $ & $ 0.31 \pm 0.27 $ & $ 0.37 \pm 0.31 $ \\ \hline
all embed. & $ 0.77 \pm 0.16 $ & $ 0.32 \pm 0.3 $ & $ 0.35 \pm 0.33 $ \\ \hline
all topol. \& model  & $ 0.85 \pm 0.12 $ & $ 0.45 \pm 0.33 $ & $ 0.38 \pm 0.34 $ \\ \hline
all topol. \& embed.  & $ 0.83 \pm 0.14 $ & $ 0.41 \pm 0.34 $ & $ 0.38 \pm 0.34 $ \\ \hline
all model \& embed.  & $ 0.82 \pm 0.14 $ & $ 0.34 \pm 0.3 $ & $ 0.39 \pm 0.33 $ \\ \hline
all topol., model \& embed. & $ 0.84 \pm 0.13 $ & $ 0.41 \pm 0.34 $ & $ 0.38 \pm 0.35 $ \\ \hline
	\end{tabular}
	\label{tab:xgb_f}  
\end{table*} 

\begin{table*}[t!]
\centering
\caption{Average AUC, precision, and recall performances of the link prediction algorithms over 548 networks as a subset of CommunityFitNet corpus. A XGBoost is used for supervised stacking of methods. Here, the predictors are adjusted for maximum AUC using a model selection through a cross validation on training set. The results are reported on 20\% holdout test set.}
\begin{tabular}{l|l|l|l}
\hline \hline
Algorithm & AUC & Precision & Recall\\
\hline\hline
all topol. & $ 0.86 \pm 0.11 $ & $ 0.38 \pm 0.32 $ & $ 0.5 \pm 0.35 $ \\ \hline
all model-based & $ 0.84 \pm 0.12 $ & $ 0.24 \pm 0.25 $ & $ 0.55 \pm 0.34 $ \\ \hline
all embed. & $ 0.77 \pm 0.16 $ & $ 0.31 \pm 0.31 $ & $ 0.32 \pm 0.36 $ \\ \hline
all topol. \& model  & $ 0.87 \pm 0.11 $ & $ 0.38 \pm 0.33 $ & $ 0.49 \pm 0.36 $ \\ \hline
all topol. \& embed.  & $ 0.84 \pm 0.14 $ & $ 0.43 \pm 0.34 $ & $ 0.36 \pm 0.37 $ \\ \hline
all model \& embed.  & $ 0.83 \pm 0.13 $ & $ 0.31 \pm 0.3 $ & $ 0.44 \pm 0.36 $ \\ \hline
all topol., model \& embed. & $ 0.84 \pm 0.13 $ & $ 0.43 \pm 0.35 $ & $ 0.36 \pm 0.37 $ \\ \hline	\end{tabular}
	\label{tab:xgb_AUC}  
\end{table*} 

\begin{table*}[t!]
\centering
\caption{Average AUC, precision, and recall performances of the link prediction algorithms over 548 networks as a subset of CommunityFitNet corpus. An AdaBoost is used for supervised stacking of methods. Here, the predictors are adjusted for maximum F measure using a model selection through a cross validation on training set. The results are reported on 20\% holdout test set. }
\begin{tabular}{l|l|l|l}
\hline \hline
Algorithm & AUC & Precision & Recall\\
\hline\hline
all topol. & $ 0.82 \pm 0.13 $ & $ 0.4 \pm 0.34 $ & $ 0.42 \pm 0.33 $ \\ \hline
all model-based & $ 0.79 \pm 0.14 $ & $ 0.31 \pm 0.31 $ & $ 0.4 \pm 0.31 $ \\ \hline
all embed. & $ 0.74 \pm 0.16 $ & $ 0.27 \pm 0.32 $ & $ 0.36 \pm 0.3 $ \\ \hline
all topol. \& model  & $ 0.81 \pm 0.13 $ & $ 0.38 \pm 0.36 $ & $ 0.43 \pm 0.34 $ \\ \hline
all topol. \& embed.  & $ 0.8 \pm 0.14 $ & $ 0.33 \pm 0.35 $ & $ 0.45 \pm 0.32 $ \\ \hline
all model \& embed.  &  $ 0.79 \pm 0.14 $ & $ 0.29 \pm 0.33 $ & $ 0.46 \pm 0.32 $ \\ \hline
all topol., model \& embed. & $ 0.81 \pm 0.14 $ & $ 0.33 \pm 0.35 $ & $ 0.44 \pm 0.33 $ \\ \hline
\end{tabular}
	\label{tab:ada_f}  
\end{table*} 

\begin{table*}[t!]
\centering
\caption{Average AUC, precision, and recall performances of the link prediction algorithms over 548 networks as a subset of CommunityFitNet corpus. An AdaBoost is used for supervised stacking of methods. Here, the predictors are adjusted for maximum AUC using a model selection through a cross validation on training set. The results are reported on 20\% holdout test set.}
\begin{tabular}{l|l|l|l}
\hline \hline
Algorithm & AUC & Precision & Recall\\
\hline\hline
all topol. & $ 0.86 \pm 0.12 $ & $ 0.3 \pm 0.3 $ & $ 0.62 \pm 0.3 $ \\ \hline
all model-based & $ 0.83 \pm 0.13 $ & $ 0.25 \pm 0.29 $ & $ 0.57 \pm 0.32 $ \\ \hline
all embed. & $ 0.76 \pm 0.16 $ & $ 0.25 \pm 0.33 $ & $ 0.41 \pm 0.32 $ \\ \hline
all topol. \& model  & $ 0.85 \pm 0.12 $ & $ 0.32 \pm 0.35 $ & $ 0.58 \pm 0.34 $ \\ \hline
all topol. \& embed.  & $ 0.82 \pm 0.14 $ & $ 0.31 \pm 0.36 $ & $ 0.51 \pm 0.35 $ \\ \hline
all model \& embed.  & $ 0.8 \pm 0.14 $ & $ 0.26 \pm 0.31 $ & $ 0.5 \pm 0.33 $ \\ \hline
all topol., model \& embed. & $ 0.82 \pm 0.13 $ & $ 0.29 \pm 0.35 $ & $ 0.51 \pm 0.36 $ \\ \hline
\end{tabular}
	\label{tab:ada_AUC}  
\end{table*} 


\begin{thebibliography}{10}

\bibitem{kossinets2006effects}
Kossinets G (2006) Effects of missing data in social networks.
\newblock {\em Social Networks} 28(3):247--268.

\bibitem{fire2013computationally}
Fire M, et~al. (2013) Computationally efficient link prediction in a variety of
  social networks.
\newblock {\em ACM Transactions on Intelligent Systems and Technology (TIST)}
  5(1):10.

\bibitem{lu2011link}
L{\"u} L, Zhou T (2011) Link prediction in complex networks: A survey.
\newblock {\em Physica A: statistical mechanics and its applications}
  390(6):1150--1170.

\bibitem{nagarajan2015predicting}
Nagarajan M, et~al. (2015) Predicting future scientific discoveries based on a
  networked analysis of the past literature in {\em Proceedings of the 21th ACM
  SIGKDD International Conference on Knowledge Discovery and Data Mining}.
\newblock (ACM), pp. 2019--2028.

\bibitem{kane2012s}
Kane GC, Alavi M, Labianca GJ, Borgatti S (2014) What’s different about
  social media networks? a framework and research agenda.
\newblock {\em MIS Quarterly} 38(1):274--304.

\bibitem{burgess2016link}
Burgess M, Adar E, Cafarella M (2016) Link-prediction enhanced consensus
  clustering for complex networks.
\newblock {\em {PLoS ONE}} 11(5):e0153384.

\bibitem{mirshahvalad2012significant}
Mirshahvalad A, Lindholm J, Derlen M, Rosvall M (2012) Significant communities
  in large sparse networks.
\newblock {\em PloS one} 7(3):e33721.

\bibitem{ghasemian2018evaluating}
Ghasemian A, Hosseinmardi H, Clauset A (2019) Evaluating overfit and underfit
  in models of network community structure.
\newblock {\em IEEE Trans.\ Knowledge and Data Engineering (TKDE)}.

\bibitem{valles2018consistencies}
Vall{\`e}s-Catal{\`a} T, Peixoto TP, Sales-Pardo M, Guimer{\`a} R (2018)
  Consistencies and inconsistencies between model selection and link prediction
  in networks.
\newblock {\em Physical Review E} 97(6):062316.

\bibitem{arlot2010survey}
Arlot S, Celisse A, , et~al. (2010) A survey of cross-validation procedures for
  model selection.
\newblock {\em Statistics Surveys} 4:40--79.

\bibitem{trevor2009elements}
Hastie T, Tibshirani R, Friedman J (2009) {\em The elements of statistical
  learning: data mining, inference, and prediction}.
\newblock (New York, NY: Springer).

\bibitem{clauset2008hierarchical}
Clauset A, Moore C, Newman MEJ (2008) Hierarchical structure and the prediction
  of missing links in networks.
\newblock {\em Nature} 453(7191):98.

\bibitem{martinez2017survey}
Mart{\'\i}nez V, Berzal F, Cubero JC (2017) A survey of link prediction in
  complex networks.
\newblock {\em ACM Computing Surveys (CSUR)} 49(4):69.

\bibitem{al2011survey}
Al~Hasan M, Zaki MJ (2011) A survey of link prediction in social networks in
  {\em Social Network Data Analytics}.
\newblock (Springer), pp. 243--275.

\bibitem{liben2007link}
Liben-Nowell D, Kleinberg J (2007) The link-prediction problem for social
  networks.
\newblock {\em Journal of the Association for Information Science and
  Technology} 58(7):1019--1031.

\bibitem{zhou2009predicting}
Zhou T, L{\"u} L, Zhang YC (2009) Predicting missing links via local
  information.
\newblock {\em The European Physical Journal B} 71(4):623--630.

\bibitem{grover2016node2vec}
Grover A, Leskovec J (2016) node2vec: Scalable feature learning for networks in
  {\em Proceedings of the 22nd ACM SIGKDD International Conference on Knowledge
  Discovery and Data Mining}.
\newblock (ACM), pp. 855--864.

\bibitem{cai2018comprehensive}
Cai H, Zheng VW, Chang KCC (2018) A comprehensive survey of graph embedding:
  Problems, techniques, and applications.
\newblock {\em IEEE Transactions on Knowledge and Data Engineering}
  30(9):1616--1637.

\bibitem{wolpert:macready:1997}
Wolpert DH, Macready WG (1997) No free lunch theorems for optimization.
\newblock {\em IEEE Transactions on Evolutionary Computation} 1(1):67--82.

\bibitem{peel2017ground}
Peel L, Larremore DB, Clauset A (2017) The ground truth about metadata and
  community detection in networks.
\newblock {\em Science Advances} 3(5):e1602548.

\bibitem{schapire1990strength}
Schapire RE (1990) The strength of weak learnability.
\newblock {\em Machine Learning} 5(2):197--227.

\bibitem{breiman1996bagging}
Breiman L (1996) Bagging predictors.
\newblock {\em Machine Learning} 24(2):123--140.

\bibitem{srivastava2014dropout}
Srivastava N, Hinton G, Krizhevsky A, Sutskever I, Salakhutdinov R (2014)
  Dropout: a simple way to prevent neural networks from overfitting.
\newblock {\em Journal of Machine Learning Research} 15(1):1929--1958.

\bibitem{wolpert1992stacked}
Wolpert DH (1992) Stacked generalization.
\newblock {\em Neural Networks} 5(2):241--259.

\bibitem{schapire1999brief}
Schapire RE (1999) A brief introduction to boosting in {\em Proceedings of the
  16th International Joint Conference on Artificial intelligence, Volume 2}.
\newblock (Morgan Kaufmann Publishers Inc.), pp. 1401--1406.

\bibitem{koren2009bellkor}
Koren Y (2009) The {BellKor} solution to the {Netflix Grand Prize}.
\newblock {\em Netflix prize documentation 81} pp. 1--10.

\bibitem{freund:schapire:1997}
Freund Y, Schapire RE (1997) A decision-theoretic generalization of on-line
  learning and an application to boosting.
\newblock {\em Journal of Computer and System Sciences} 55:119--139.

\bibitem{epasto2019single}
Epasto A, Perozzi B (2019) Is a single embedding enough? {Learning} node
  representations that capture multiple social contexts in {\em The World Wide
  Web Conference}.
\newblock (ACM), pp. 394--404.

\bibitem{makridakis2018m4}
Makridakis S, Spiliotis E, Assimakopoulos V (2018) The {M4} competition:
  {Results}, findings, conclusion and way forward.
\newblock {\em International Journal of Forecasting} 34(4):802--808.

\bibitem{makridakis2018statistical}
Makridakis S, Spiliotis E, Assimakopoulos V (2018) Statistical and machine
  learning forecasting methods: Concerns and ways forward.
\newblock {\em {PLoS ONE}} 13(3):e0194889.

\bibitem{newman:networks:2018}
Newman M (2019) {\em Networks}.
\newblock (Oxford University Press).

\bibitem{cukierski2011graph}
Cukierski W, Hamner B, Yang B (2011) Graph-based features for supervised link
  prediction in {\em Neural Networks (IJCNN), The 2011 International Joint
  Conference on}.
\newblock (IEEE), pp. 1237--1244.

\bibitem{networkx}
Hagberg A, Swart P, S~Chult D (2008) Exploring network structure, dynamics, and
  function using networkx, (Los Alamos National Lab.(LANL), Los Alamos, NM
  (United States)), Technical report.

\bibitem{leicht2006vertex}
Leicht EA, Holme P, Newman MEJ (2006) Vertex similarity in networks.
\newblock {\em Physical Review E} 73(2):026120.

\bibitem{newman2004finding}
Newman MEJ, Girvan M (2004) Finding and evaluating community structure in
  networks.
\newblock {\em Phys. Rev. E} 69(2):026113.

\bibitem{newman2016community}
Newman MEJ (2016) Community detection in networks: Modularity optimization and
  maximum likelihood are equivalent.
\newblock {\em arXiv:1606.02319}.

\bibitem{zhang2014scalable}
Zhang P, Moore C (2014) Scalable detection of statistically significant
  communities and hierarchies, using message passing for modularity.
\newblock {\em Proc. Natl. Acad. Sci. USA} 111(51):18144--18149.

\bibitem{newman2016estimating}
Newman MEJ, Reinert G (2016) Estimating the number of communities in a network.
\newblock {\em Phys. Rev. Lett.} 117(7):078301.

\bibitem{hayashi2016tractable}
Hayashi K, Konishi T, Kawamoto T (2016) A tractable fully {Bayesian} method for
  the stochastic block model.
\newblock {\em arXiv:1602.02256}.

\bibitem{rosvall2008maps}
Rosvall M, Bergstrom CT (2008) Maps of random walks on complex networks reveal
  community structure.
\newblock {\em Proc. Natl. Acad. Sci. USA} 105(4):1118--1123.

\bibitem{peixoto2013parsimonious}
Peixoto TP (2013) Parsimonious module inference in large networks.
\newblock {\em Phys. Rev. Lett.} 110(14):148701.

\bibitem{Krzakala2013}
Krzakala F, et~al. (2013) Spectral {R}edemption in {C}lustering {S}parse
  {N}etworks.
\newblock {\em Proc. Natl. Acad. Sci.} 110(52):20935--20940.

\bibitem{perozzi2014deepwalk}
Perozzi B, Al-Rfou R, Skiena S (2014) Deepwalk: Online learning of social
  representations in {\em Proceedings of the 20th ACM SIGKDD International
  Conference on Knowledge Discovery and Data Mining}.
\newblock (ACM), pp. 701--710.

\bibitem{kipf2016variational}
Kipf TN, Welling M (2016) Variational graph auto-encoders.
\newblock {\em preprint arXiv:1611.07308}.

\bibitem{hamilton2017representation}
Hamilton WL, Ying R, Leskovec J (2017) Representation learning on graphs:
  Methods and applications.
\newblock {\em preprint arXiv:1709.05584}.

\bibitem{dietterich2000ensemble}
Dietterich T (2000) Ensemble methods in machine learning.
\newblock {\em Multiple Classifier Systems} pp. 1--15.

\bibitem{sewell2008ensemble}
Sewell M (2008) Ensemble learning.
\newblock {\em RN} 11(02).

\bibitem{chen2016xgboost}
Chen T, Guestrin C (2016) Xgboost: A scalable tree boosting system in {\em
  Proceedings of the 22nd ACM SIGKDD International Conference on Knowledge
  Discovery and Data Mining}.
\newblock pp. 785--794.

\bibitem{freund1997decision}
Freund Y, Schapire RE (1997) A decision-theoretic generalization of on-line
  learning and an application to boosting.
\newblock {\em Journal of computer and system sciences} 55(1):119--139.

\bibitem{karrer2011stochastic}
Karrer B, Newman MEJ (2011) Stochastic blockmodels and community structure in
  networks.
\newblock {\em Physical review E} 83(1):016107.

\bibitem{decelle2011asymptotic}
Decelle A, Krzakala F, Moore C, Zdeborov{\'a} L (2011) Asymptotic {A}nalysis of
  the {S}tochastic {B}lock {M}odel for {M}odular {N}etworks and {I}ts
  {A}lgorithmic {A}pplications.
\newblock {\em Phys. Rev. E} 84(6):066106.

\bibitem{clauset2016ICON}
Clauset A, Tucker E, Sainz M (2016) {The Colorado Index of Complex Networks}.
  (\url{https://icon.colorado.edu/}).

\bibitem{al2006link}
Al~Hasan M, Chaoji V, Salem S, Zaki M (2006) Link prediction using supervised
  learning in {\em SDM06: workshop on link analysis, counter-terrorism and
  security}.

\bibitem{ahmed2016supervised}
Ahmed C, ElKorany A, Bahgat R (2016) A supervised learning approach to link
  prediction in twitter.
\newblock {\em Social Network Analysis and Mining} 6(1):24.

\bibitem{lichtenwalter2010new}
Lichtenwalter RN, Lussier JT, Chawla NV (2010) New perspectives and methods in
  link prediction in {\em Proceedings of the 16th ACM SIGKDD international
  conference on Knowledge discovery and data mining}.
\newblock (ACM), pp. 243--252.

\bibitem{cover2012elements}
Cover TM, Thomas JA (2012) {\em Elements of information theory}.
\newblock (John Wiley \& Sons).

\end{thebibliography}

\end{document}